\documentclass[10pt,twocolumn,letterpaper]{article}

\usepackage{cvpr_arxiv}
\usepackage{times}
\usepackage{epsfig}
\usepackage{graphicx}
\usepackage{amsmath}
\usepackage{amssymb}
\usepackage[]{algorithm2e}
\usepackage{subfigure}
\usepackage{placeins}
\usepackage{hyphenat}

\usepackage[pagebackref=true,breaklinks=true,letterpaper=true,colorlinks,bookmarks=false]{hyperref}

\cvprfinalcopy % *** Uncomment this line for the final submission

\def\BibPath{.}  

\def\cvprPaperID{622} % *** Enter the CVPR Paper ID here

\begin{document}

%%%%%%%%% TITLE
\title{Using Self-Contradiction to Learn Confidence Measures in Stereo Vision}

\author{
Christian Mostegel
~~~~~~~~~~
Markus Rumpler
~~~~~~~~~~
Friedrich Fraundorfer
~~~~~~~~~~
Horst Bischof\\
Institute for Computer Graphics and Vision, Graz University of Technology\thanks{The research leading to these results has received funding from the EC FP7 project 3D-PITOTI (ICT-2011-600545)
and from the Austrian Research Promotion Agency (FFG) as Vision+ project 836630 and together with OMICRON electronics GmbH as Bridge1 project 843450.}\\
{\tt\small \{surname\}@icg.tugraz.at}
}

\maketitle
%\thispagestyle{empty}

%%%%%%%%% ABSTRACT
\begin{abstract}

Learned confidence measures gain increasing importance for outlier removal 
and quality improvement in stereo vision.
However, acquiring the necessary training data is typically a tedious and 
time consuming task that involves manual interaction, active sensing devices
and/or synthetic scenes.
To overcome this problem, we propose a new, flexible, and scalable way 
for generating training data that only requires a set of stereo images as input.
The key idea of our approach is to use 
different view points for reasoning about contradictions and consistencies between multiple depth maps
generated with the same stereo algorithm.
This enables us to generate a huge amount of training data
in a fully automated manner.
Among other experiments, we demonstrate the potential of our approach 
by boosting the performance of three learned confidence measures on the KITTI2012 dataset by simply training
them on a vast amount of automatically generated training data rather than a limited amount of laser ground truth data.

\end{abstract}

%%%%%%%%% BODY TEXT
\section{Introduction}

Many works have demonstrated that machine learning can be 
greatly beneficial for stereo vision~\cite{park15,spyro14,zbontar15a,peris12,haeusler13}.
All these works have one thing in common:\\
They require training data -- the more the better.

\begin{figure}
  \centering
%   \vspace{-15pt}
\includegraphics[width=0.9\columnwidth]{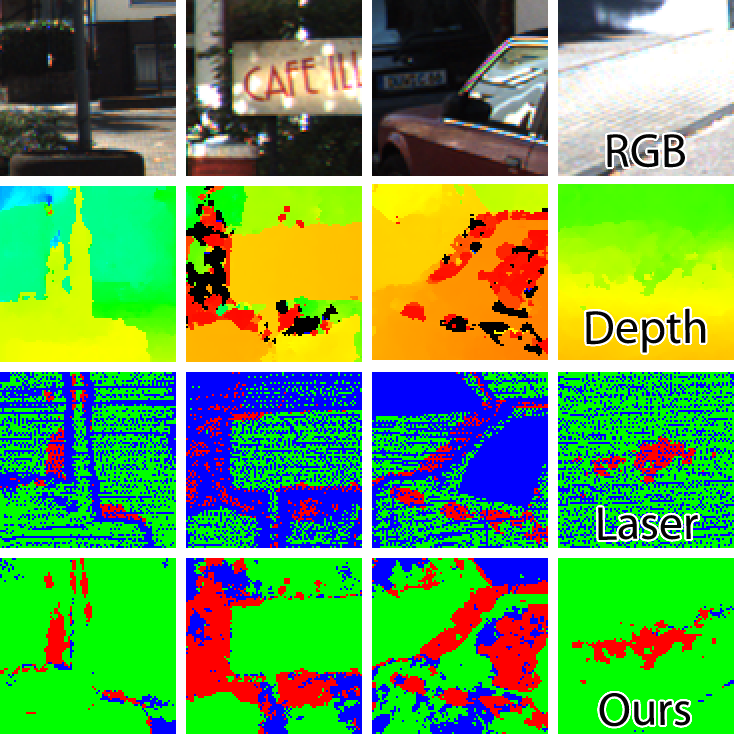} 
    
  \caption{Our approach automatically detects and classifies contradictions and consistencies between multiple depth maps
  to generate labeled training images, which can be used for training confidence measures.
  From top to bottom: RGB input images, depth maps created with a query stereo algorithm (here~\cite{rothermel12}),
  label images based on laser ground truth~\cite{geiger12} and
  our automatically generated label images.
  In the label images, green stands for positive samples,
  red for negative and blue is ignored during training.}
  \vspace{-8pt}
  \label{fig:concept}
\end{figure}
Previous approaches used three main sources of training data.
The first source is manual labeling. While this is the traditional approach 
in the fields of classification and segmentation (e.g.~\cite{everingham10,torralba10,shotton09}),
it requires hundreds of man-hours even in 2D.
Because the task becomes even more taxing in 3D, only very few manually labeled datasets exist in this domain (e.g.~\cite{ladicky12}).
The second source is synthetic data generation~\cite{butler12,peris12}.
Unfortunately, pure synthetic data generation has not yet reached the level where it generalizes to natural 
images without an extreme modeling effort.
The third source is to record ground truth data with active depth sensors,
which is currently the most popular source~\cite{strecha08dataset,geiger12,menze15,scharstein14}.
If a projector based setup is used~\cite{scharstein14}, the ground truth can 
achieve a very high accuracy, but the data acquisition takes a lot of time and 
is restricted to indoor scenes.
For outdoor scenes the method of choice is typically the use of a laser scanner~\cite{strecha08dataset,geiger12,menze15}.
Aside from requiring a non-trivial registration between the 
laser reconstruction and the recorded images, 
this method is also
subject to a range of assumptions itself.
This fact makes a manual removal of obviously incorrect ground truth data necessary
for outdoor datasets~\cite{geiger12,menze15}.
Some approaches, like~\cite{menze15}, combine these three sources.
They combine active sensing with synthetic car models and manual annotation
to increase the quality of ground truth data.

None of these methods 
is easily portable to very specific application areas
such as under water reconstruction or 3D reconstruction with micro aerial vehicles.
Furthermore, none of these methods shows good scaling properties in the sense 
of required man-hours per training data.

This motivated us to propose a novel way of generating training data without a
synthetic model, active devices or manual interaction.
Instead of explicitly generating a ground truth,
we compare multiple depth maps of the same scene obtained with the same stereo approach with each 
other and thus collect positive and negative training data.
As input we require a set of stereo images
observing a static scene from multiple viewing angles.
After computing the relative poses and generating depth maps, we evaluate which parts of the depth maps can likely  be  trusted,
which parts contradict each other and for which parts we simply do not have enough information
available to make this decision.
This results in a set of partially labeled images, similar to ground truth, which can be used for training (see Fig.~\ref{fig:concept}).

For the evaluation of our method we use three publicly available datasets,
which are namely the multi-view stereo dataset of Strecha et al.~\cite{strecha08dataset},
the Middlebury2014 dataset~\cite{scharstein14} and the KITTI2012 stereo dataset~\cite{geiger12}.
On these datasets we demonstrate that 
the performance of learned confidence measures can be boosted by simply training them
on large amounts of domain specific training data,
which our approach can cheaply provide.

%-------------------------------------------------------------------------

\section{Related Work}
To the best of our knowledge, we are the first to approach the topic of 
stereo training data generation in a self-supervised manner.
In order to demonstrate the usefulness of our groundtruth generation,
we show that existing stereo vision approaches, which
already have been evaluated on one or more of the afore mentioned stereo datasets,
can also be successfully trained on our generated data.

Thus, we first give a short overview of the most relevant learning based stereo approaches.
While most approaches pose the problem of learning reconstruction errors as a binary classification
problem (correct matches/depth values versus incorrect matches/depth values),
Kong and Tao~\cite{kong04} propose to use an additional class for failures due to
foreground fattening. Using these predicted class probabilities
they adjust the initial matching cost.
Peris et al.~\cite{peris12} train a multi-class  Linear Discriminant Analysis (LDA) classifier
to compute disparities together with a confidence map. 
\v{Z}bontar and LeCun~\cite{zbontar15a}
improve the matching cost computation by learning a similarity measure between small image patches using a convolutional neural network.

The approaches mentioned above are very specific in their formulation, but many other works
use a so called "confidence measure" as a basis for improving the stereo output.
A confidence measure should predict the likelihood of a depth value being correct
and is typically computed using image intensities, disparity values and/or matching costs.
Some surveys about confidence measures are available in~\cite{hu12,egnal04,egnal02}.
In the simplest way a confidence measure can be used to remove very likely 
wrong measurements from the depth map.
This process is called sparsification. The most common
way for sparsification without training is the left-right consistency check~\cite{hu12}.
While this check already detects many outliers, it cannot detect errors
caused by a systematic problem of an approach (e.g. foreground fattening).
Haeusler et al.~\cite{haeusler13} showed that ensemble learning
of many different features with random decision forests can significantly improve the sparsification
performance.
Note that confidence measures are also learned in similar fashion in the domain of optical flow, e.g.~\cite{aodha13,gehrig11}.
Spyropoulos et al.~\cite{spyro14} used the confidence prediction as a soft-constraint
in a Markov random field to improve the stereo output.
In the very recent work of Park and Yoon~\cite{park15} the confidence
prediction is used to modulate the matching cost of a semi-global matcher~\cite{hirschmuller08} and 
thus increase its performance.
As the performance of the above mentioned approaches depends
on how well the confidence of a measurement can be predicted,
the area under the sparsification curve is one of the most important
evaluation criteria in this domain~\cite{haeusler13,park15}.
Hence, we 
found that this criterion is ideally suited to benchmark the quality of our training data generation,
aside from comparing it directly to the ground truth.
In our experiments, we use three recent approaches~\cite{haeusler13,spyro14,park15} that compute
confidence measures and analyze the change of performance depending on the used training data (laser ground truth vs. automatically generated training data) on the KITTI2012 dataset~\cite{geiger12}.

Aside from stereo vision, there exist
some works that deal with learning the 
matchability of features.
Some of these works~\cite{brown11,trzcinski12,han15} use ground truth data collected by~\cite{brown11}.
To generate the ground truth data they use the dense multi-view stereo reconstruction algorithm provided by Goesele~et~al.~\cite{goesele07}
and trust this approach to be accurate enough.
The problem with applying this approach to dense stereo is that a learning algorithm will try to tune its output
to reproduce any systematic error made by~\cite{goesele07}.
Philbin et al.~\cite{philbin10} use SIFT~\cite{Lowe04SIFT} nearest-neighbors together with a RANSAC verification to generate 
negative and positive training data,
whereas Simonyan et al.~\cite{simonyan14} first compute a homography between images using SIFT and RANSAC and
then establish region correspondences using the homography.
Hartmann~et~al.~\cite{hartmann14matchability} learn the matchability of SIFT features
by
 collecting features that survive the matching stage and
those which are rejected as positive and negative training data.
All of these approaches 
 focus on a specific type of sparse feature
 and do not generalize well to dense stereo.

%-------------------------------------------------------------------------

\section{Fully Automatic Training Data Generation}

\begin{figure}
  \centering
\includegraphics[width=1\columnwidth]{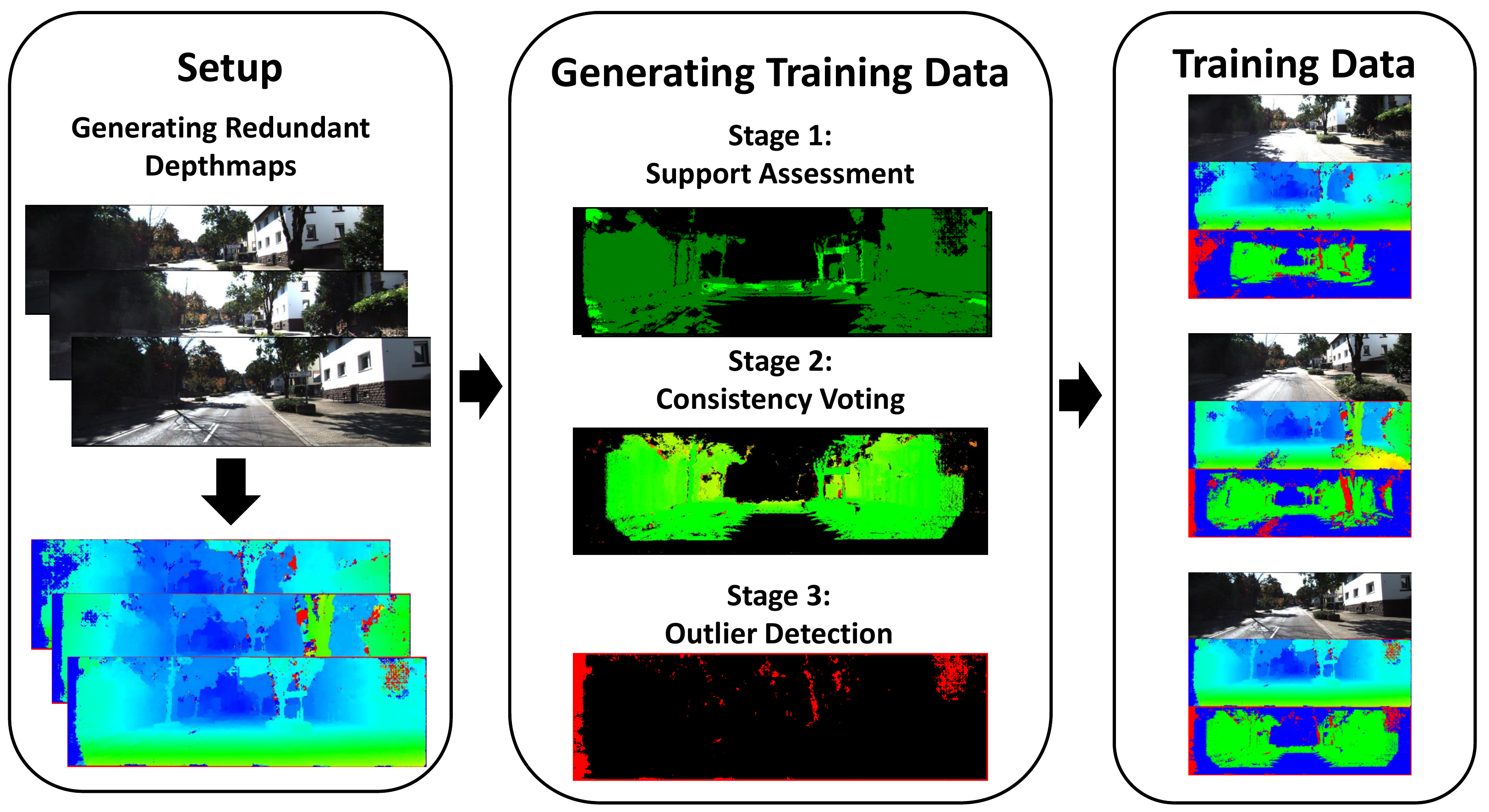} 
    
  \caption{Fully Automatic Training Data Generation.
  }
  \vspace{-15pt}
  \label{fig:data_generation}
\end{figure}

The core idea of our training data generation approach is to relax the aim from 
labeling each pixel as \emph{correct/incorrect} to
labeling them as \emph{self-consistent/contradicting}.
We use the word \emph{self-consistent} in the sense that
depth maps generated with the same algorithm from \emph{different view points}
shall not contradict each other through free space violation or occlusion.
Note that the use of \emph{different view points} is important as the errors in depth maps
are in general 
strongly correlated, if they are generated from the same view point with the same stereo algorithm.
In this work we use the observation that this correlation is small when
the relative observation angle between two depth maps is large to 
reduce the influence of systematic errors.
The basic steps of our approach are visualized in Fig.~\ref{fig:data_generation}.

As input our approach requires a set of stereo images with known poses.
First, we execute the \emph{query algorithm}, which yields a set of depth maps (Setup).
Then we assess which parts of a depth map are supported by other depth maps 
with a significantly different observation angle (Stage 1).
In the next stage (Stage 2), we then use this support to 
influence the voting process.
In the final stage (Stage 3), we detect outliers which were missed in the previous stage using an 
augmented depth map.
In the remainder of this section, we describe all involved steps in more detail.

\subsection{Stage 1: Support Assessment}

Given many depth maps of the same scene,
we want to separate parts of the scene 
where many depth maps agree on the structure (consistent parts) from 
those where they either disagree or 
we simply do not have enough view points to rule out 
systematic errors.

In performing this separation, we have to account for two problems.
The first problem is that if the camera poses of two stereo pairs
are too similar, the depth maps will very likely contain the same systematic error.
To remedy this situation, we try to decrease the error correlation
in using different observation angles.
The second problem originates from the finite precision of cameras, which introduces 
a depth uncertainty. This means
that a measurement with a high  uncertainty is not well suited for determining whether a 
measurement with a lower uncertainty is correct or not. To estimate
this uncertainty we use the model proposed by~\cite{haner11vslam_viewplanning}.
This model allows us to compute a covariance matrix for each
3D point corresponding to a depth value
through first-order backward covariance propagation under the assumption of isotropic Gaussian image noise, 
which is explained in more detail in~\cite{hartley04multiview}.

As we aim to produce 2D label images, we address this problem on a per-pixel basis.
So for each pixel of a depth map, we first collect the support of other depth maps. 
A reference
depth map is only allowed to express its support for the 3D point $\mathbf{p}_{\text{query}}$ associated with a query pixel
if it fulfills the following two criteria.

First, the viewpoints shall be sufficiently different.
We define that a viewpoint is different enough if the observation angle difference between two stereo pairs
 is sufficiently large ( $\alpha_\text{diff} > \alpha_\text{min}$).
We compute this observation angle as
$\alpha_{\text{diff}} = \measuredangle (\overrightarrow{\mathbf{p}_{\text{query}} \mathbf{c}_{\text{ref}} },\overrightarrow{\mathbf{p}_{\text{query}} \mathbf{c}_{\text{query}} })$,
where $\mathbf{c}_{\text{x}}$ is the mean camera center of a stereo pair.

Second, the reference measurement shall be within a fixed theoretical tolerance $\sigma_\text{max}$ of the query measurement.
For this evaluation we use the Mahalanobis distance based on 
the covariance matrix with the smaller uncertainty (either reference or query).

In order to avoid being too much biased by a single observation direction,
a 3D point fulfilling these criteria is not directly allowed to vote, but 
instead can activate its corresponding bin depending on the observation angle.
We use angular bins of $\alpha_\text{min}$ degree,
where each activated bin increases the \emph{support} for the query point by one.

\subsection{Stage 2: Consistency Voting}

 \begin{figure}
  \centering
\includegraphics[width=0.8\columnwidth]{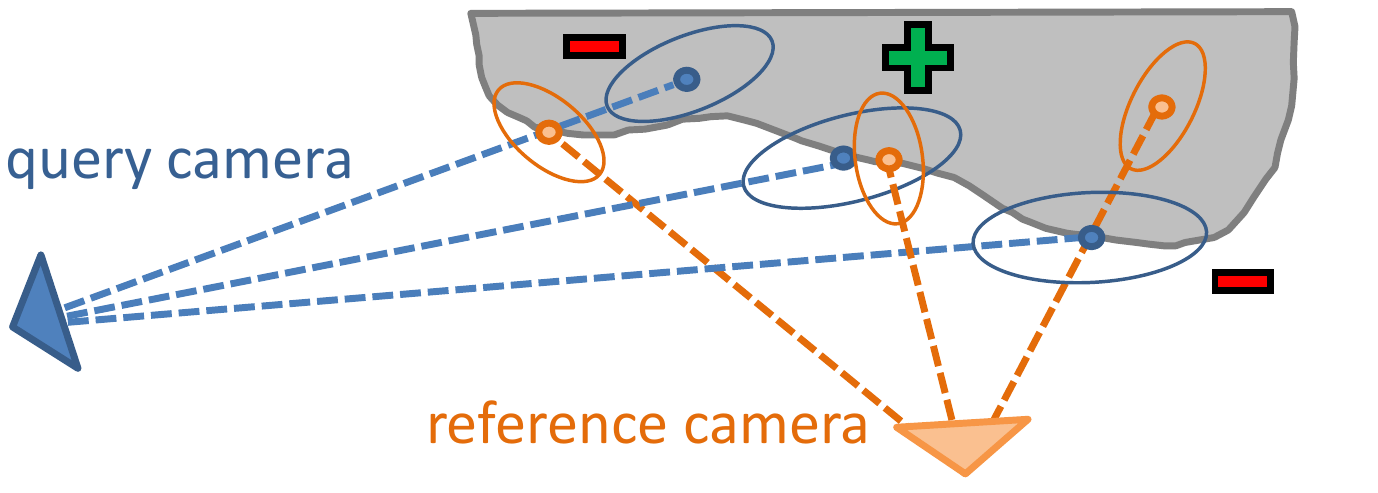}     
  \caption{Consistency Voting. There are three possibilities for voting. A positive vote (center) is only cast
  if the reference measurement is within the uncertainty boundary of the query measurement.
  A negative vote is either cast if a reference measurement would block the line of sight of the query camera (left) or 
  the other way around (right).}
  \vspace{-15pt}
  \label{fig:supp_n_contra}
\end{figure}
The basic idea of this stage is to let all depth maps vote for the (in)consistency 
of a query depth map.
Similar to works in depth map fusion (e.g.~\cite{merrel07}), negative votes are cast by free space violations and occlusions
and positive votes are cast by measurements which are sufficiently close to each other (see Fig.~\ref{fig:supp_n_contra}).
Opposed to fusion approaches, we aim for a completely different output.
While works in depth map fusion try to improve/fuse the depth map,
we only aim to decide
which parts of the depth map cause contradictions and which parts are sufficiently consistent.
Furthermore, we have to reduce the influence of systematic errors in the voting scheme,
which we achieve with the support of a reference measurement computed in the previous stage.
In particular this means that only parts which have a support from at least one significantly different observation angle 
are eligible for voting.

The proposed voting scheme looks as follows.
For casting a positive vote $v_{+}$ a reference measurement has to fulfill two properties.
First, it shall be more accurate than
the query measurement. We evaluate this property with the largest Eigen value of the corresponding
covariance matrix.
Second, the reference measurement has to be within a fixed theoretical
tolerance of $\sigma_\text{max}$ of the query measurement. For this evaluation we use the Mahalanobis distance based on 
the covariance matrix of the query 3D point.
We define a positive vote as:
\begin{equation}
 v_{+} = \sqrt{ i_{\text{ref}} } \cdot \text{\emph{support}}_\text{ref}
\end{equation}
where $i_{\text{ref}}$ is the smallest Eigen value of the Fisher information matrix
of the reference 3D point. This means that measurements with a low theoretic uncertainty
get a higher voting strength, as $\sqrt{ i_{\text{ref}} }  = 1/\sqrt{u_{\text{ref}}} $,
where $u_{\text{ref}}$ is the largest Eigen value of the covariance matrix and hence $\sqrt{u_{\text{ref}}}$ can be interpreted 
as the standard deviation along the axis of the highest uncertainty.

For casting a negative vote a reference measurement has to fulfill three properties.
First, it also has to be more accurate than the query measurement.
Second, it has to be outside the fixed theoretical
tolerance of $\sigma_\text{max}$.
Third, it has to cause a free space violation or occlusion as depicted in Fig.~\ref{fig:supp_n_contra}.
In a free space violation, a reference measurement would block the line of sight of a query measurement (left side in  Fig.~\ref{fig:supp_n_contra}),
whereas the other way around would cause an occlusion (right side in  Fig.~\ref{fig:supp_n_contra}).
If these properties are met, a negative vote is cast:
\begin{equation}
 v_{-} = - \sqrt{ i_{\text{ref}} } \cdot \text{\emph{support}}_\text{ref}
\end{equation}

For each pixel in the query depth map the votes are collected.
The label of a pixel with more than zero votes is then set depending 
on the sign of the final sum of votes.

\subsection{Stage 3: Outlier Detection}
In the previous step, we only allowed measurements with a minimal support from a different
observation angle to vote and only then if they are more accurate than the query measurement.
This restriction is necessary because otherwise the training data would contain
a great percentage of incorrectly labeled samples, i.e. false positives (consistent but incorrect) and 
false negatives (inconsistent but correct).
However, this also causes many regions to be missed
in which absolutely no consensus can be reached, because they only contain outliers (e.g. top left corner in Fig.~\ref{fig:example_labels_kitti}).
In this stage, we aim to detect these outliers for enhancing our negative training data.

First, we label trivial outliers which either lie behind the camera or 
do not project into the second stereo camera.
For the remaining unlabeled regions we compare the depth values of the query camera
to a specially augmented depth map.
Our procedure for obtaining this augmented depth map is inspired by the stability-based depth map fusion proposed by Merrell~et~al.~\cite{merrel07}.
The main difference is that we do not aim for high performance or even the perfect depth map,
but a depth map which rather prefers lower depth values which are sufficiently plausible.
We found that underestimating the depth values helps us to keep the number of false negatives (inconsistent but correct) low,
while at the same time allowing us to recover many true negatives (inconsistent and incorrect).
Further, we avoid any smoothness assumptions to preserve fine objects.

For computing the augmented depth map, we collect all depth values of the other depth maps that would project into a pixel
of the query image.
Then we sort these depth values and search for the closest depth value which 
 obtains a positive score in a voting scheme.
 This voting scheme is very similar to 
the one proposed in the previous stage, but many more depth values 
will end up with a positive score although they are incorrect.

There are 4 differences to the other voting scheme:
(1) Every depth map can vote (without accuracy restrictions),
(2) the border between consistent and contradicting vote is set to $ (1/\sqrt{u_{\text{query}}} +1/\sqrt{u_{\text{ref}}}) \cdot \sigma_\text{max} $,
(3) $\text{\emph{support}}_\text{ref} = 1$ for all measurements and
(4) a depth value has to obtain at least three votes to be considered valid.
If no such depth value is found, the original depth is kept.

Using the augmented depth map, we now treat a depth value as a negative sample if the following two
criteria are met.
First, the query depth value has to be smaller than the depth value of the augmented depth map.
Second, the difference between those two depth values has to be larger than 
$ \sigma_\text{max} \cdot 1/\sqrt{u_{\text{augmented}}}$, where
$u_{\text{augmented}}$ stands for the largest Eigen value of the covariance matrix of the augmented measurement if 
we pretend that it is only visible from the query stereo pair.
The final training data is then a combination of the negative samples from this stage with
the positive and negative training samples from the previous stage.

\section{Experiments}

In our experiments, we use three 
publicly available datasets, which are namely the KITTI2012 dataset~\cite{geiger12},
the Middlebury2014 dataset~\cite{scharstein14}, and the Strecha fountain dataset~\cite{strecha08dataset}.

The main focus of our experiments is on the KITTI2012 dataset~\cite{geiger12} because
it is well-suited to demonstrate our approach and has already been used before for the evaluation of
confidence prediction algorithms~\cite{haeusler13,park15}.
The KITTI2012 dataset does not only let us evaluate the coverage and accuracy of our approach,
but also lets us highlight the usefulness of our approach 
in boosting the performance of confidence prediction approaches
by simply training them on the automatically generated training data.

\subsection{General Setup}

For all experiments we used the same set of parameters.
The parameter $\alpha_{\text{min}}$ ($=10^\circ$) can be used
to adjust the trade-off between coverage and label error.
As a general rule, we can say that if one increases this parameter, 
the false positive rate becomes lower, but at the same time the label coverage decreases as well.
The parameter $\sigma_\text{max}$ ($=2$) can be used to express desired
accuracy of a query algorithm as a multiple of the $\sigma$ bound.

As query algorithms, we use two different stereo algorithms.
The first algorithm is a Semi-Global Matching (SGM)~\cite{hirschmuller08} implementation by
Rothermel et al.~\cite{rothermel12} which uses the census transform for computing the matching cost.
As a second algorithm we chose the recently proposed Slanted Plane Smoothing (SPS) approach of Yamaguchi et al.~\cite{yamaguchi14}.
We chose this approach because it shows a very good performance on the KITTI datasets~\cite{geiger12,menze15},
and gives a completely different output than a SGM (piece-wise planar super pixels vs. unrestricted transitions).

For analyzing the benefit of our approach for learning, we have chosen three
different recent approaches~\cite{haeusler13,spyro14,park15} which are based on confidence prediction.
All three approaches use random forests for the confidence prediction, which made it possible
to reimplement them in a common framework.
The difference between the approaches lies in which hand-crafted features they feed to the random forest.
Ensemble learning~\cite{haeusler13} uses the peak ratio, entropy of disparities, perturbation, left-right disparity difference,
horizontal gradient, disparity map variance, disparity ambiguity, zero mean sum of absolute differences and the local SGM energy,
which results through consideration of multiple scales in a feature vector of 23 dimensions.
Ground Control Point (GCP) learning~\cite{spyro14} uses eight features, which are 
the matching cost, distance to border, maximum margin, attainable maximum likelihood,
left-right consistency, left-right difference, distance to discontinuity and difference 
with median disparity.
Park et al.~\cite{park15} use a feature vector with 22 dimensions, which contains
the peak ratio, naive peak ratio, matching score, maximum margin, winner margin,
maximum likelihood, perturbation, negative entropy, left-right difference, local curvatures,
local variance of disparity values,
distance to discontinuity, median deviations of disparities, left-right consistency,
magnitude of image gradient and the distance to border.

For the implementation we used the publicly available random forest framework of Schulter at al.~\cite{schulter14a}.
For training the forest we used the same settings in all our experiments.
We used 20 trees with a maximum depth of 20 and a minimum leaf size of 100.
For choosing a split function we use the standard entropy and draw 2000 random samples per node 
and 500 random thresholds per feature channel.
For every training setup we balanced the dataset on image basis.
This means that every image contributed as many positive training examples 
as negative examples.
For the final evaluation, we always considered the complete image.
For obtaining the pose estimation on the KITTI2012 dataset we use~\cite{geiger11}.

\subsection{KITTI Dataset}

\begin{figure}[top]
% 
%      \vspace{-15pt}
  \centering
 \subfigure[SGM Sequence 102.]
    {
                \includegraphics[width=0.30\textwidth]{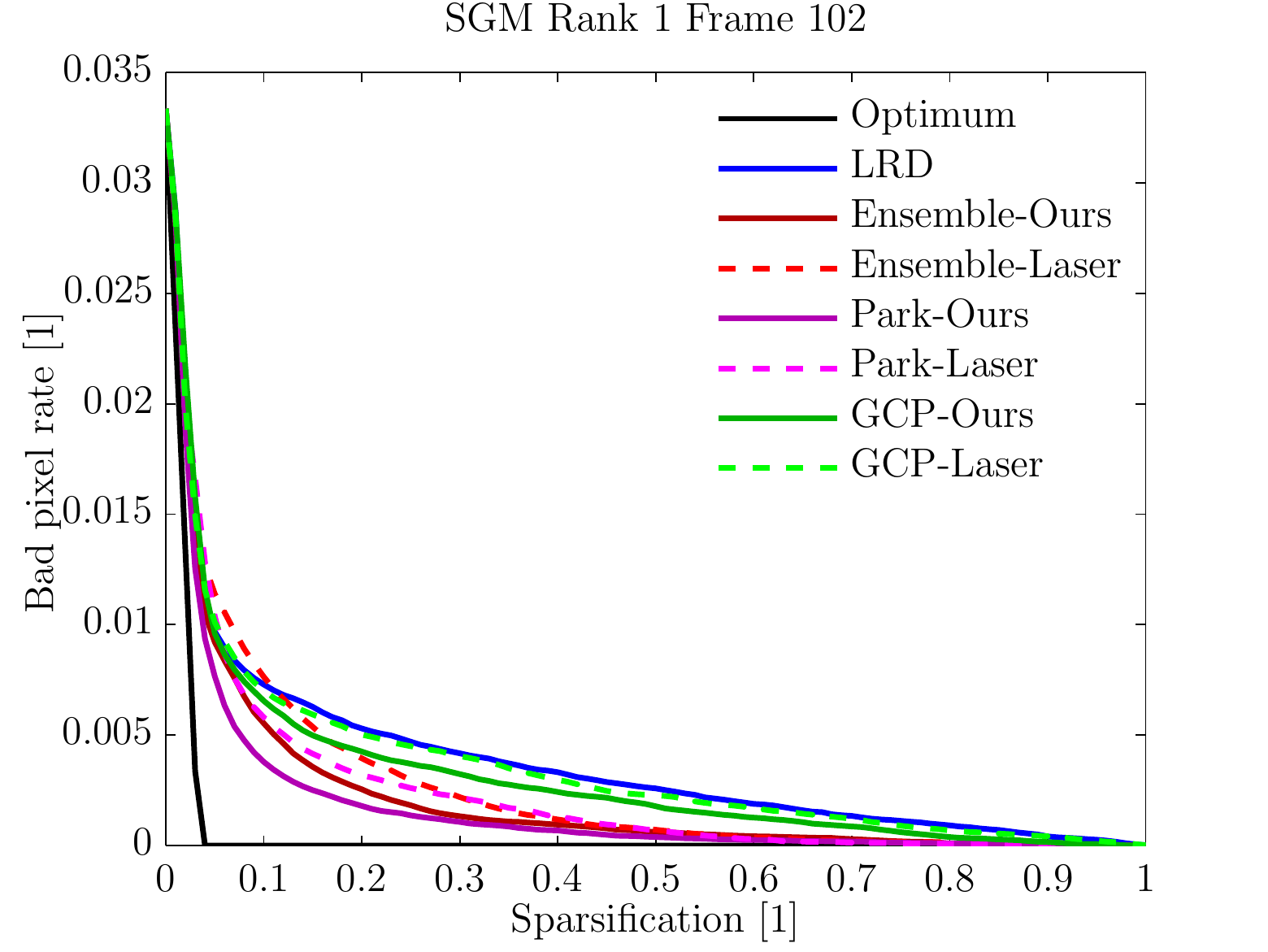} 
}\quad
 \subfigure[SPS Sequence 102.]
    {
                \includegraphics[width=0.30\textwidth]{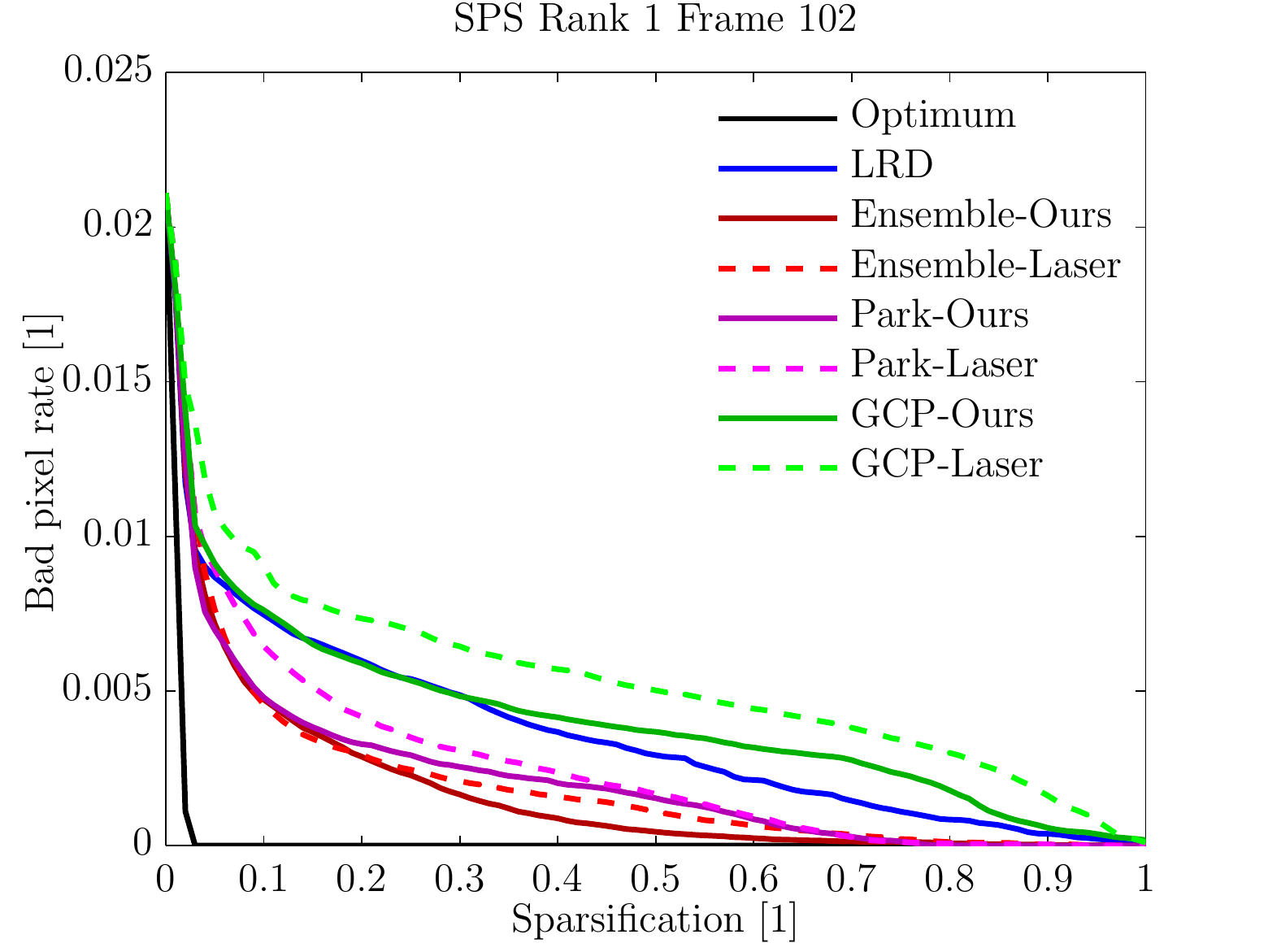} 
}

%\end{minipage}
    \caption{Sparsification curves for sequence 102 of the \textbf{KITTI} training dataset. We display all combinations of 
    query  algorithm (SGM~\cite{rothermel12} and SPS~\cite{yamaguchi14}), confidence prediction algorithm (Ensemble~\cite{haeusler13}, GCP~\cite{spyro14}, Park~\cite{park15}) and
    training data (Laser and Ours). As a baseline method we also show the Left-Right disparity Difference (LRD).}
 \vspace{-15pt}
  \label{fig:slines}
\end{figure}

We use the KITTI2012 dataset~\cite{geiger12} to evaluate 
three properties of our ground truth generation,
which are namely accuracy, coverage and training performance.
The first two, we obtain by comparing our automatically generated 
label images to label images produced with the laser ground truth provided for the training dataset.
For the SGM~\cite{rothermel12} data we reach an \textbf{accuracy of  97.3}\% (STD: 1.4\%) at an average coverage of the laser ground truth
of  47.8\% (STD: 11.8\%).
For the SPS~\cite{yamaguchi14} data we obtain an \textbf{accuracy of 95.3\%} (STD: 5.7\%) at an average coverage of  48.6\% (STD: 13.4\%).
Note that the coverage mostly depends on the camera motion.
The ideal case to demonstrate our approach would be a circular motion around an object,
whereas no motion will result in no labeled images.
As the KITTI dataset contains some sequences with very little motion, this results 
in a high standard deviation of the coverage.
\begin{figure*}[top]
% 
%     \vspace{-15pt}
  \centering
\includegraphics[width=1\textwidth]{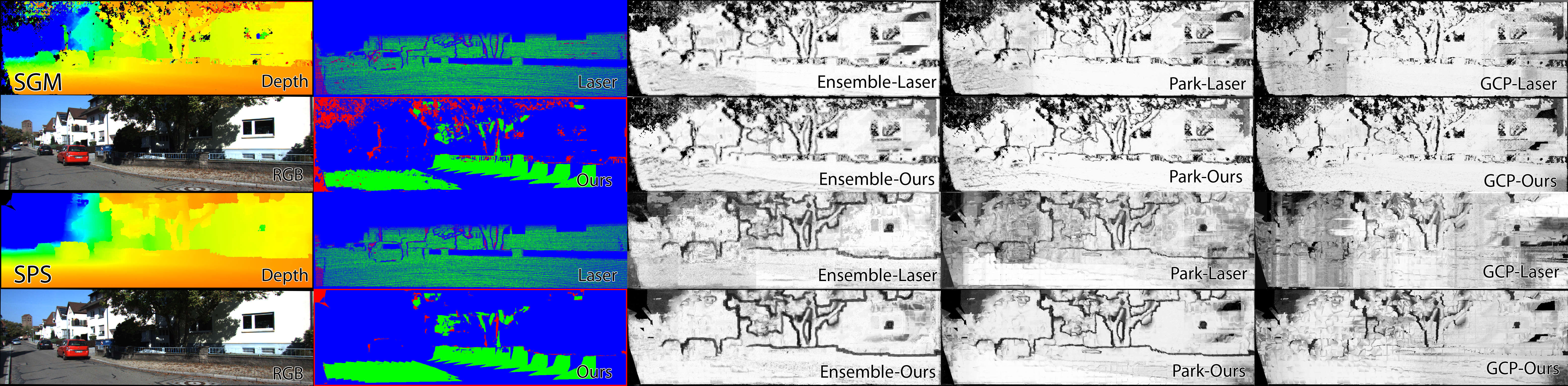} 
%\end{minipage}
    \caption{Qualitative results for sequence 102 of the \textbf{KITTI} training dataset. In the first column we show the depth maps of SGM~\cite{rothermel12} and
    SPS~\cite{yamaguchi14} together with the RGB input image. The second column shows the resulting label images once produced with the laser ground truth (Laser) and
    once with our approach (Ours). Note that our approach only assigns a positive label to parts of the scene that are observed under significantly different view points (the car is making a turn to the left in the sequence).
    The remaining 3 columns show the confidence prediction output of Ensemble~\cite{haeusler13}, GCP~\cite{spyro14} and Park~\cite{park15} once trained on Laser and once on Ours.
    The confidence ranges from low (black) to high (white).
    Note the confidence prediction is much smoother for Ours and contains less artifacts (especially for GCP).
    }
 \vspace{-15pt}
  \label{fig:example_labels_kitti}
\end{figure*}

While accuracy and coverage are relevant, the much more interesting
factor is how well the data is suited for training an algorithm.
To analyze this factor, we benchmark the change of the confidence prediction performance of three recent confidence prediction approaches,
which we further refer to as Ensemble~\cite{haeusler13}, GCP~\cite{spyro14} and Park~\cite{park15}.
For benchmarking this performance we evaluate the Area Under the Sparsification Curve (AUSC) as in ~\cite{hu12,haeusler13,park15}.
A sparsification curve plots the bad pixel rate over the sparsification factor.
For drawing the curve the pixels are sorted by confidence values and 
always the lowest values are removed.
The AUSC is a very good indicator for the prediction performance of a confidence measure.
Sparsification curves for frame 102 of the dataset are shown in Fig.~\ref{fig:slines},
while further sparsification curves can be found in the supplementary material.

For training on the laser ground truth, we follow the evaluation protocol of~\cite{haeusler13,park15}.
This means that we select the frames 43, 71, 82, 87, 94, 120, 122 and 180 of the KITTI \emph{training} dataset 
for training. The labels correct/incorrect are set by comparing the query depth maps with the laser
ground truth using the standard three pixel disparity threshold. Further on, we will mark a confidence measure trained on this data with the suffix "Laser".
As our approach requires multiple images that view the same scene,
we use the 195 sequences of 21 stereo pairs of the KITTI \emph{testing} dataset
for automatically generating our label images. Further on, we will mark a confidence measure trained on this data with the suffix "Ours".
Example label images can be found in Fig.~\ref{fig:example_labels_kitti} and the supplementary material.
For testing we once again follow the protocol of~\cite{haeusler13,park15} and evaluate the 
confidence prediction on the KITTI \emph{training} dataset minus the eight sequences that were used for training on the laser ground truth.
Thus, there is no overlap between training and testing for Laser as well as Ours.
Also note that Ours has not seen a single ground truth laser scan.
In training, we used all available training samples from the laser ground truth and roughly ten times
this number from our automatically generated data.
Note that this is less than one percent of all available training data.
With this setup our implementation used $\sim$20GB of memory for training.

In Fig.~\ref{fig:kitti_ausc} we show the mean, minimum and maximum AUSC values of the three confidence prediction algorithms for all combinations of query algorithm and
training data.
In Tab.~\ref{tab:relative_areas_kitti} we show the AUSC for each approach divided by the optimal AUSC
over all evaluated sequences of the KITTI dataset.
In all cases, using our training data resulted in a performance boost.
In some cases the AUSC even dropped by 10\%.
A visual comparison of the difference in the confidence prediction can be found 
in Fig.~\ref{fig:example_labels_kitti} and the supplementary material.
Note that our training data leads to a smoother confidence prediction with significantly fewer artifacts.

\begin{figure}[t]
  \centering
    \subfigure
    {
    \hspace{-40pt}
\includegraphics[width=0.62\textwidth]{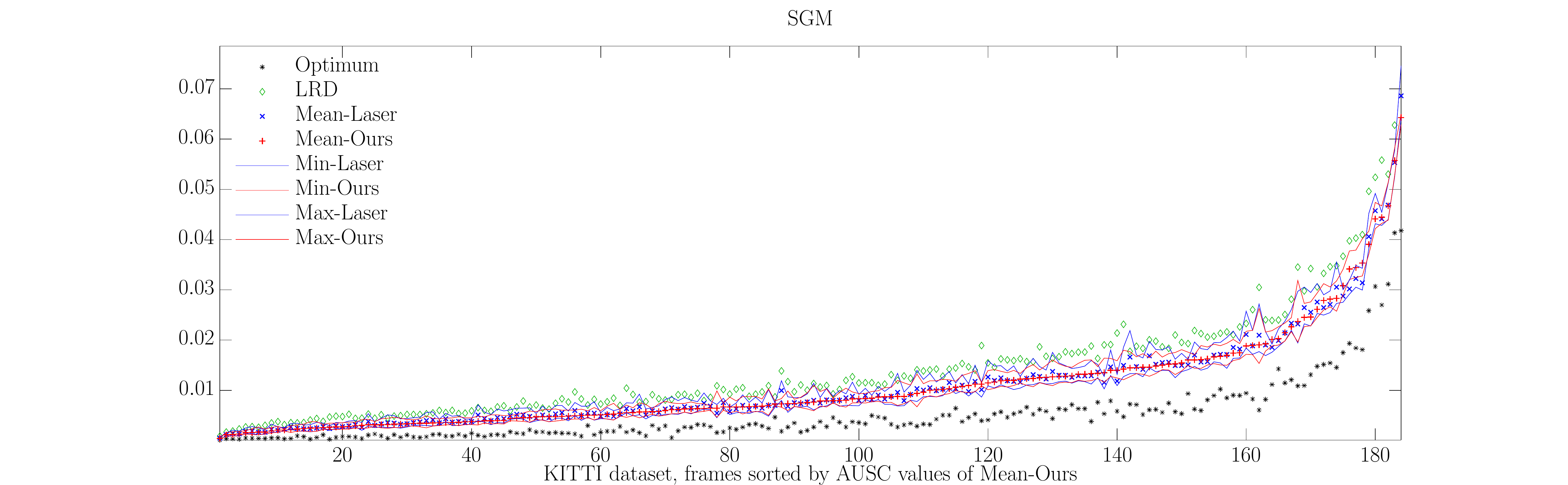} 
} 
\quad
\subfigure
    {
\hspace{-40pt}
\includegraphics[width=0.62\textwidth]{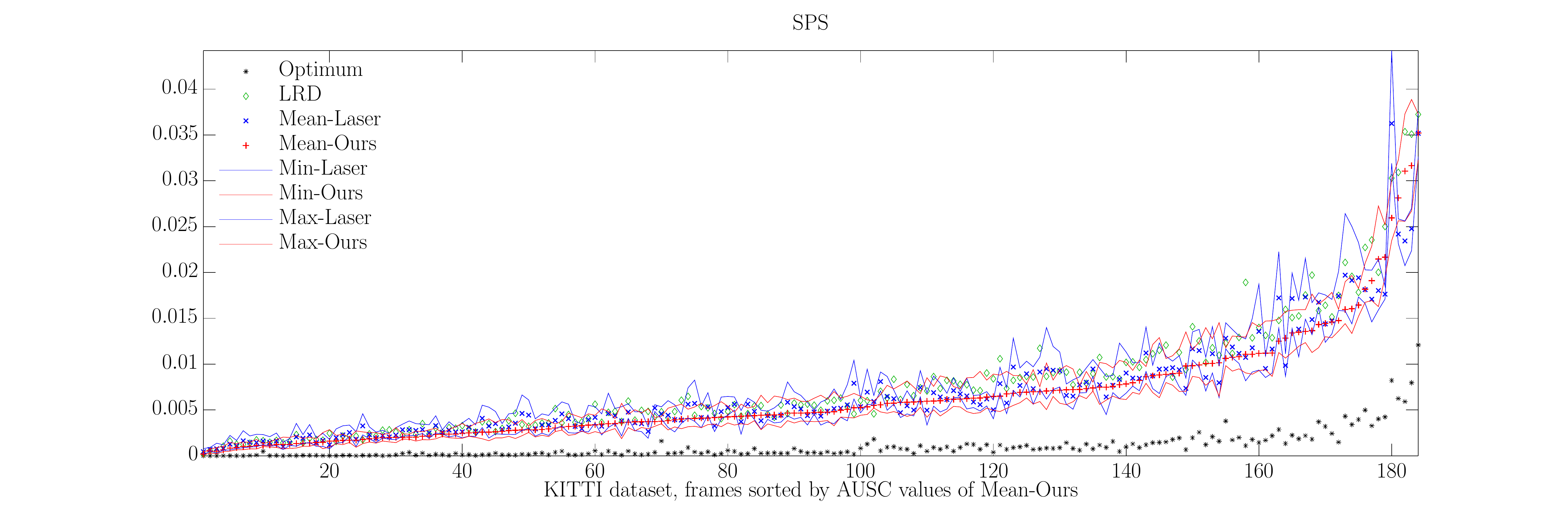} 
    }
  \caption{Mean, minimum and maximum AUSC values over the three confidence prediction algorithms (Ensemble~\cite{haeusler13}, GCP~\cite{spyro14}, Park~\cite{park15})
  for all frames of the \textbf{KITTI} training dataset minus the eight frames used for training.  We display all combinations of 
    query  algorithm (SGM~\cite{rothermel12} and SPS~\cite{yamaguchi14}) and
    training data (Laser and Ours).
    The frames were sorted according to mean AUSC value of Ours.
   As a baseline method we also show the Left-Right disparity Difference (LRD).
    Note that Ours (red) is lower than Laser (blue) in most cases.
    For SGM, all approaches  perform always better than LRD if they are trained on Ours, while if they are trained on Laser they sometimes perform worse (e.g. 142).
    For SPS stereo, the number of severe errors is significantly higher for Laser than for Ours (compare blue versus red peaks above 160). }
  \vspace{-5pt}
  \label{fig:kitti_ausc}
\end{figure}

As a matter of completeness, we executed our training data generation only 
on the eight same sequences that were used for training Laser.
One has to note that the coverage of our approach depends on the camera motion
and one of the sequences (180) contains no useful motion, which leaves our approach with 7 sequences.
Using only this limited amount of training data, the AUSC increased by $\sim$10\% for all approaches
compared to using the 195 testing sequences.
This is not surprising, as each of our training images can be considered as weaker compared to
the laser ground truth, in the sense that consistency alone cannot uncover all errors and 
that the coverage of our labeling depends on the camera motion.
But this experiment clearly shows that using 
ten times more "weak" training samples, which can be cheaply generated with our method, still
leads to a better performance than fewer "strong" training samples.

\begin{table}
\centering
 \begin{tabular}[b]{|c|c|c|c|c|}
  \hline
   & LRD  & Ens.\cite{haeusler13}  & Park\cite{park15} & GCP\cite{spyro14}\\\hline\hline 
   SGM-Laser & 2.81 & 1.97&  1.93&  2.50 \\\hline 
   SGM-Ours &  2.81 &  1.95&  \textbf{1.92} & 2.45\\\hline
   Reduction & - & 0.94\% & 0.78\% & \textbf{2.02\%}\\\hline\hline 
   SPS-Laser & 7.60& 5.86&  6.23&  8.28 \\\hline 
   SPS-Ours & 7.60 &  \textbf{5.43}& 5.61& 7.95\\\hline 
   Reduction & - &  7.28\% & \textbf{9.93\%} & 3.98\% \\\hline 
\end{tabular}
\caption{Area under the sparsification curve divided by optimal area on the \textbf{KITTI} dataset.
We display all combinations of 
    query  algorithm (SGM~\cite{rothermel12} and SPS~\cite{yamaguchi14}), confidence prediction algorithm (Ensemble~\cite{haeusler13}, GCP~\cite{spyro14}, Park~\cite{park15}) and
    training data (Laser and Ours).
    The reduction is computed as $1-AUSC_{Ours}/AUSC_{Laser}$.
} \vspace{-15pt}
\label{tab:relative_areas_kitti}
\end{table}

\subsection{Middlebury Dataset}

\begin{figure}[t]
  \centering
    \subfigure
    {
\includegraphics[width=0.9\columnwidth]{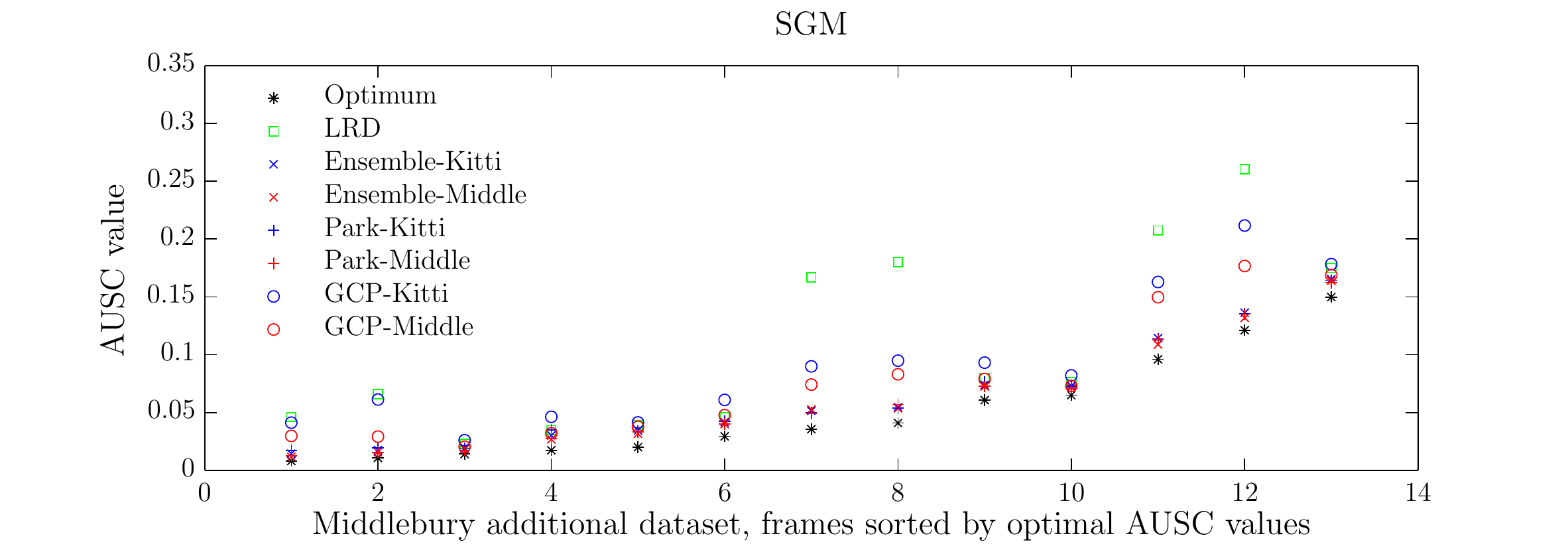} 
} \vspace{-20pt}\quad
\subfigure
    {
\includegraphics[width=0.9\columnwidth]{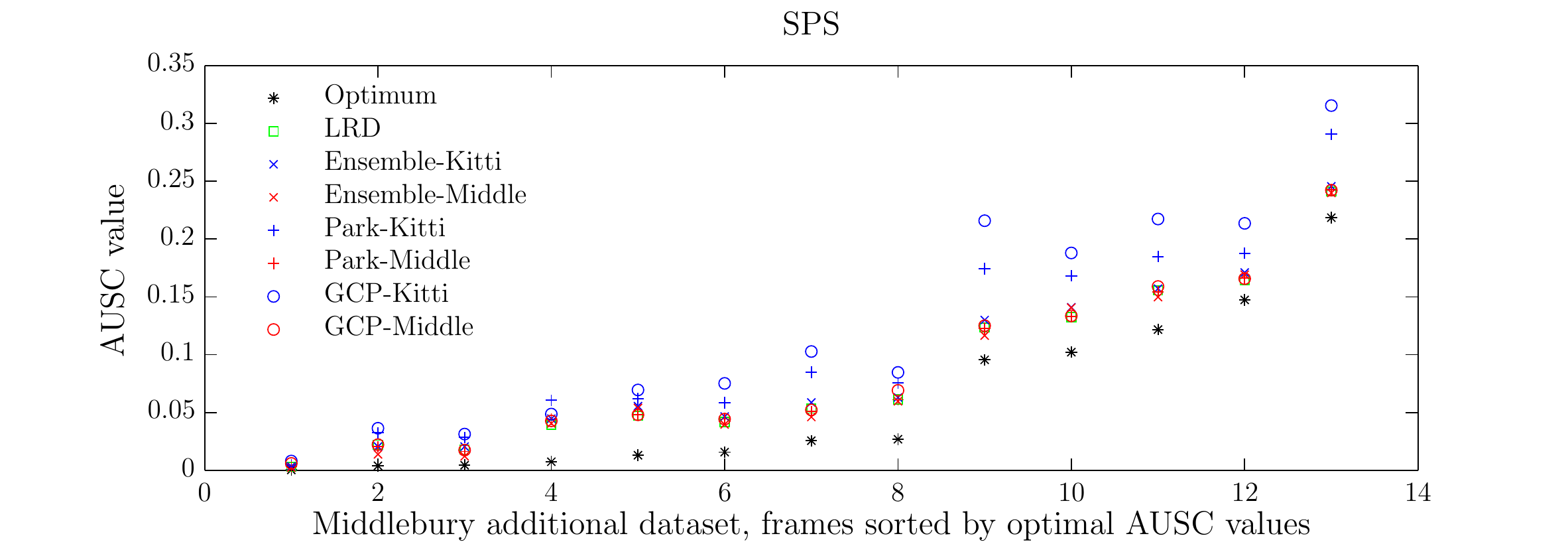} 
    }
  \caption{Area under the Sparsification Curve (AUSC) values for all 13 frames of the additional \textbf{Middlebury} dataset. The frames were sorted according to the 
  optimal area under the curve value. We display all combinations of 
    query  algorithm (SGM~\cite{rothermel12} and SPS~\cite{yamaguchi14}), confidence prediction algorithm (Ensemble~\cite{haeusler13}, GCP~\cite{spyro14}, Park~\cite{park15}) and
    training data (Kitti~\cite{geiger12} and Middle~\cite{scharstein14}). As a baseline method we also show the Left-Right disparity Difference (LRD). Note that the red symbols (Middle)
    are in many cases drastically lower than their blue counter parts (Kitti).}
  \vspace{-7pt}
  \label{fig:middlebury_ausc}
\end{figure}

\begin{table}
\centering
 \begin{tabular}[b]{|c|c|c|c|c|}
  \hline
   & LRD  & Ens.\cite{haeusler13} & Park\cite{park15} & GCP\cite{spyro14}  \\\hline \hline 
   SGM-Kitti & 2.10  &  1.24    &  1.25    &   1.78  \\\hline 
   SGM-Middle & 2.10    & \textbf{1.19}    & 1.20   & 1.50\\\hline 
   Reduction & -   & 3.29\% & 3.30\% & \textbf{15.86\%} \\\hline \hline 
   SPS-Kitti & 1.41 &   1.48  &    1.81   &  2.05 \\\hline 
   SPS-Middle & 1.41 &     \textbf{1.39}   &   1.42   &   1.44\\\hline 
   Reduction & - &  6.32\% & 21.63\%& \textbf{29.82\%}\\\hline 
\end{tabular}
\caption{Area under the sparsification curve divided by optimal area on the \textbf{Middlebury} dataset.
We display all combinations of 
    query  algorithm (SGM~\cite{rothermel12} and SPS~\cite{yamaguchi14}), confidence prediction algorithm (Ensemble~\cite{haeusler13}, GCP~\cite{spyro14}, Park~\cite{park15}) and
    training data (Kitti~\cite{geiger12} and Middle~\cite{scharstein14}).
    The reduction is computed as $1-AUSC_{Middle}/AUSC_{Kitti}$.}
     \vspace{-15pt}
\label{tab:relative_areas}
\end{table}

The Middlebury2014~\cite{scharstein14} dataset contains a set of 23 high resolution stereo pairs for which known camera calibration parameters and
ground truth disparity maps obtained with a structured light scanner are available. The set is divided into 10 stereo pairs for training and additional 13 stereo pairs that we used for testing.
The images in the Middlebury dataset all show static indoor scenes with varying difficulties including repetitive structures, occlusions, wiry objects as well as untextured areas.

Due to the limitation that only stereo pairs and no multi-view sequences are provided, we are not able to evaluate the accuracy performance of our ground truth generation.
But we can still evaluate the performance of the confidence measures previously learned on the KITTI to evaluate their generalization performance from outdoor to indoor scenes.
Figure~\ref{fig:middlebury_ausc} shows the resulting AUSC curve for SGM~\cite{rothermel12}  and SPS~\cite{yamaguchi14}, respectively. 
In Tab.~\ref{tab:relative_areas} we show the AUSC over the optimal values.

For all combinations of query algorithm and confidence prediction approach,
training on the Middlebury increased the performance compared 
to training on the KITTI and evaluating on the Middlebury.
The percentage of area reduction strongly depends on the used confidence prediction approach.
We assume that the large variation in area reduction (3\%-30\%) is caused 
by features which are very setup specific (e.g. distance to border).
Despite the large reduction variation, all approaches benefit from
training on the Middlebury rather than the KITTI.
This means that tuning towards a special setup can make a large difference in performance.

\subsection{Strecha Dataset}

\begin{figure}[top]
  \centering
    \subfigure[SGM Pair 2+3.]
    {

                \includegraphics[width=0.30\textwidth]{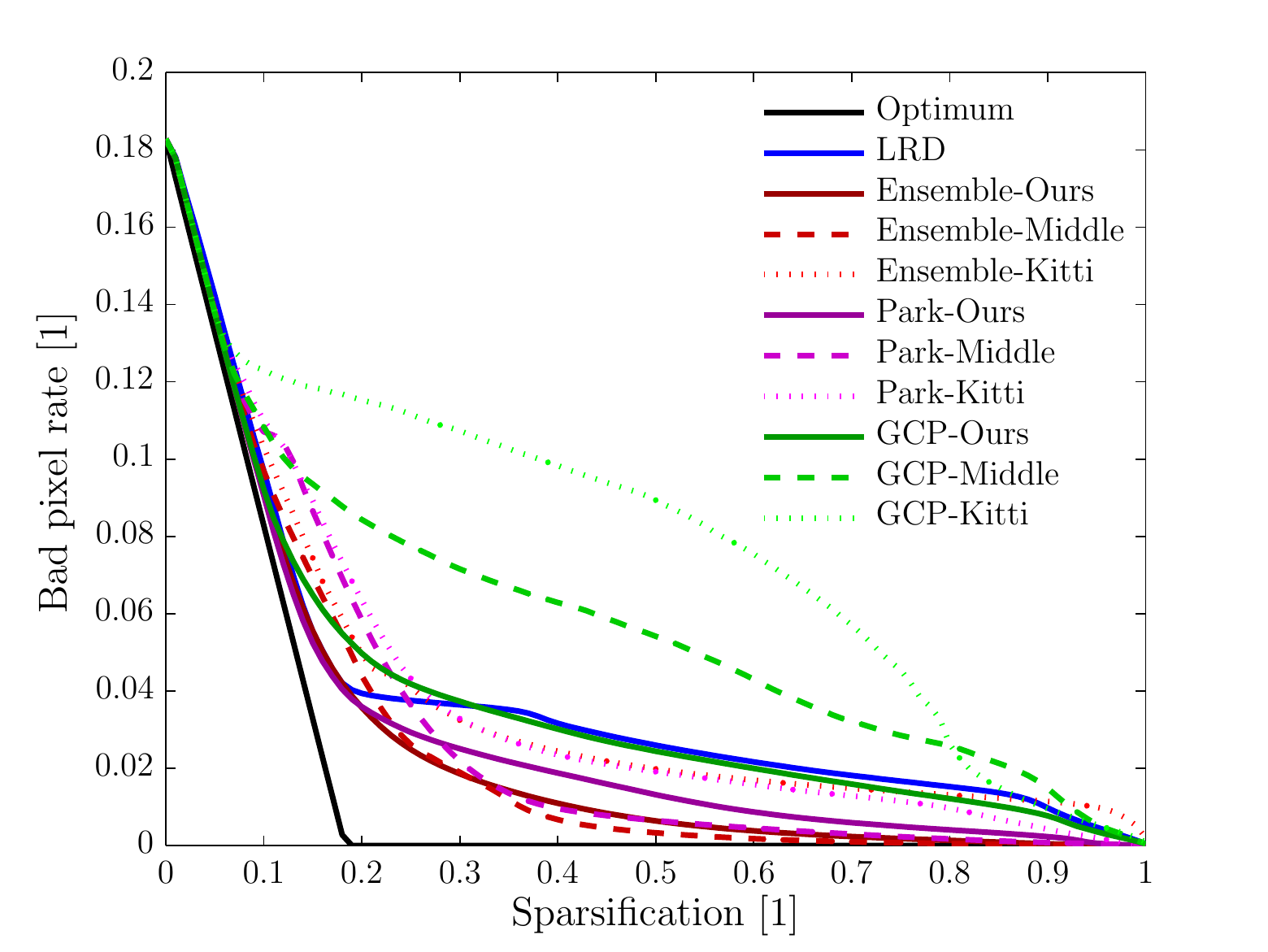} 
}
\vspace{-5pt} \quad
\subfigure[SGM Pair 6+7]
    {

                \includegraphics[width=0.30\textwidth]{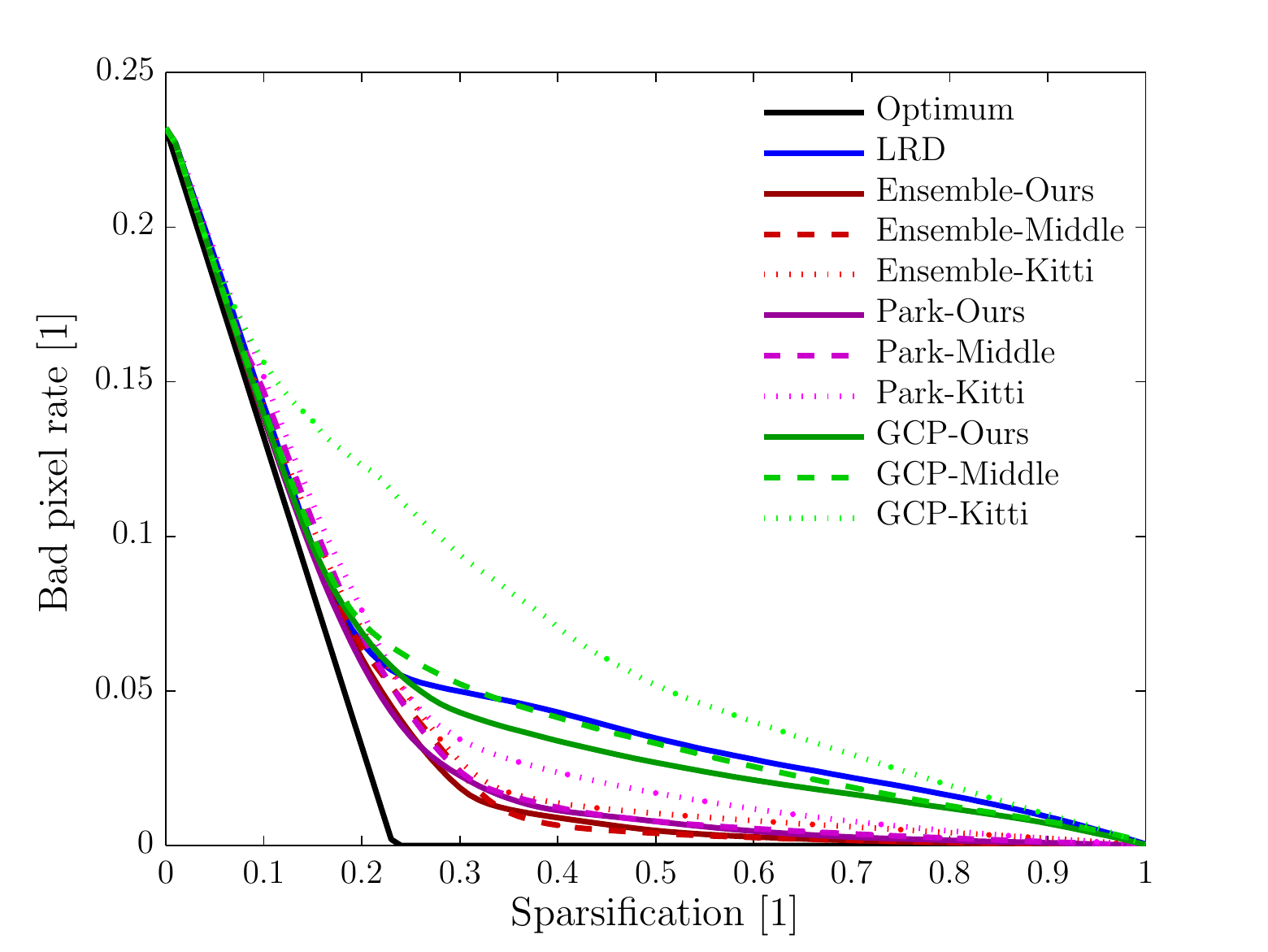} 
}
    \caption{Sparsification curves for testing stereo pair on the \textbf{Strecha} fountain dataset. 
    We display all combinations of confidence prediction algorithm (Ensemble~\cite{haeusler13}, GCP~\cite{spyro14}, Park~\cite{park15}) and
    training data (Kitti~\cite{geiger12}, Middle~\cite{scharstein14} and Ours) for the SGM output~\cite{rothermel12}. 
    As a baseline method we also show the Left-Right disparity Difference (LRD).}
    \vspace{-15pt}
  \label{fig:strecha_slines}
\end{figure}

To further demonstrate the value of our approach,
we analyze the sparsification performance in a completely different setup.
For this experiment we used the multi-view stereo dataset of Strecha et al.~\cite{strecha08dataset}.
This dataset provides images together with camera poses and two ground truth meshes.
From the two available meshes, the Herz-Jesu mesh is a good example that 
also active sensors have their limitations. In this mesh all the thin structures (hand rails and bars) are 
simply missing. 
As these errors would cause problems in the evaluation,
we only used the second dataset (Fountain), which does not contain any thin structures.
This dataset consists of 11 images aligned to the ground truth mesh.
For this experiment we split the images into a training set containing 3 image pairs and 
a test set with 2 image pairs. The training pairs are made of images 0+1, 4+5 and 8+9 and
the testing pairs of 2+3 and 6+7.
Each pair was then rectified using~\cite{rothermel12}.
As the SPS implementation~\cite{yamaguchi14} failed to produce any reasonable output on this 
kind of data, we limit this experiment to the SGM~\cite{rothermel12} reconstruction.

In this setup our ground truth generation reached an \nohyphens{\textbf{accuracy of 95.1\%}} (STD: 2.6\%) at a coverage 30.4\% (STD: 5.0\%).
In Fig.~\ref{fig:strecha_slines} we show the resulting two sparsification curves and the AUSC reduction statistics in Tab.~\ref{tab:relative_area_reduction}.
All combinations of query algorithms and confidence prediction approaches performed better trained on the Middlebury than
on the KITTI.
In all cases the performance was further increased by tuning them specifically to this scene in using our automatically
generated training data.

\begin{table}
\centering
 \begin{tabular}[b]{|c|c|c|c|c|}
  \hline
   & LRD & Ens.\cite{haeusler13} & Park\cite{park15} & GCP\cite{spyro14} \\\hline 
   Kitti RA & 2.12   & 1.81   &   1.91   &    3.54   \\\hline 
   Middle RA &  2.12   &   1.43   &   1.59   &   2.60   \\\hline 
   Ours RA &  2.12   &  \textbf{1.40}    &    1.51    &    2.01   \\\hline \hline 
   Kitti Red& -   & 22.34\%  &21.04\% & \textbf{45.30\%} \\\hline 
   Middle Red & - & 1.86\%  & 5.02\%& \textbf{23.53\%}\\\hline 
\end{tabular}
\caption{Area under the sparsification curve divided by optimal area (Relative Area RA) on the \textbf{Strecha} fountain dataset.
We display all combinations of confidence prediction algorithm (Ensemble~\cite{haeusler13}, GCP~\cite{spyro14}, Park~\cite{park15}) and
    training data (Kitti~\cite{geiger12}, Middle~\cite{scharstein14} and Ours) for the SGM output~\cite{rothermel12}.
    The reduction is computed as $1-AUSC_{x}/AUSC_{Ours}$ for each confidence prediction approach.}
    \vspace{-15pt}
\label{tab:relative_area_reduction}
\end{table}

\section{Conclusion}
In this paper we present a novel way to train confidence prediction
approaches for stereo vision in a cheap and scalable manner.
We collect positive and negative training data
by analyzing the consistency between depthmaps that observe the same physical scene.
Consistency is a necessary but not sufficient criterion for correctness.
On the one hand, this means that consistency is perfectly suited for unveiling incorrect depth values and thus
to collect negative training data.
On the other hand, it can never be guaranteed that all incorrect depth values are detected through consistency
alone, as they can be consistent and incorrect at the same time.
% which can potentially lead to incorrect samples in the positive training data.
To keep the number of incorrect samples in the positive training data low, 
we only consider parts of the scene which have been viewed from significantly different observation angles
for the generation of positive training data.
In our experiments, we demonstrate that the resulting training data
can be a great benefit for learning-based confidence prediction.
On the KITTI2012 dataset, the amount and diversity of our training data allowed us to
improve the average confidence prediction performance of three different approaches
by 1 to 10\%
without changing
the algorithms themselves.
Further, we demonstrated that all three confidence prediction
approaches 
can significantly benefit from learning application specific properties.
With our approach,
these specific properties can be learned at low cost;
even for applications, such as aerial or under water robotics, that typically lack ground truth data.

\FloatBarrier
{\small
\bibliographystyle{ieee}
%\bibliography{egbib}
\bibliography{\BibPath/icg_abbrevs,\BibPath/user-bibtex}
}

\clearpage
\appendix
\section{Supplementary Material}

\begin{figure*}[p]
     \vspace{-15pt}
  \centering

 \subfigure[SGM Frame 102 (details in Fig.~\ref{fig:sgm102}).]
    {

                \includegraphics[width=0.45\textwidth]{sgm_sparsification102.pdf} 
}\quad
 \subfigure[SPS Frame 102 (details in Fig.~\ref{fig:sps102}).]
    {

                \includegraphics[width=0.45\textwidth]{sps_sparsification102.pdf} 
       
}\quad 

 \subfigure[SGM Frame 56 (details in Fig.~\ref{fig:sgm056}).]
    {

                \includegraphics[width=0.45\textwidth]{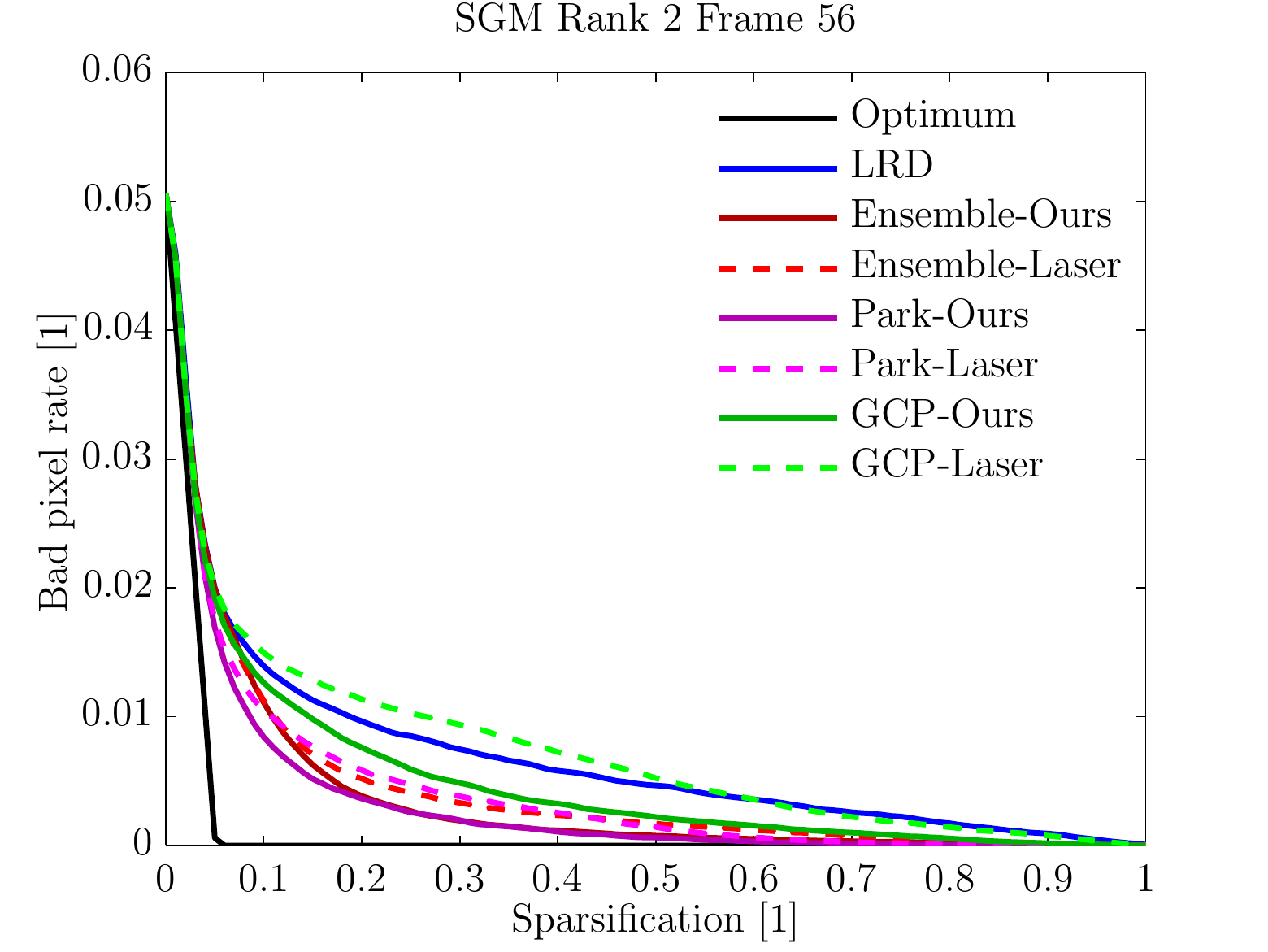} 
}\quad
 \subfigure[SPS Frame 56 (details in Fig.~\ref{fig:sps056}).]
    {

                \includegraphics[width=0.45\textwidth]{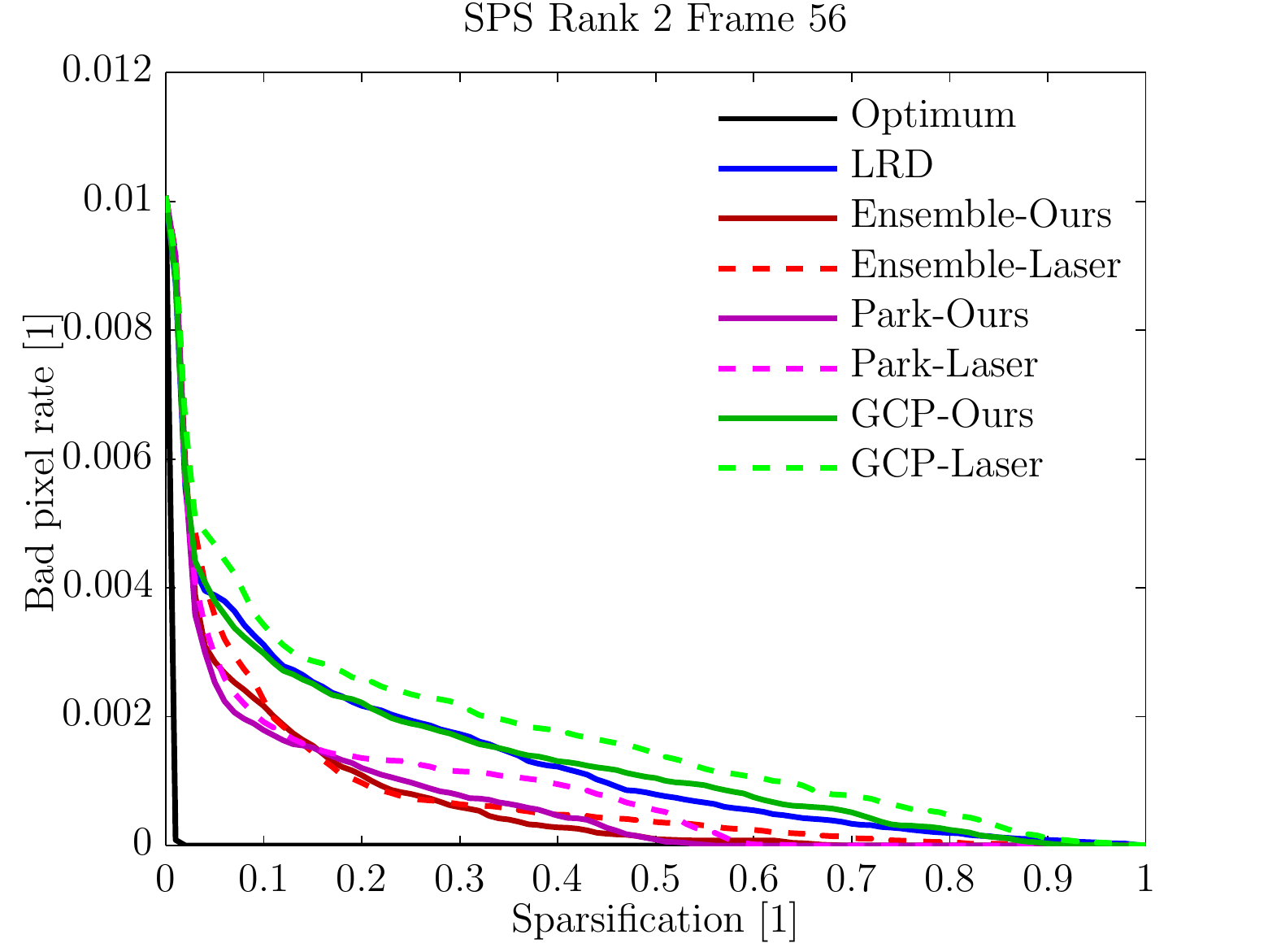} 
       
}\quad 

 \subfigure[SGM Frame 0 (details in Fig.~\ref{fig:sgm000}).]
    {

                \includegraphics[width=0.45\textwidth]{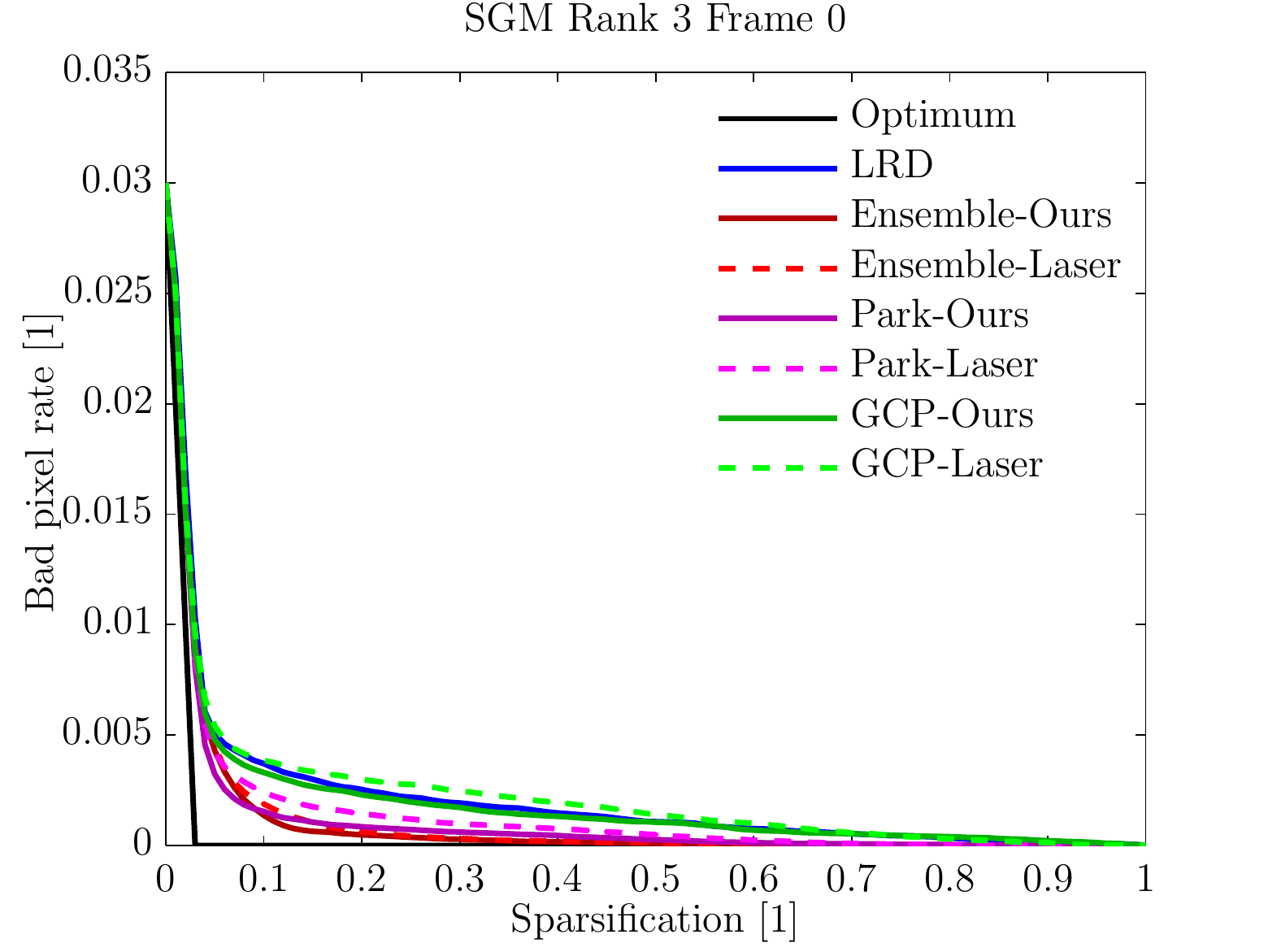} 
}\quad
 \subfigure[SPS Frame 0 (details in Fig.~\ref{fig:sps000}).]
    {

                \includegraphics[width=0.45\textwidth]{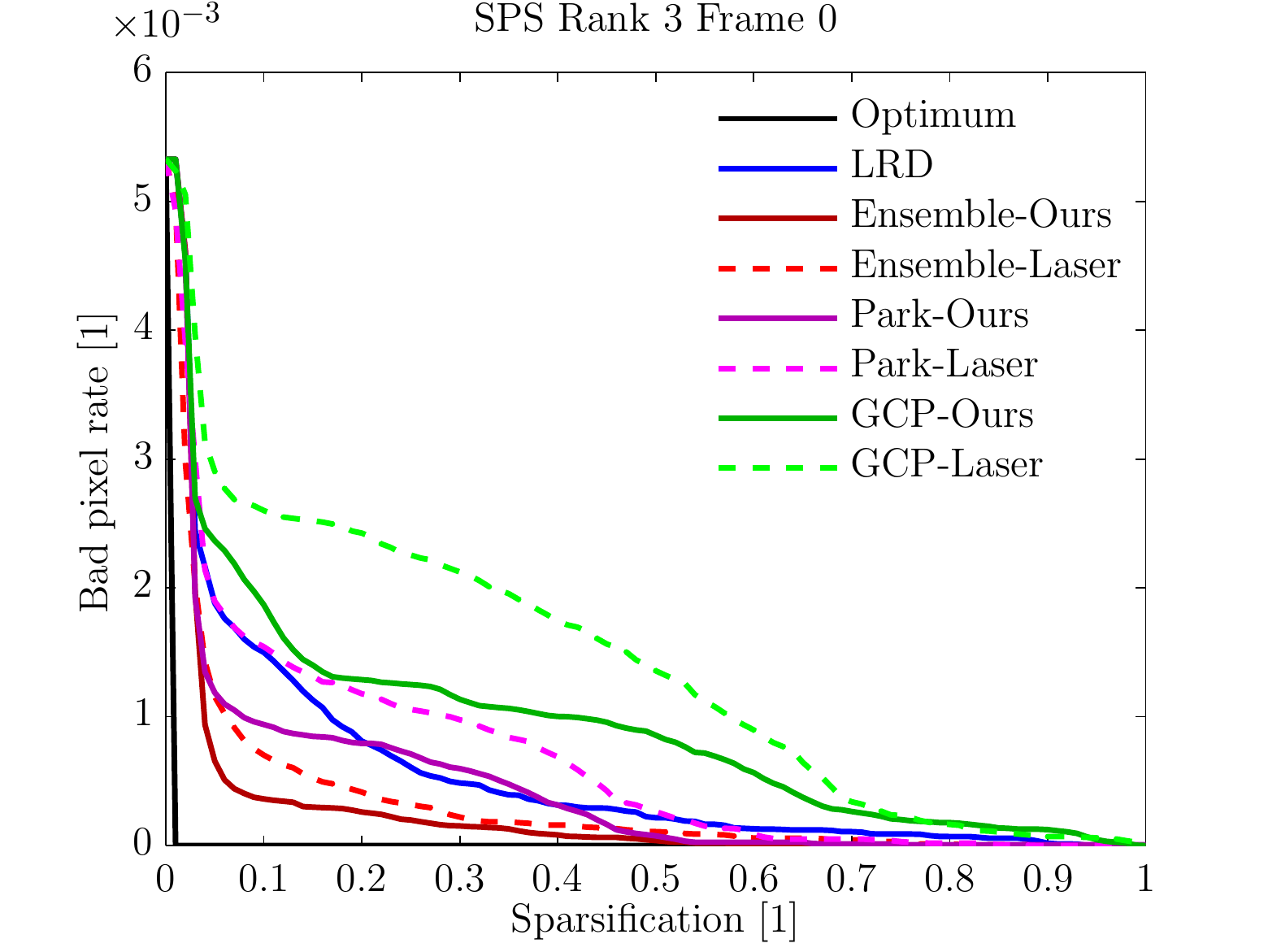} 
       
}\quad 

%\end{minipage}
    \caption{Sparsification curves for the three \textbf{best frames} with  (102,56, and 0) of the KITTI training dataset.
    The ranking was obtained by the ratio $AUSC_{SGM-Park-Ours}/AUSC_{SGM-Park-Laser}$. 
    We display all combinations of 
    query  algorithm (SGM~\cite{rothermel12} and SPS~\cite{yamaguchi14}), confidence prediction algorithm (Ensemble~\cite{haeusler13}, GCP~\cite{spyro14}, Park~\cite{park15}) and
    training data (Laser and Ours). As a baseline method we also show the Left-Right disparity Difference (LRD).}
 \vspace{-20pt}
  \label{fig:slines_lowest}
\end{figure*}

\begin{figure*}[p]
     \vspace{-15pt}
  \centering

 \subfigure[SGM Frame 85 (details in Fig.~\ref{fig:sgm085}).]
    {

                \includegraphics[width=0.45\textwidth]{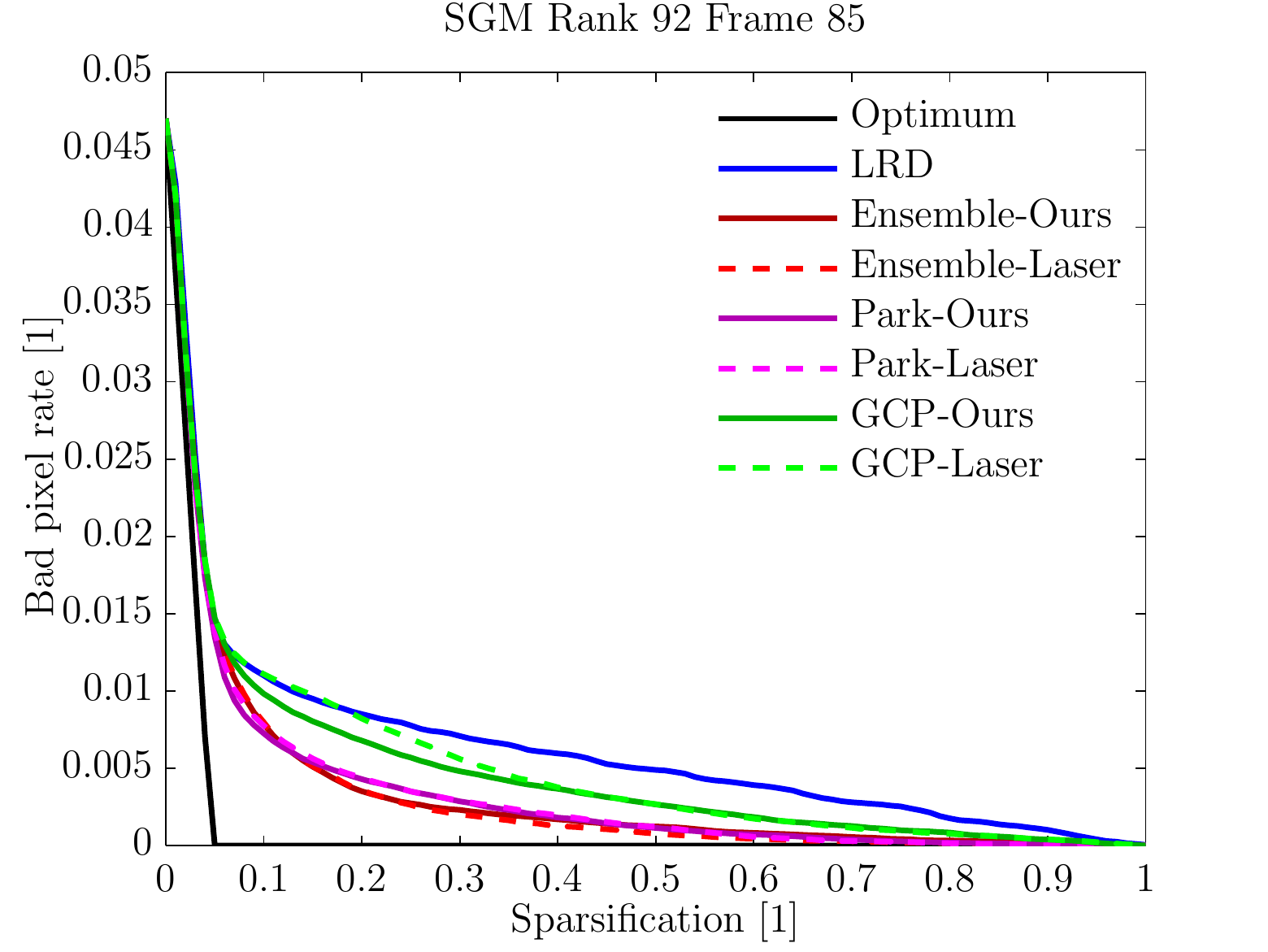} 
}\quad
 \subfigure[SPS Frame 85 (details in Fig.~\ref{fig:sps085}).]
    {

                \includegraphics[width=0.45\textwidth]{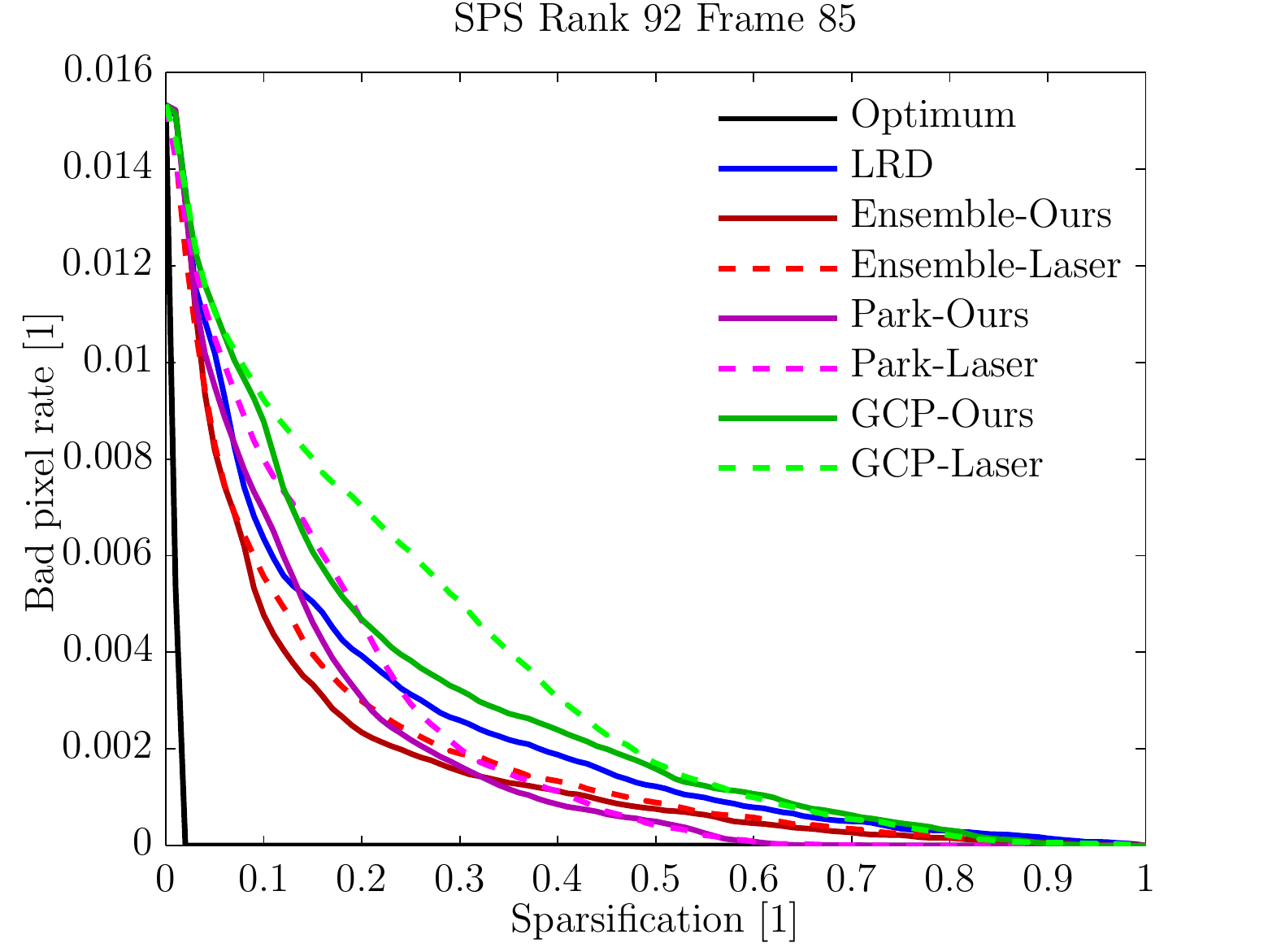} 
       
}\quad 

 \subfigure[SGM Frame 126 (details in Fig.~\ref{fig:sgm126}).]
    {

                \includegraphics[width=0.45\textwidth]{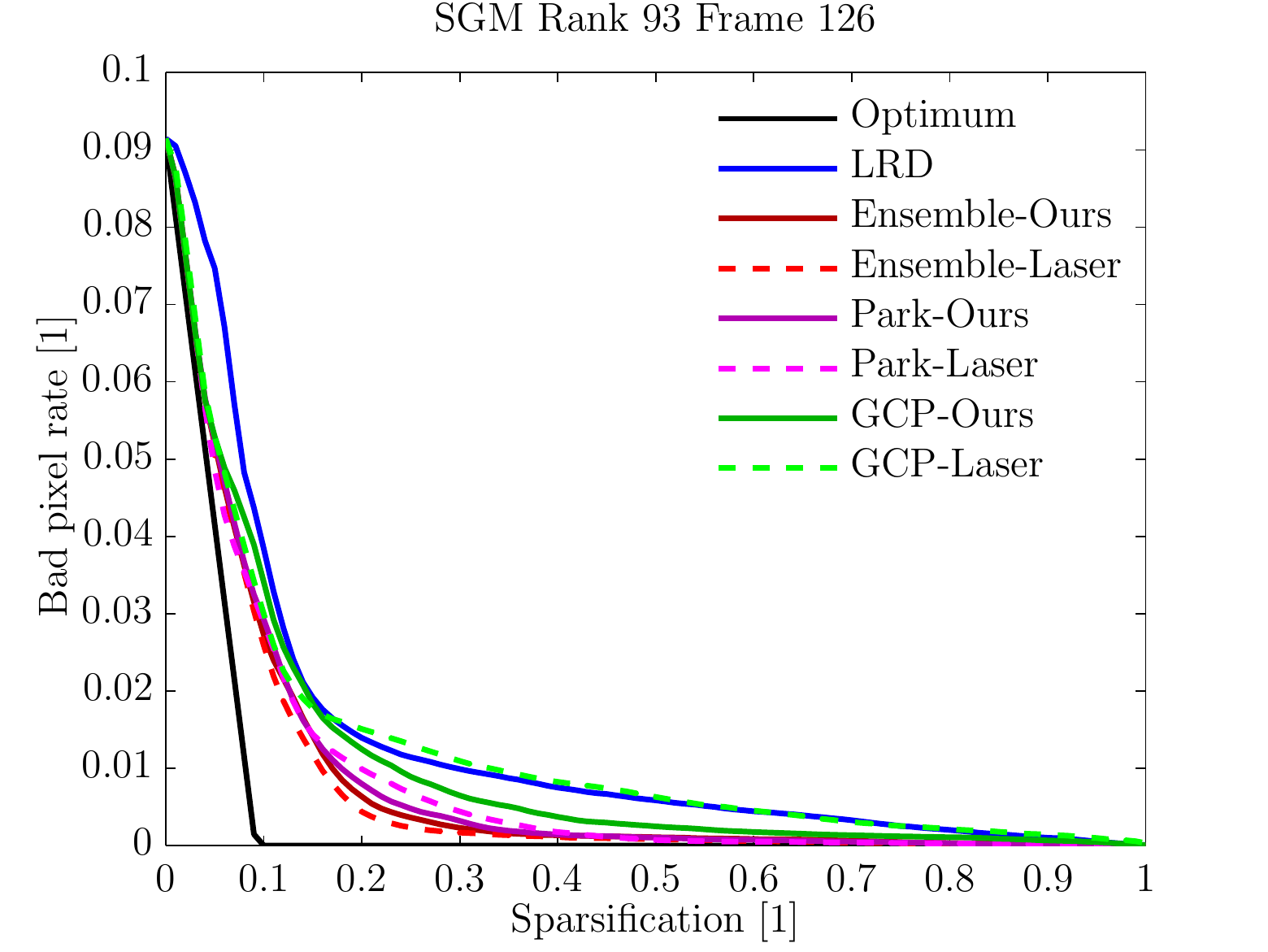} 
}\quad
 \subfigure[SPS Frame 126 (details in Fig.~\ref{fig:sps126}).]
    {

                \includegraphics[width=0.45\textwidth]{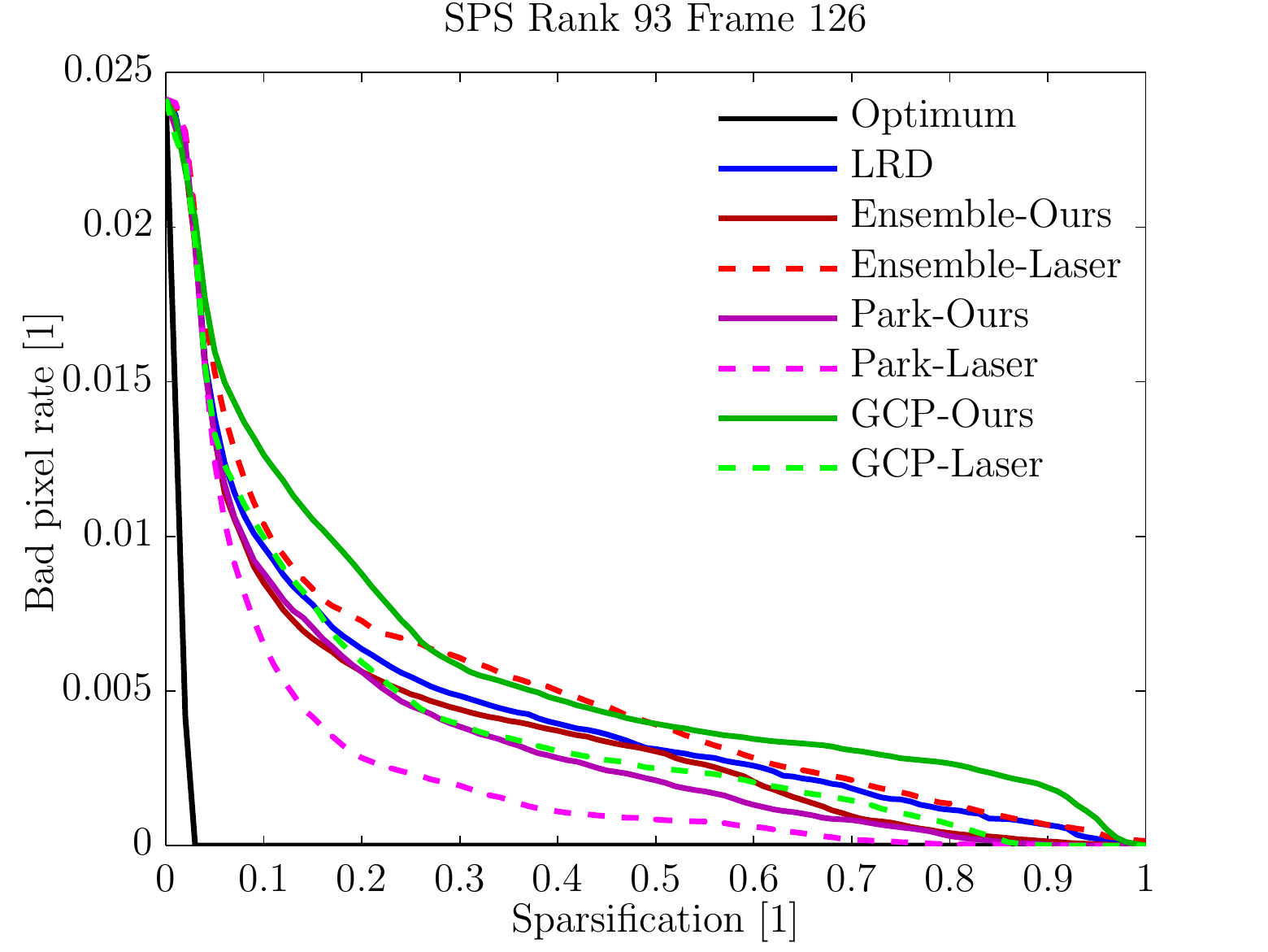} 
       
}\quad 

 \subfigure[SGM Frame 88 (details in Fig.~\ref{fig:sgm088}).]
    {

                \includegraphics[width=0.45\textwidth]{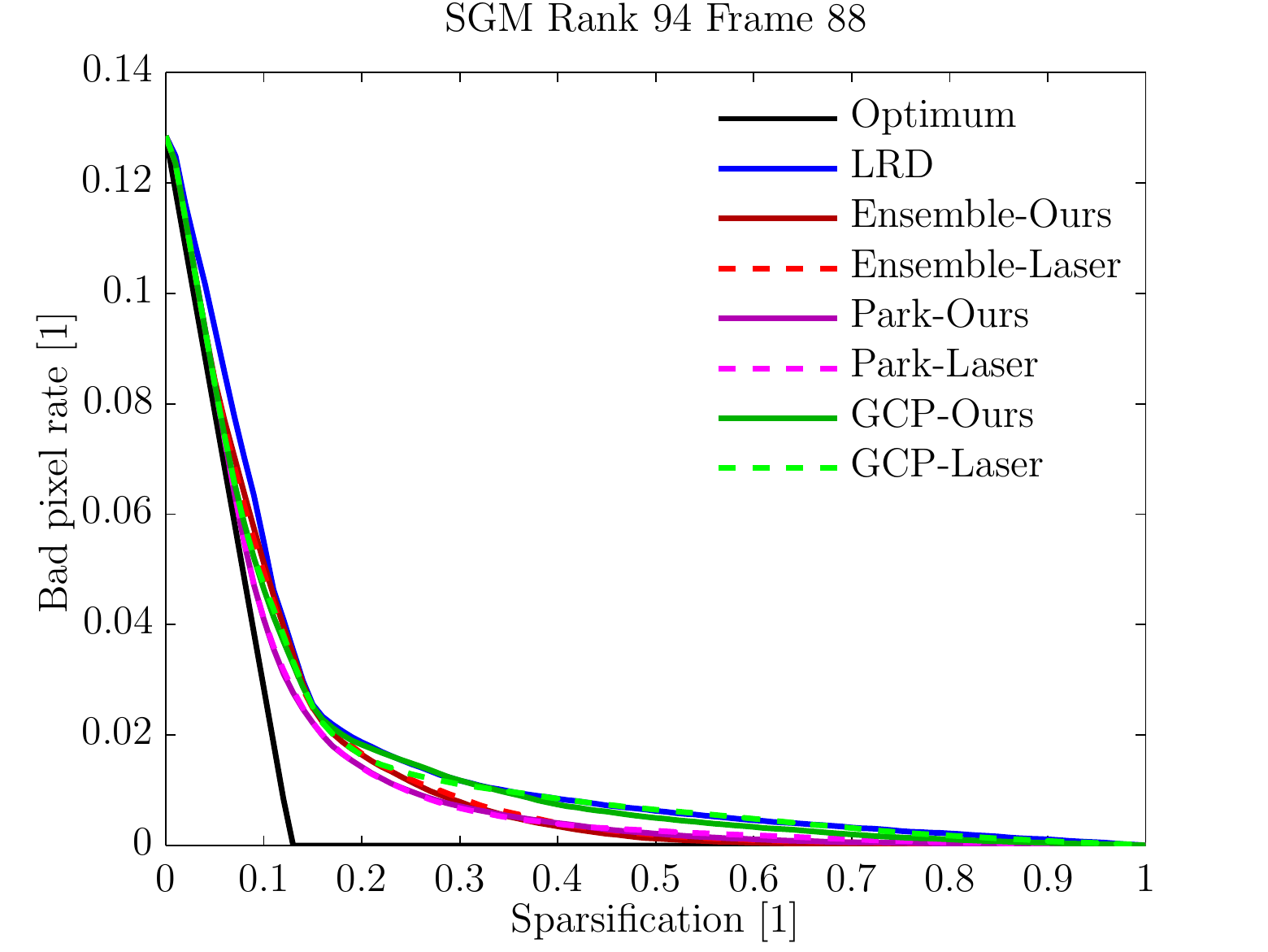} 
}\quad
 \subfigure[SPS Frame 88 (details in Fig.~\ref{fig:sps088}).]
    {

                \includegraphics[width=0.45\textwidth]{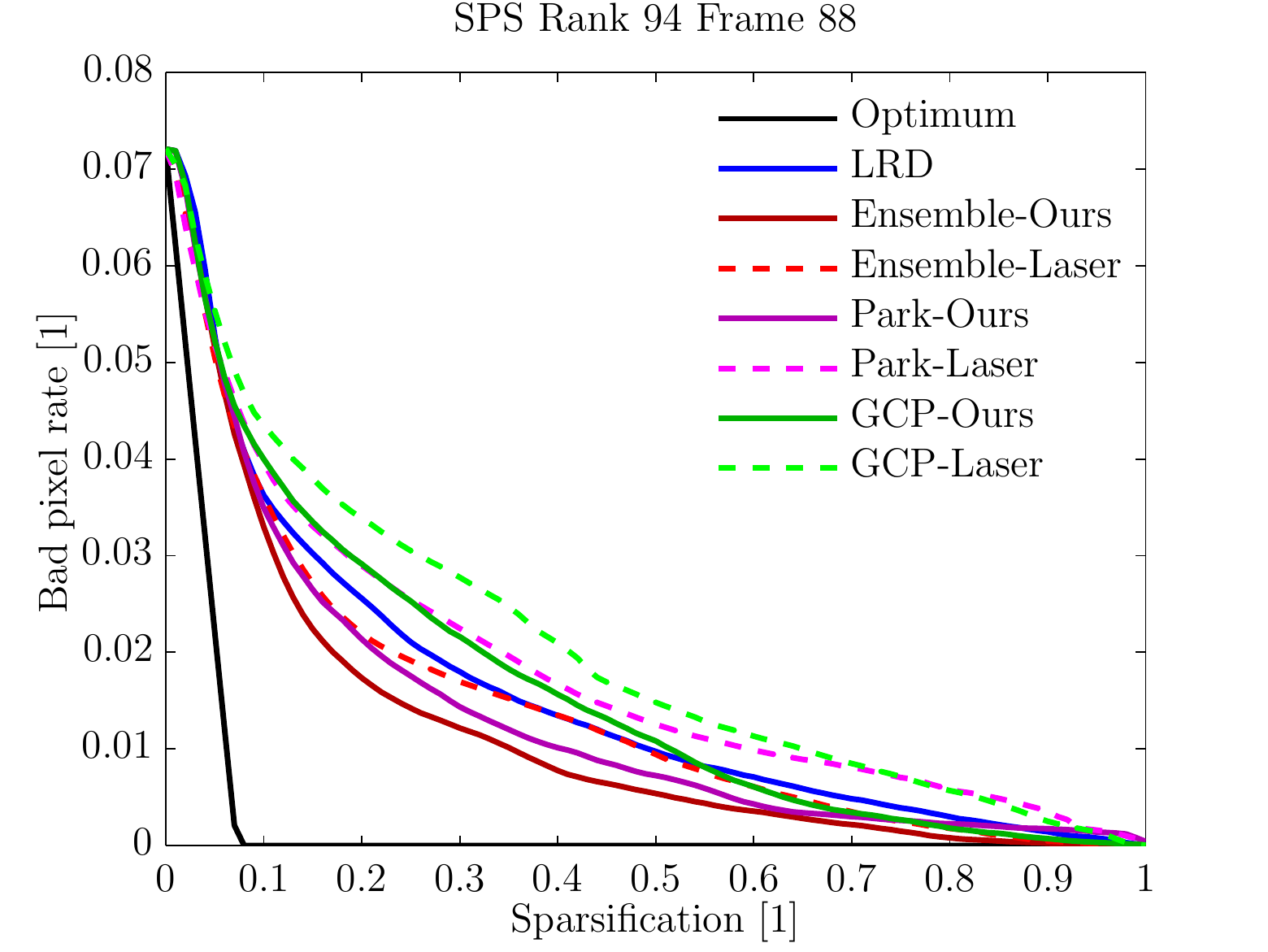} 
       
}\quad 

%\end{minipage}
    \caption{Sparsification curves for the three \textbf{ median frames} (85,126,88) of the KITTI training dataset. 
    The ranking was obtained by the ratio $AUSC_{SGM-Park-Ours}/AUSC_{SGM-Park-Laser}$. We display all combinations of 
    query  algorithm (SGM~\cite{rothermel12} and SPS~\cite{yamaguchi14}), confidence prediction algorithm (Ensemble~\cite{haeusler13}, GCP~\cite{spyro14}, Park~\cite{park15}) and
    training data (Laser and Ours). As a baseline method we also show the Left-Right disparity Difference (LRD).}
 \vspace{-20pt}
  \label{fig:slines_median}
\end{figure*}

\begin{figure*}[p]
     \vspace{-15pt}
  \centering

 \subfigure[SGM Frame 151 (details in Fig.~\ref{fig:sgm151}).]
    {

                \includegraphics[width=0.45\textwidth]{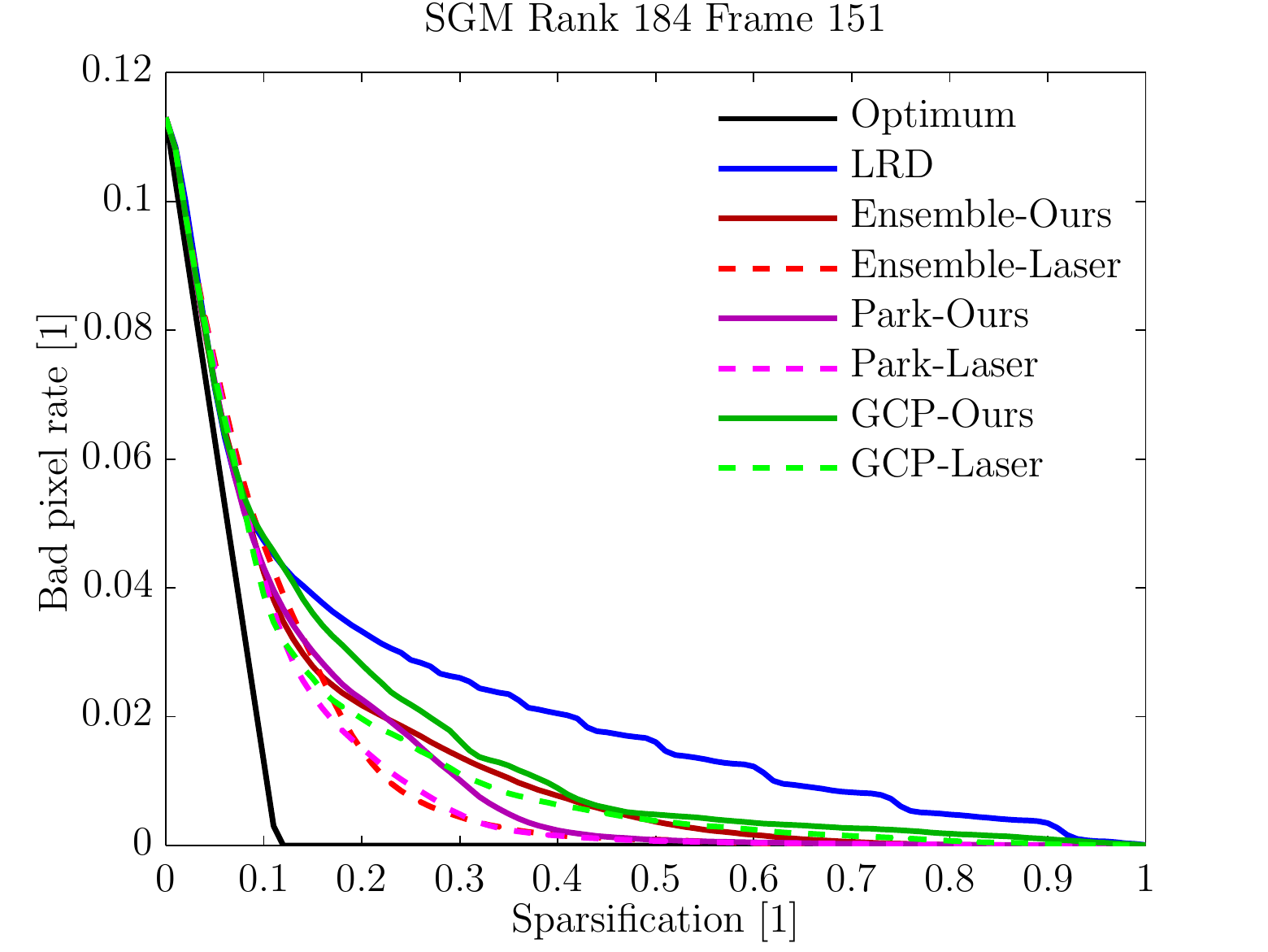} 
}\quad
 \subfigure[SPS Frame 151 (details in Fig.~\ref{fig:sps151}).]
    {

                \includegraphics[width=0.45\textwidth]{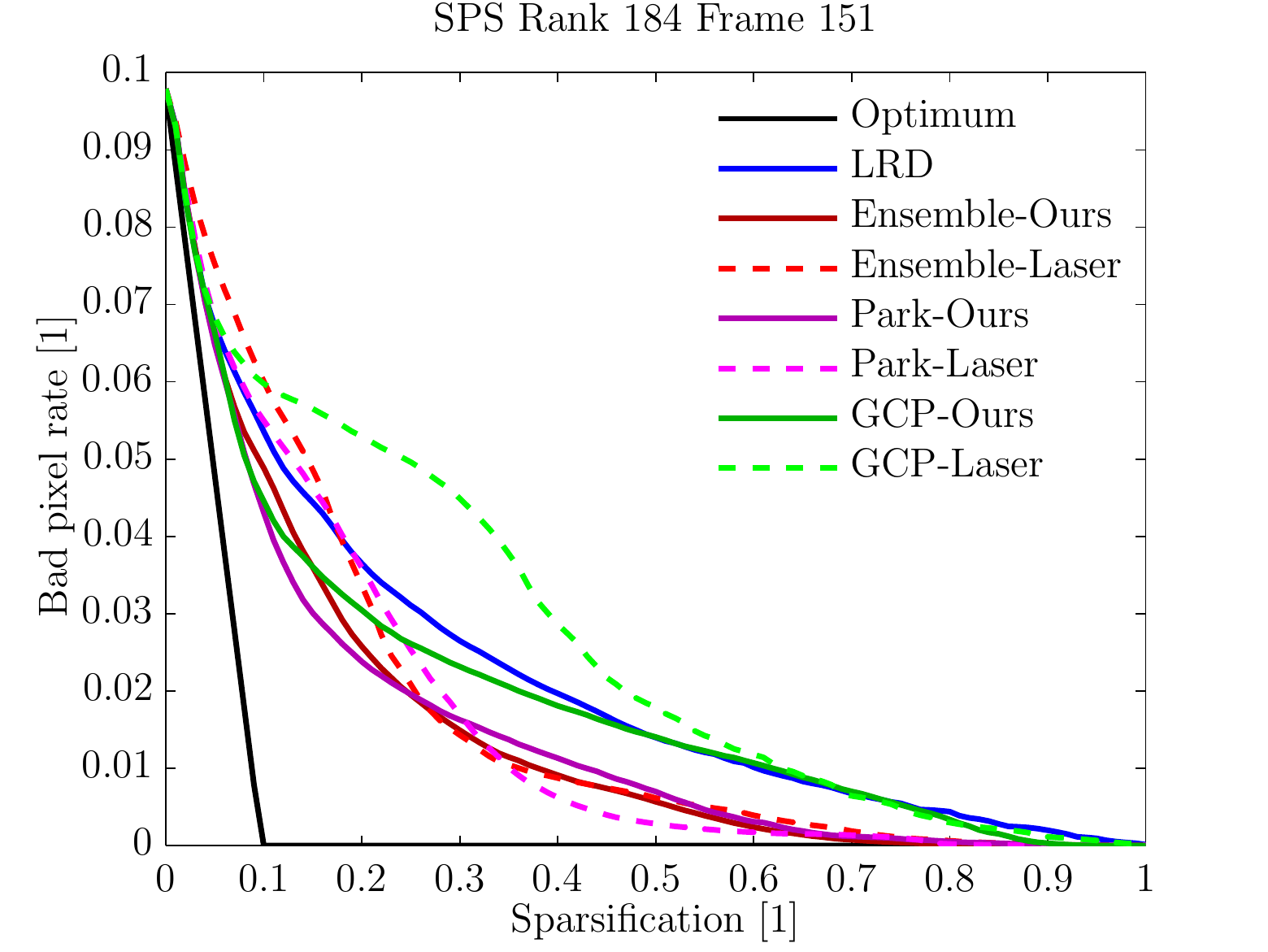} 
       
}\quad 

 \subfigure[SGM Frame 35 (details in Fig.~\ref{fig:sgm035}).]
    {

                \includegraphics[width=0.45\textwidth]{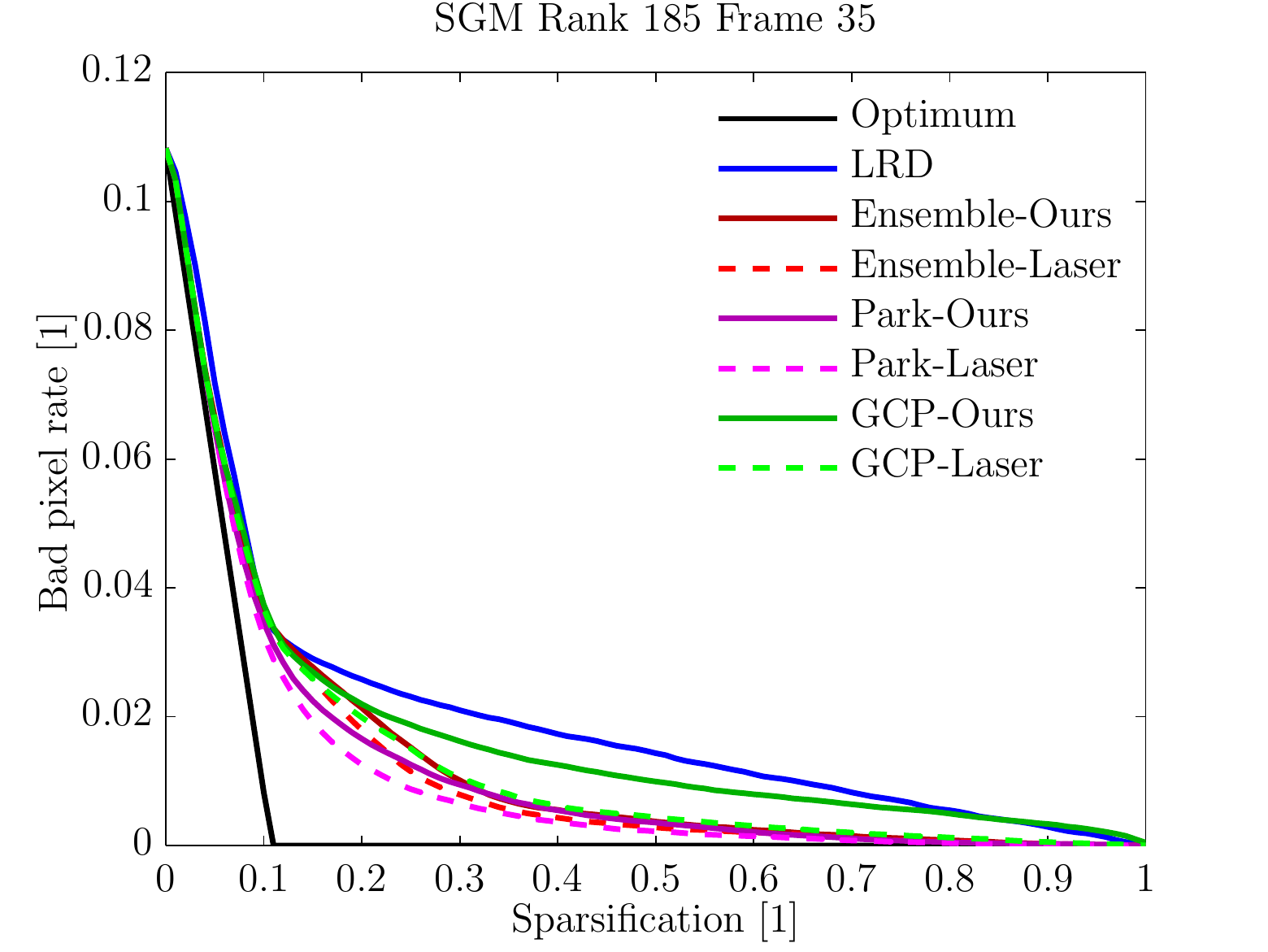} 
}\quad
 \subfigure[SPS Frame 35 (details in Fig.~\ref{fig:sps035}).]
    {

                \includegraphics[width=0.45\textwidth]{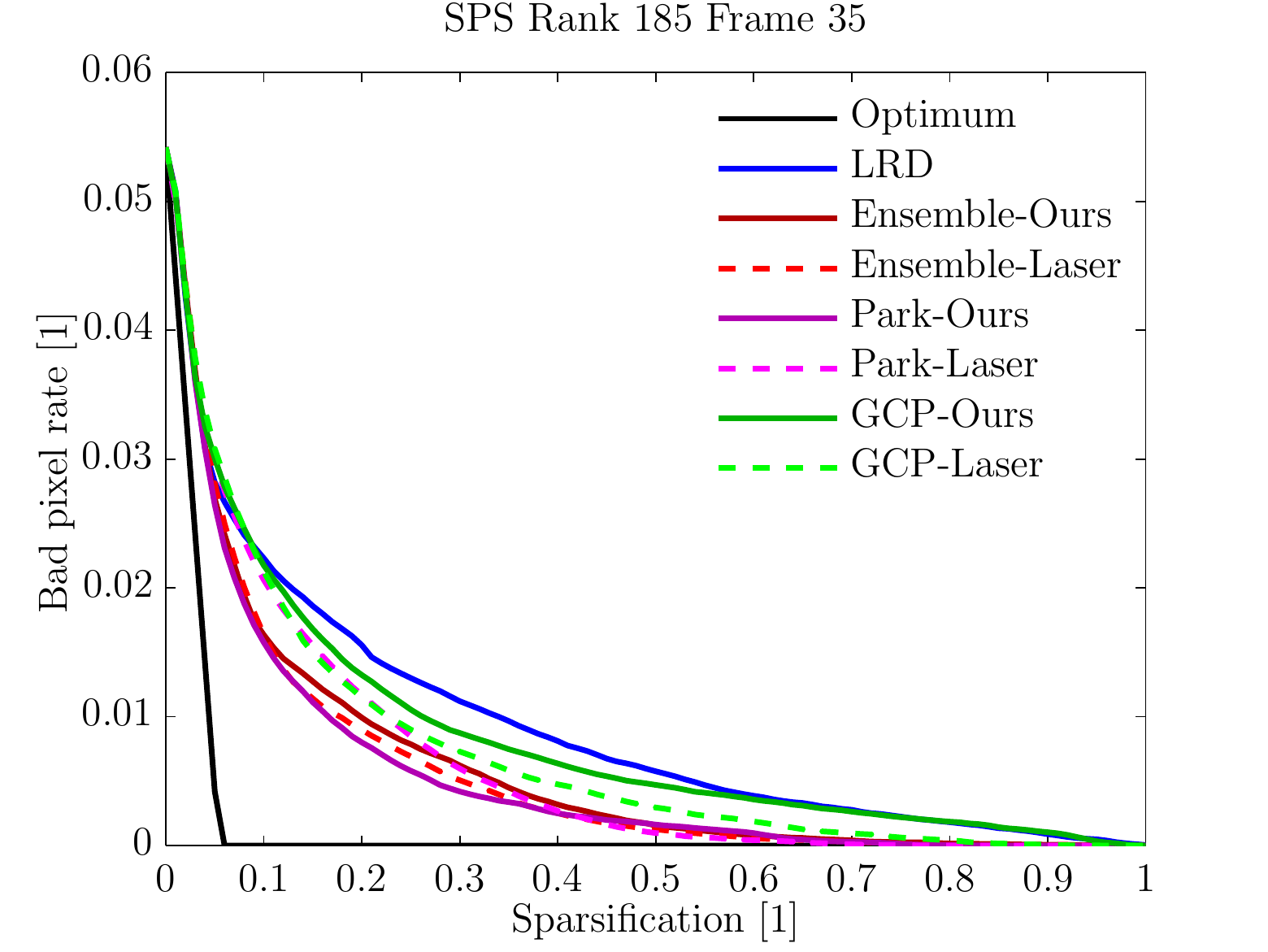} 
       
}\quad 

 \subfigure[SGM Frame 42 (details in Fig.~\ref{fig:sgm042}).]
    {

                \includegraphics[width=0.45\textwidth]{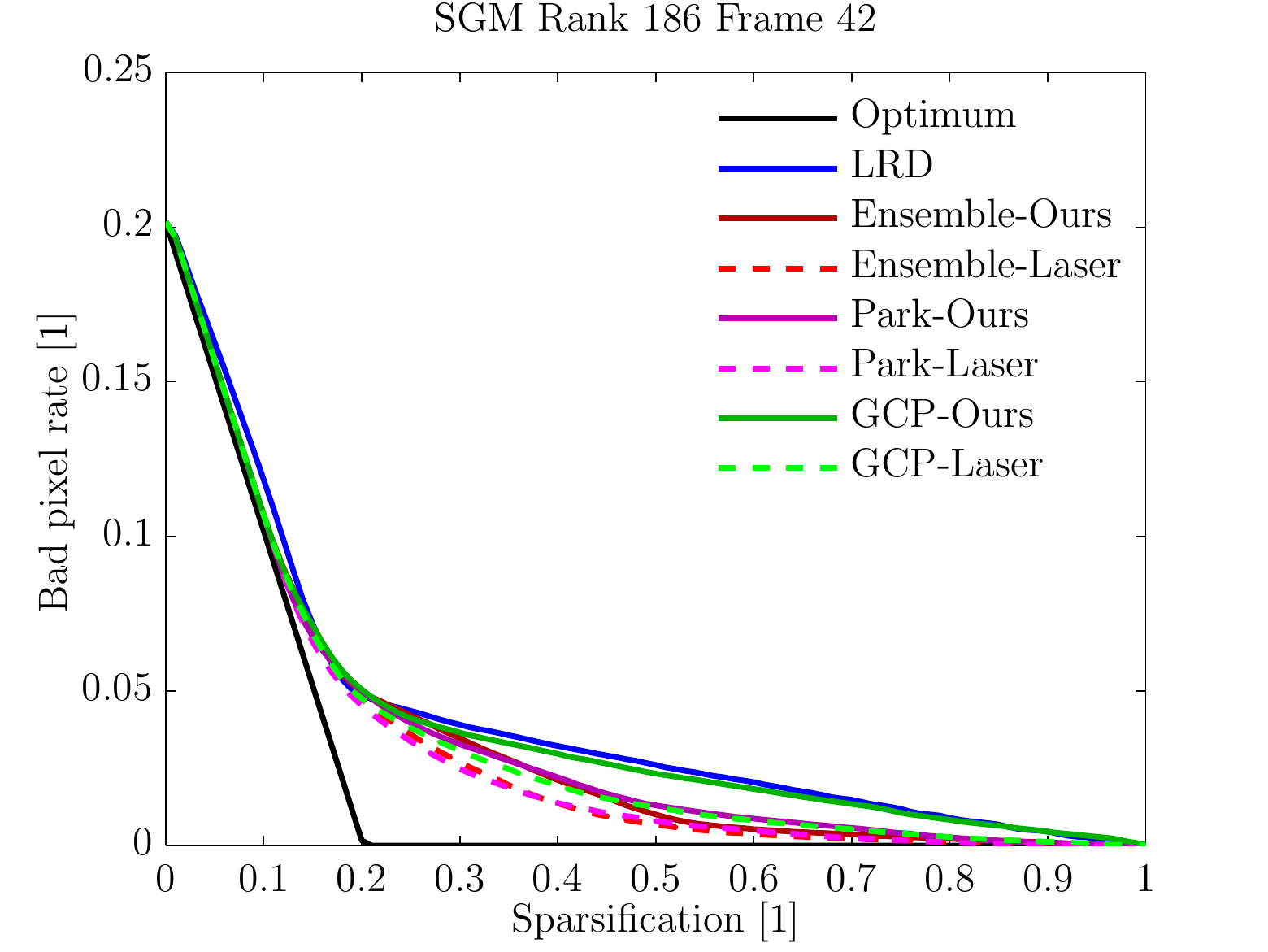} 
}\quad
 \subfigure[SPS Frame 42 (details in Fig.~\ref{fig:sps042}).]
    {

                \includegraphics[width=0.45\textwidth]{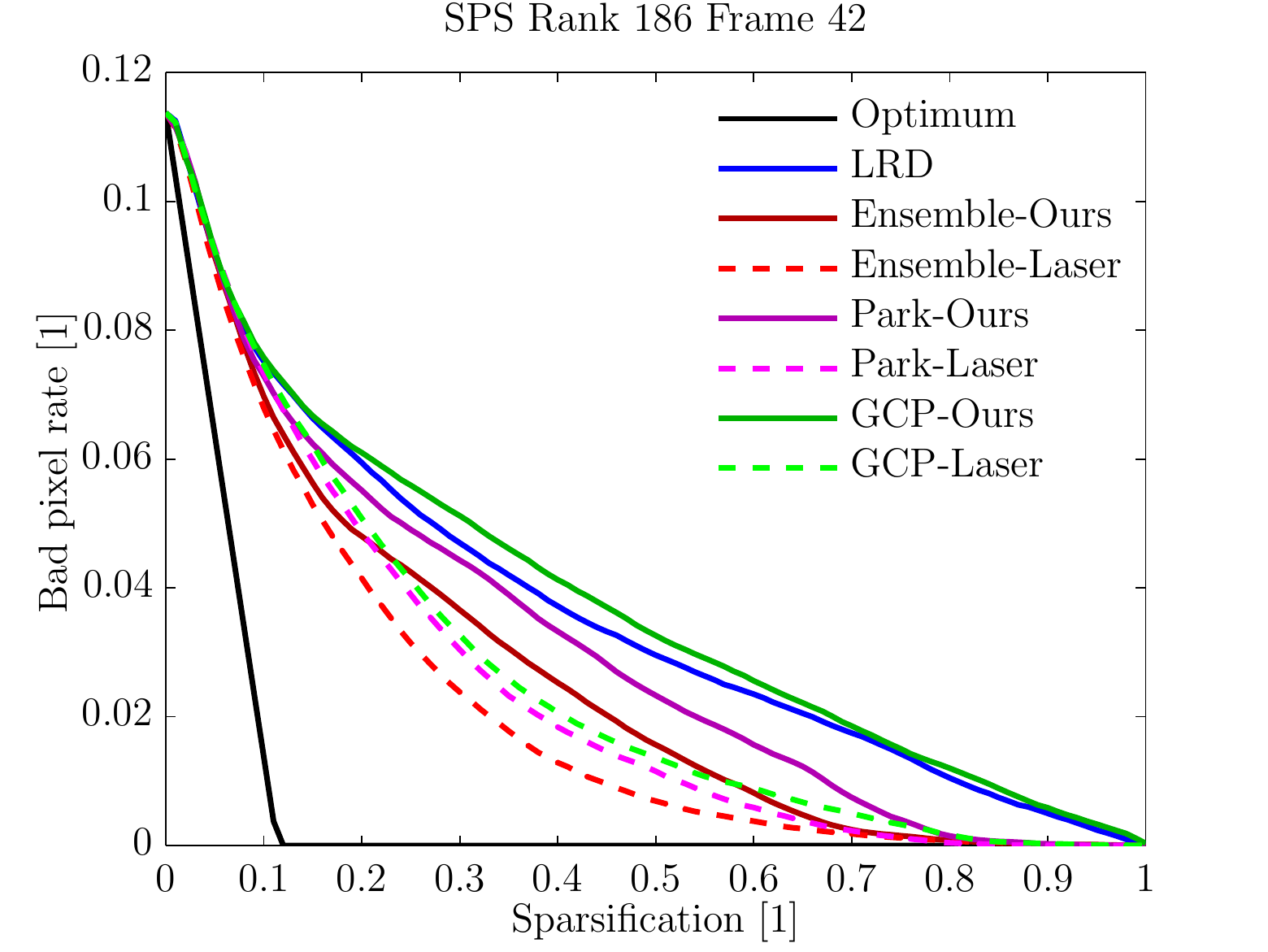} 
       
}\quad 

%\end{minipage}
    \caption{Sparsification curves for the three \textbf{worst frames} (151,35,42) of the KITTI training dataset.
    The ranking was obtained by the ratio $AUSC_{SGM-Park-Ours}/AUSC_{SGM-Park-Laser}$.  We display all combinations of 
    query  algorithm (SGM~\cite{rothermel12} and SPS~\cite{yamaguchi14}), confidence prediction algorithm (Ensemble~\cite{haeusler13}, GCP~\cite{spyro14}, Park~\cite{park15}) and
    training data (Laser and Ours). As a baseline method we also show the Left-Right disparity Difference (LRD).}
 \vspace{-20pt}
  \label{fig:slines_highest}
\end{figure*}

Figures~\ref{fig:slines_lowest},~\ref{fig:slines_median} and \ref{fig:slines_highest} are 
extensions to Fig. 5 in the original paper.
These figures each show the sparsification curves for three frames of the KITTI2012 dataset~\cite{geiger12}.
To give a fair overview of good and bad examples, we ranked the frames depending on the benefit of our training data.
As we ran multiple combinations of query  algorithm (SGM~\cite{rothermel12} and SPS~\cite{yamaguchi14}) and confidence prediction algorithm (Ensemble~\cite{haeusler13}, GCP~\cite{spyro14}, Park~\cite{park15}),
we decided to base the ranking on the combination with the lowest total Area Under the Sparsification Curve (AUSC)
which turned out to be "SGM-Park" (see Tab. 1 in the original paper).
Thus, we used the ratio $AUSC_{SGM-Park-Ours}/AUSC_{SGM-Park-Laser}$ for ranking the frames 
of the KITTI training dataset in ascending order.
Based on this ranking we selected the three best frames (Fig.~\ref{fig:slines_lowest}),
the three frames around the median index (Fig.~\ref{fig:slines_median}) and 
the three worst frames (Fig.~\ref{fig:slines_highest}).

For each of these nine frames we show two additional figures (one for SGM~\cite{rothermel12} (Fig. 4-12) and one for SPS~\cite{yamaguchi14} (Fig. 13-21)).
Each of these figures shows the RGB input image, the depth image produced by the query algorithm (SGM or SPS), the label images 
generated with the laser ground truth~\cite{geiger12} and with our approach, as well as 
the confidence prediction output for all combinations of confidence prediction algorithm (Ensemble~\cite{haeusler13}, GCP~\cite{spyro14}, Park~\cite{park15})
and training data (Laser~\cite{geiger12} and Ours).

\begin{figure*}[p]

  \centering

                \includegraphics[width=1\textwidth]{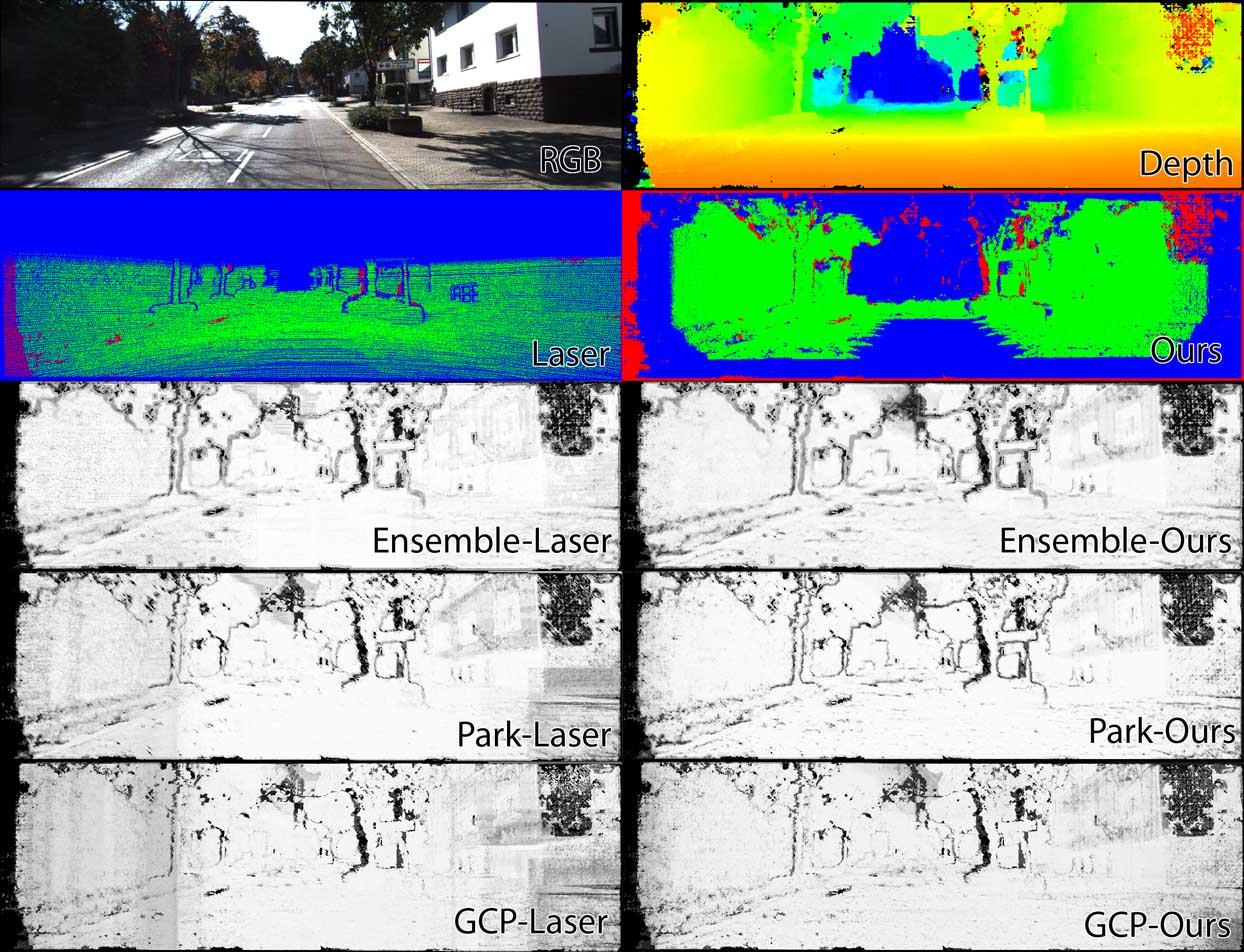}

%\end{minipage}
    \caption{Visual comparison of outputs for frame 0 and SGM~\cite{rothermel12} as query algorithm (rank 3 - best). Top row: RGB input image and SGM~\cite{rothermel12} depthmap.
    The color in the depth images ranges from blue (far away) to red (very close).
    Second row: Label images computed with laser ground truth and with our approach.
    In the label images the color green stands for positive samples,
    red for negative and blue is ignored during training and evaluation.
    In the remaining rows, we display the confidence prediction output for all combinations of 
    confidence prediction algorithm (Ensemble~\cite{haeusler13}, GCP~\cite{spyro14}, Park~\cite{park15}) and
    training data (Laser and Ours). The color ranges from black (low confidence) to white (high confidence).}
  \label{fig:sgm000}
\end{figure*}

For the label image generation, note that the amount of generated positive training data is strongly influenced
by the camera motion.
We require at least one reference measurement which is more accurate than the query measurement and has a minimum relative observation angle of at least $10^\circ$
(compare Fig.~\ref{fig:sgm000} and Fig.~\ref{fig:sgm088}).
This leads to a low coverage in the middle of the road,
while on the side of the image the labels are more dense.

For the confidence prediction, note that the prediction output trained on our label images (Ours) is less noisy and impact of the 
distance to border features becomes smoother (e.g. Fig.~\ref{fig:sgm000} left margin and bottom right corner of the image).
It seems that the random forests over-fit if we only use the limited amount of label images generated with the laser ground truth.
In contrast, our more diverse training data reduces the chance of over-fitting, which in turn leads to an improved overall prediction performance.

As a matter of completeness,
Figure~\ref{fig:kitti_ausc} shows a different version of Fig. 6 in the original paper.
While Fig. 6 in the original paper shows the mean, minimum and maximum values across the three confidence
prediction approaches (Ensemble~\cite{haeusler13}, GCP~\cite{spyro14}, Park~\cite{park15}),
Fig.~\ref{fig:kitti_ausc} shows each approach individually.
The sequences in Fig.~\ref{fig:kitti_ausc} were sorted by the optimal AUSC values.

\FloatBarrier

%%%%

\begin{figure*}[p]

  \centering

                \includegraphics[width=1\textwidth]{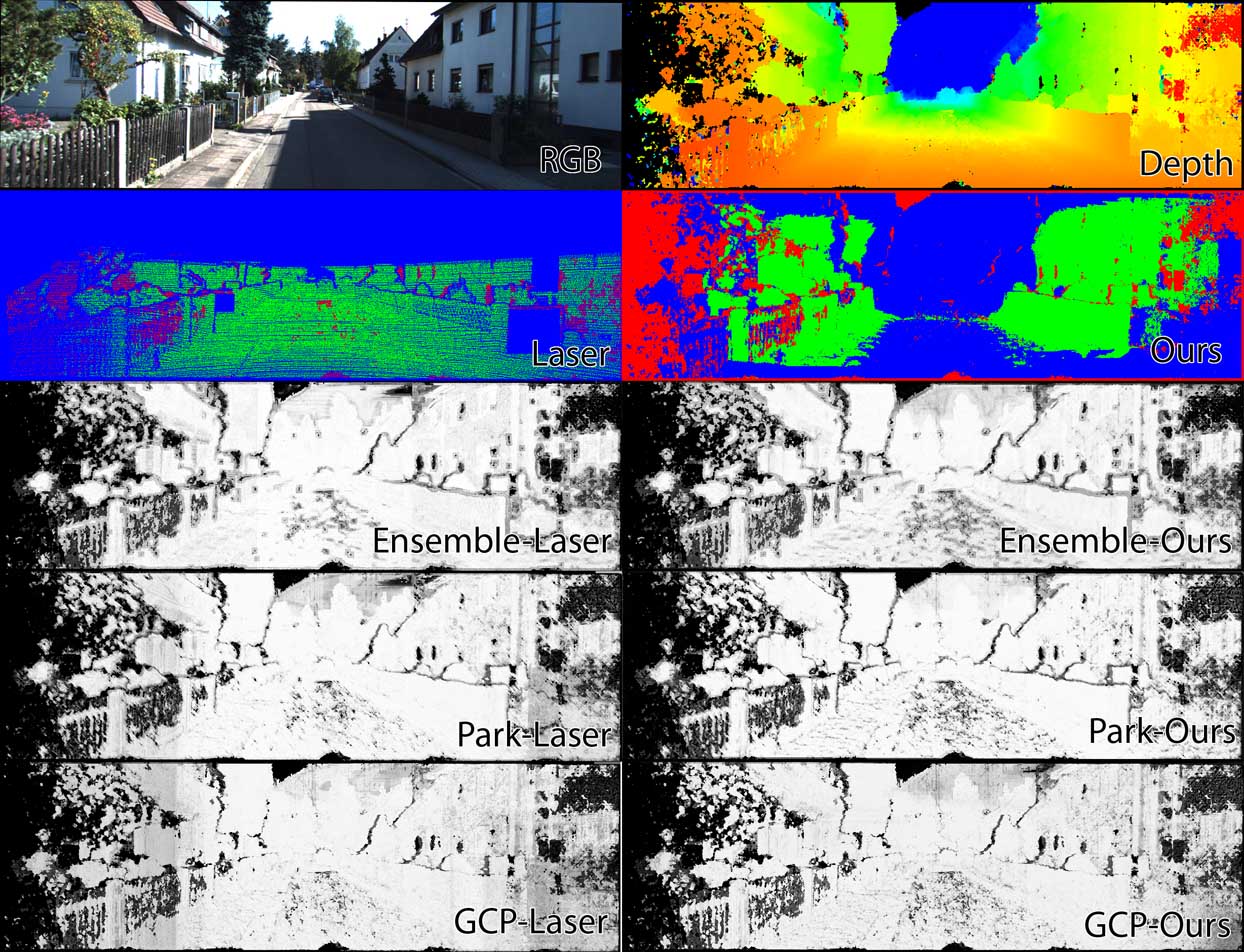}

%\end{minipage}
    \caption{Visual comparison of outputs for frame 35 and SGM~\cite{rothermel12} as query algorithm (rank 185 - worst). Top row: RGB input image and SGM~\cite{rothermel12} depthmap.
    The color in the depth images ranges from blue (far away) to red (very close).
    Second row: Label images computed with laser ground truth and with our approach.
    In the label images the color green stands for positive samples,
    red for negative and blue is ignored during training and evaluation.
    In the remaining rows, we display the confidence prediction output for all combinations of 
    confidence prediction algorithm (Ensemble~\cite{haeusler13}, GCP~\cite{spyro14}, Park~\cite{park15}) and
    training data (Laser and Ours). The color ranges from black (low confidence) to white (high confidence).}
  \label{fig:sgm035}
\end{figure*}

\begin{figure*}[p]

  \centering

                \includegraphics[width=1\textwidth]{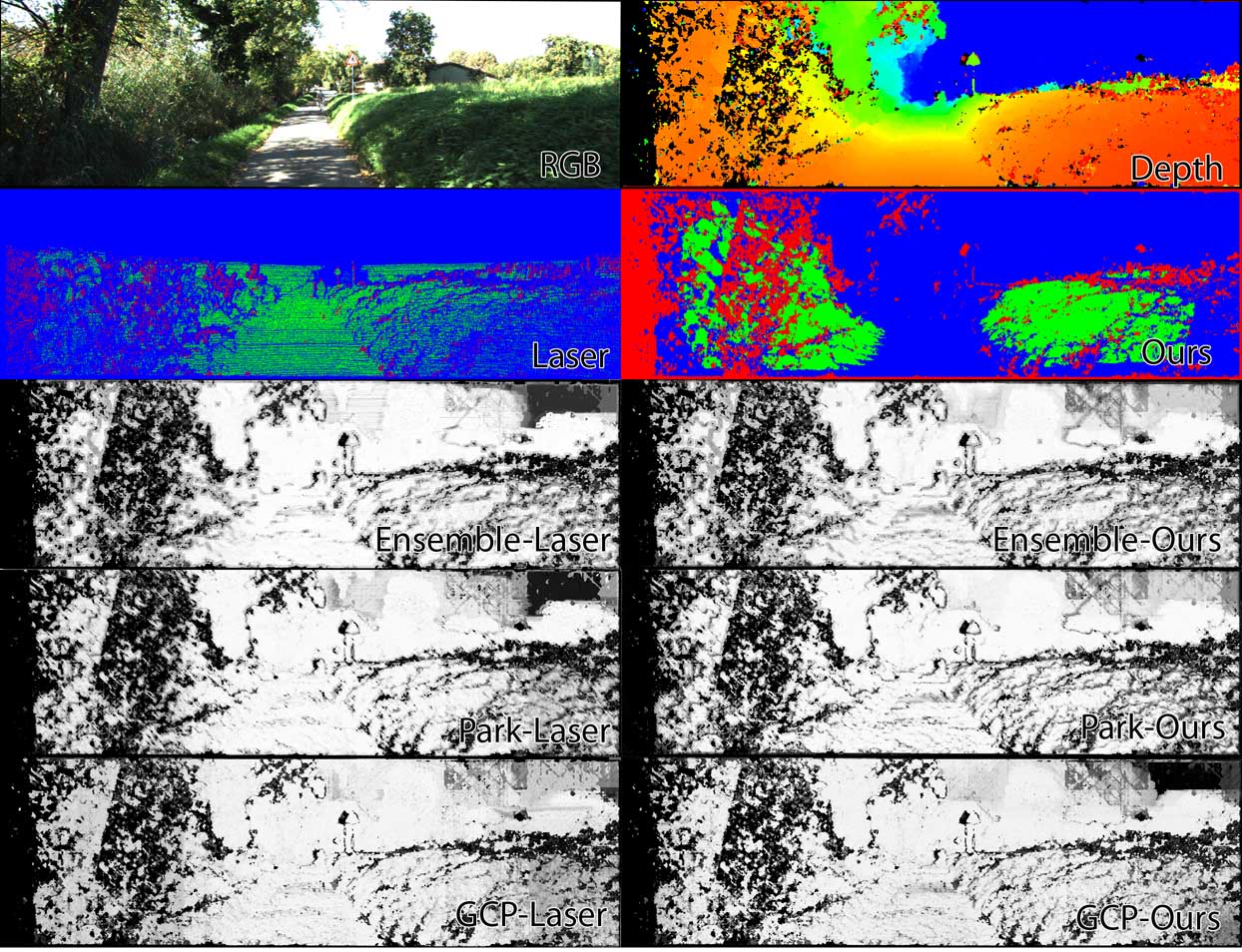}

%\end{minipage}
    \caption{Visual comparison of outputs for frame 42 and SGM~\cite{rothermel12} as query algorithm (rank 186 - worst). Top row: RGB input image and SGM~\cite{rothermel12} depthmap.
    The color in the depth images ranges from blue (far away) to red (very close).
    Second row: Label images computed with laser ground truth and with our approach.
    In the label images the color green stands for positive samples,
    red for negative and blue is ignored during training and evaluation.
    In the remaining rows, we display the confidence prediction output for all combinations of 
    confidence prediction algorithm (Ensemble~\cite{haeusler13}, GCP~\cite{spyro14}, Park~\cite{park15}) and
    training data (Laser and Ours). The color ranges from black (low confidence) to white (high confidence).}
  \label{fig:sgm042}
\end{figure*}

\begin{figure*}[p]

  \centering

                \includegraphics[width=1\textwidth]{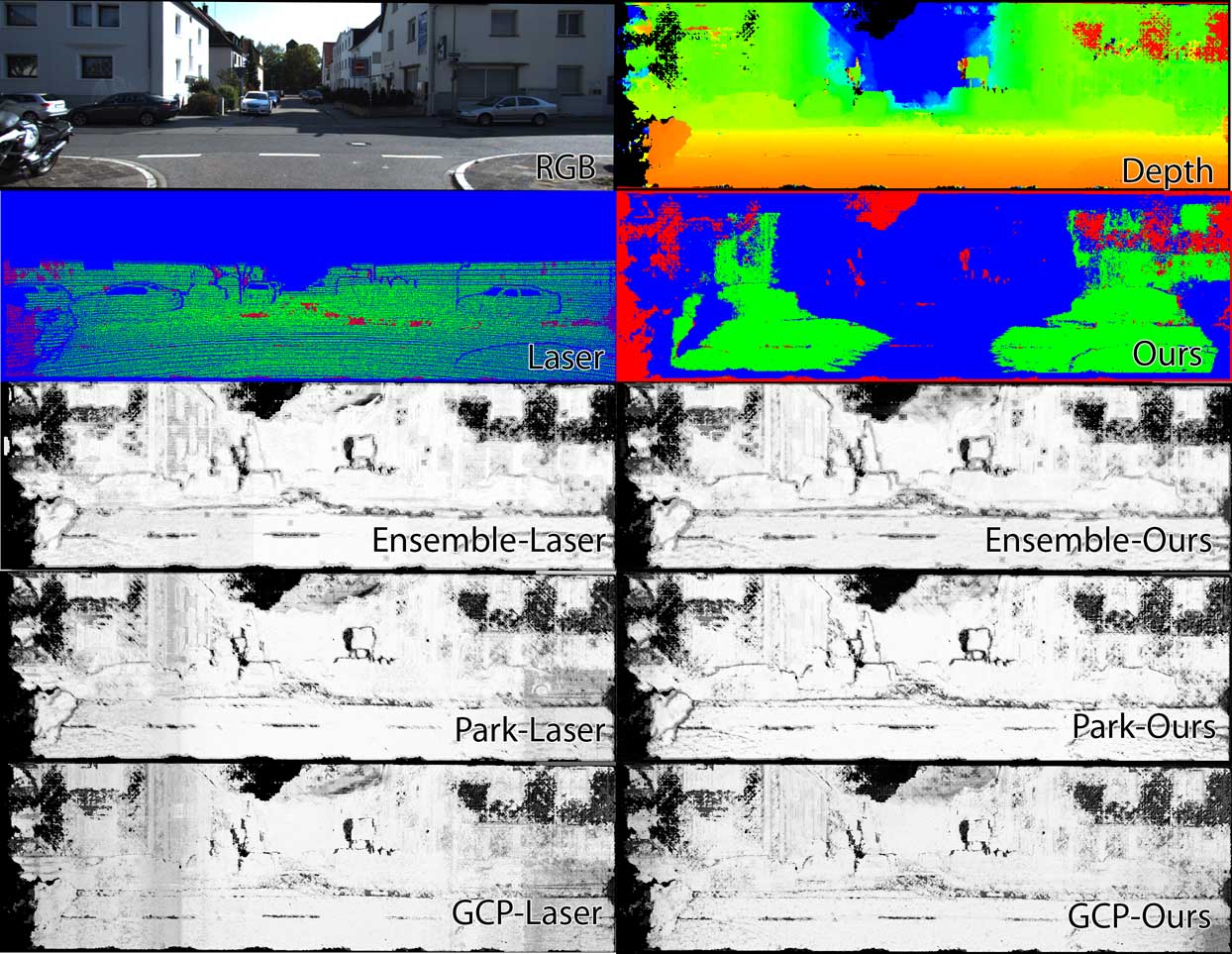}

%\end{minipage}
    \caption{Visual comparison of outputs for frame 56 and SGM~\cite{rothermel12} as query algorithm (rank 2 - best). Top row: RGB input image and SGM~\cite{rothermel12} depthmap.
    The color in the depth images ranges from blue (far away) to red (very close).
    Second row: Label images computed with laser ground truth and with our approach.
    In the label images the color green stands for positive samples,
    red for negative and blue is ignored during training and evaluation.
    In the remaining rows, we display the confidence prediction output for all combinations of 
    confidence prediction algorithm (Ensemble~\cite{haeusler13}, GCP~\cite{spyro14}, Park~\cite{park15}) and
    training data (Laser and Ours). The color ranges from black (low confidence) to white (high confidence).}
  \label{fig:sgm056}
\end{figure*}

\begin{figure*}[p]

  \centering

                \includegraphics[width=1\textwidth]{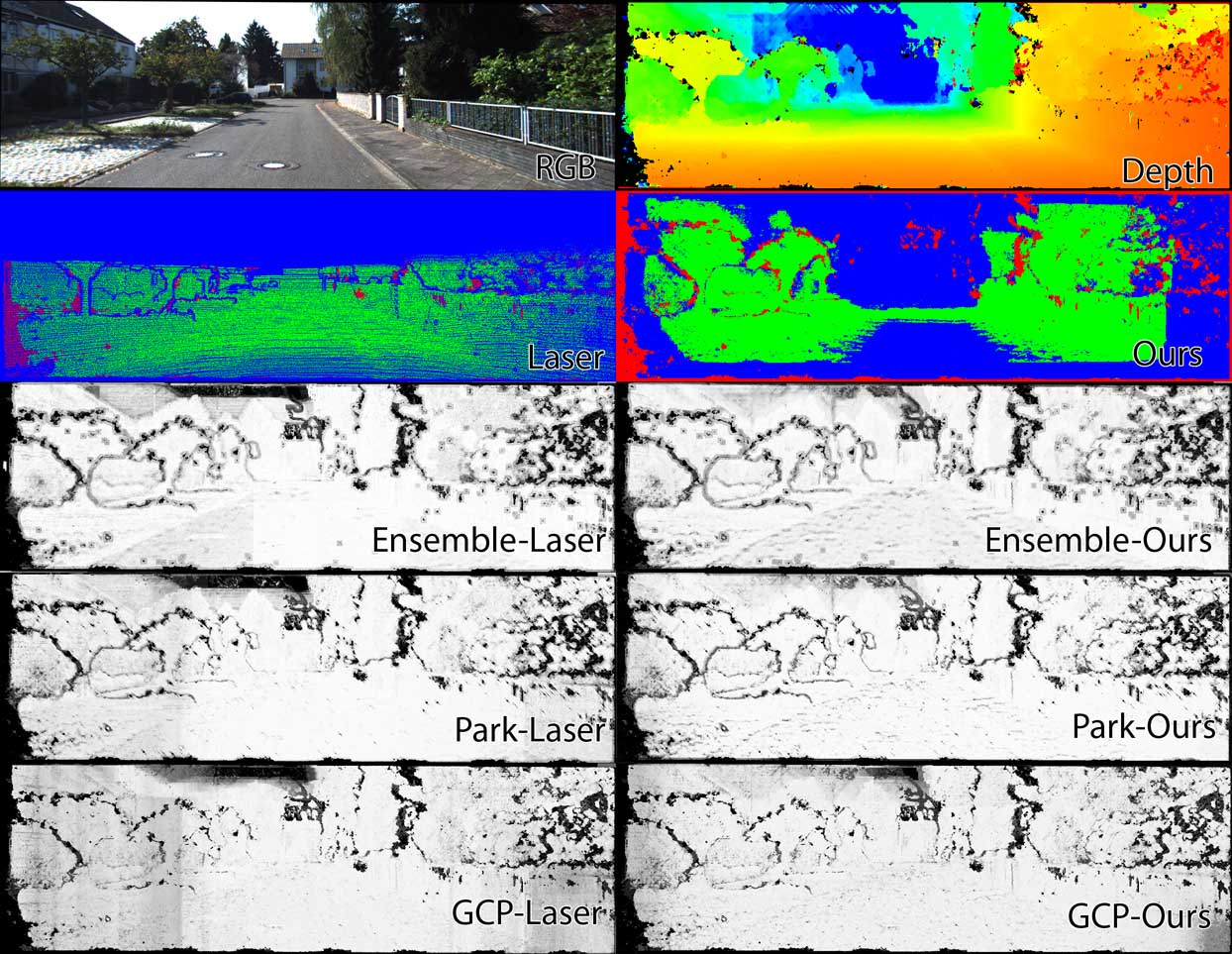}

%\end{minipage}
    \caption{Visual comparison of outputs for frame 85 and SGM~\cite{rothermel12} as query algorithm (rank 92 - median). Top row: RGB input image and SGM~\cite{rothermel12} depthmap.
    The color in the depth images ranges from blue (far away) to red (very close).
    Second row: Label images computed with laser ground truth and with our approach.
    In the label images the color green stands for positive samples,
    red for negative and blue is ignored during training and evaluation.
    In the remaining rows, we display the confidence prediction output for all combinations of 
    confidence prediction algorithm (Ensemble~\cite{haeusler13}, GCP~\cite{spyro14}, Park~\cite{park15}) and
    training data (Laser and Ours). The color ranges from black (low confidence) to white (high confidence).}
  \label{fig:sgm085}
\end{figure*}

\begin{figure*}[p]

  \centering

                \includegraphics[width=1\textwidth]{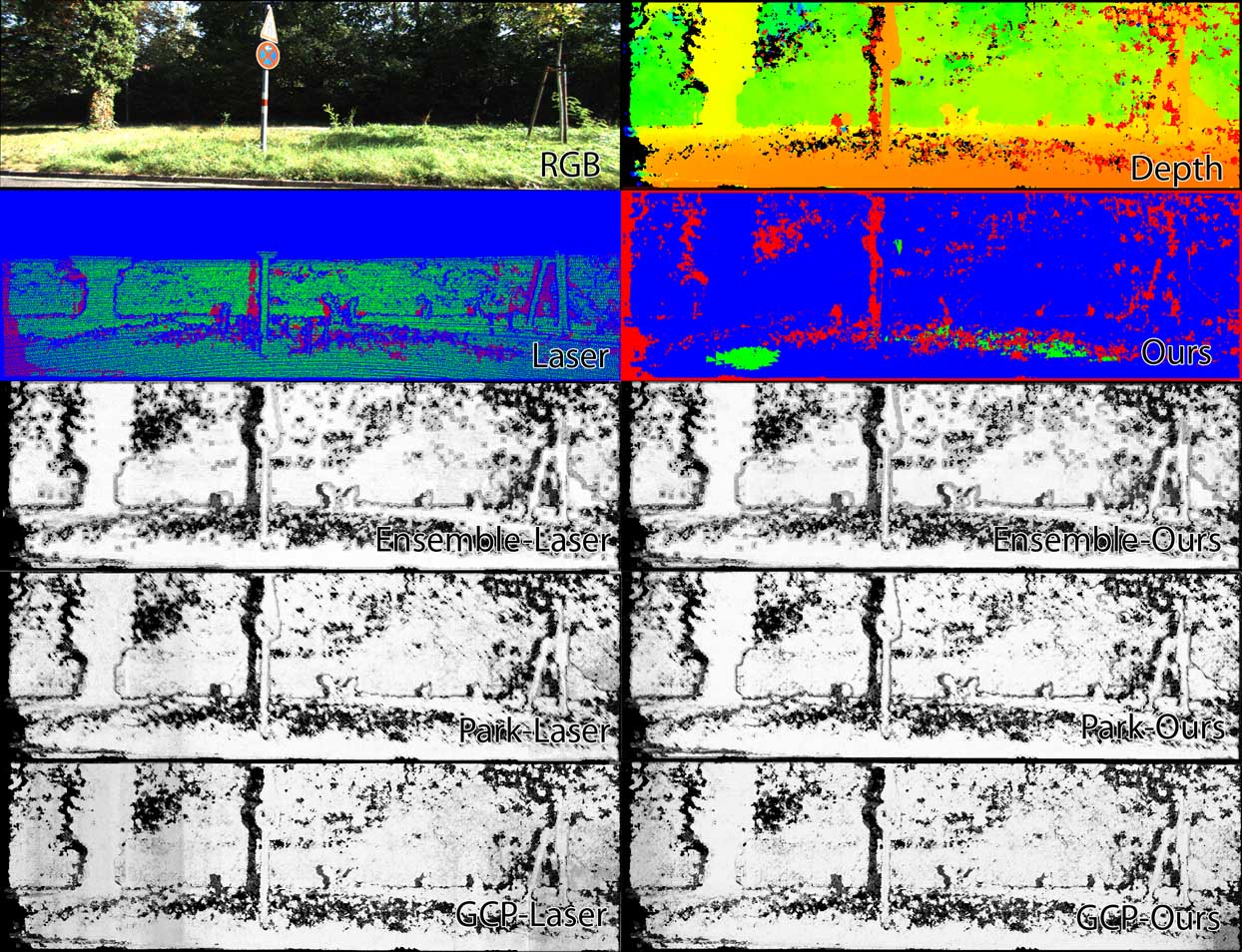}

%\end{minipage}
    \caption{Visual comparison of outputs for frame 88 and SGM~\cite{rothermel12} as query algorithm (rank 94 - median). Top row: RGB input image and SGM~\cite{rothermel12} depthmap.
    The color in the depth images ranges from blue (far away) to red (very close).
    Second row: Label images computed with laser ground truth and with our approach.
    In the label images the color green stands for positive samples,
    red for negative and blue is ignored during training and evaluation.
    In the remaining rows, we display the confidence prediction output for all combinations of 
    confidence prediction algorithm (Ensemble~\cite{haeusler13}, GCP~\cite{spyro14}, Park~\cite{park15}) and
    training data (Laser and Ours). The color ranges from black (low confidence) to white (high confidence).}
  \label{fig:sgm088}
\end{figure*}

\begin{figure*}[p]

  \centering

                \includegraphics[width=1\textwidth]{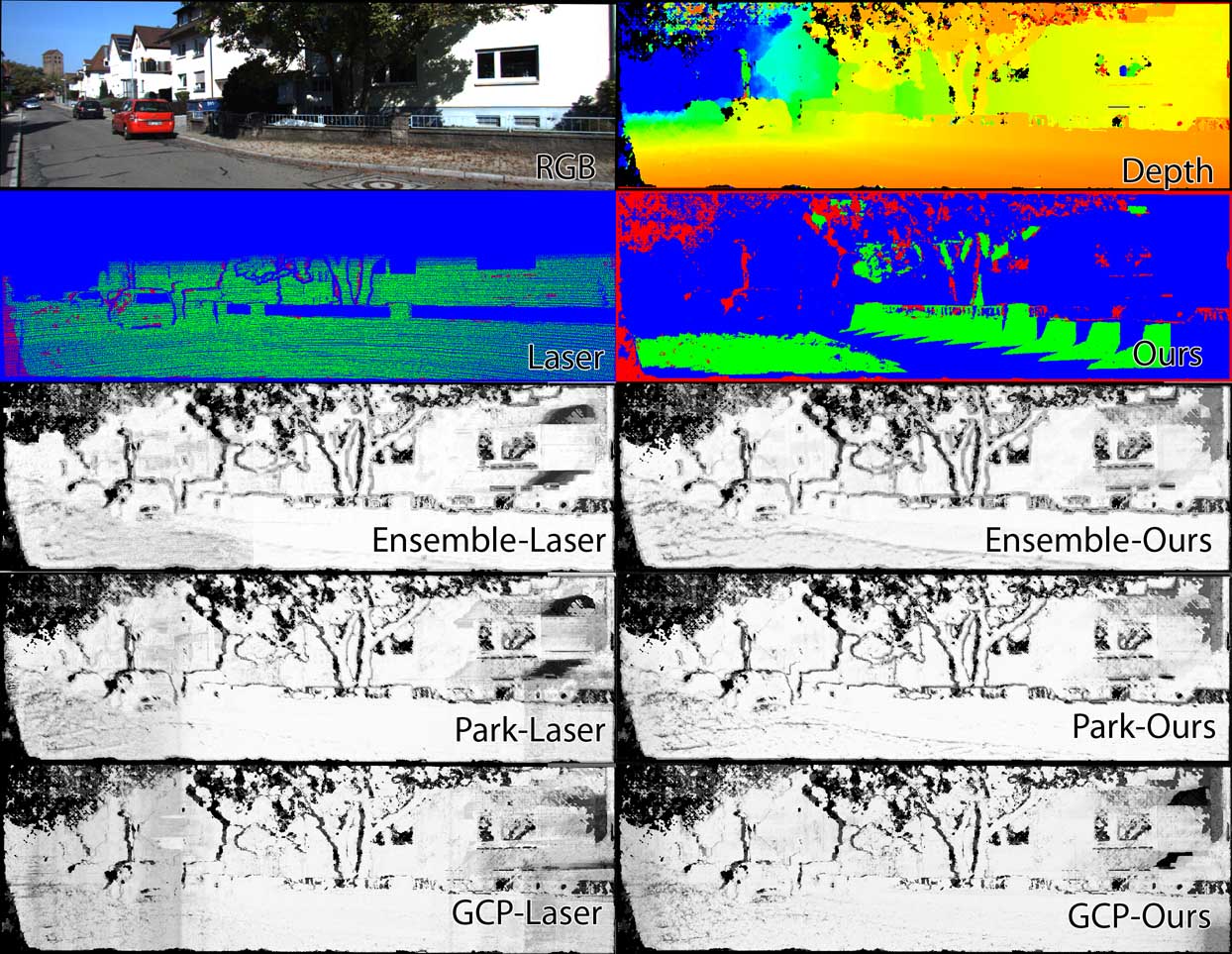}

%\end{minipage}
    \caption{Visual comparison of outputs for frame 102 and SGM~\cite{rothermel12} as query algorithm (rank 1 - best). Top row: RGB input image and SGM~\cite{rothermel12} depthmap.
    The color in the depth images ranges from blue (far away) to red (very close).
    Second row: Label images computed with laser ground truth and with our approach.
    In the label images the color green stands for positive samples,
    red for negative and blue is ignored during training and evaluation.
    In the remaining rows, we display the confidence prediction output for all combinations of 
    confidence prediction algorithm (Ensemble~\cite{haeusler13}, GCP~\cite{spyro14}, Park~\cite{park15}) and
    training data (Laser and Ours). The color ranges from black (low confidence) to white (high confidence).}
  \label{fig:sgm102}
\end{figure*}

\begin{figure*}[p]

  \centering

                \includegraphics[width=1\textwidth]{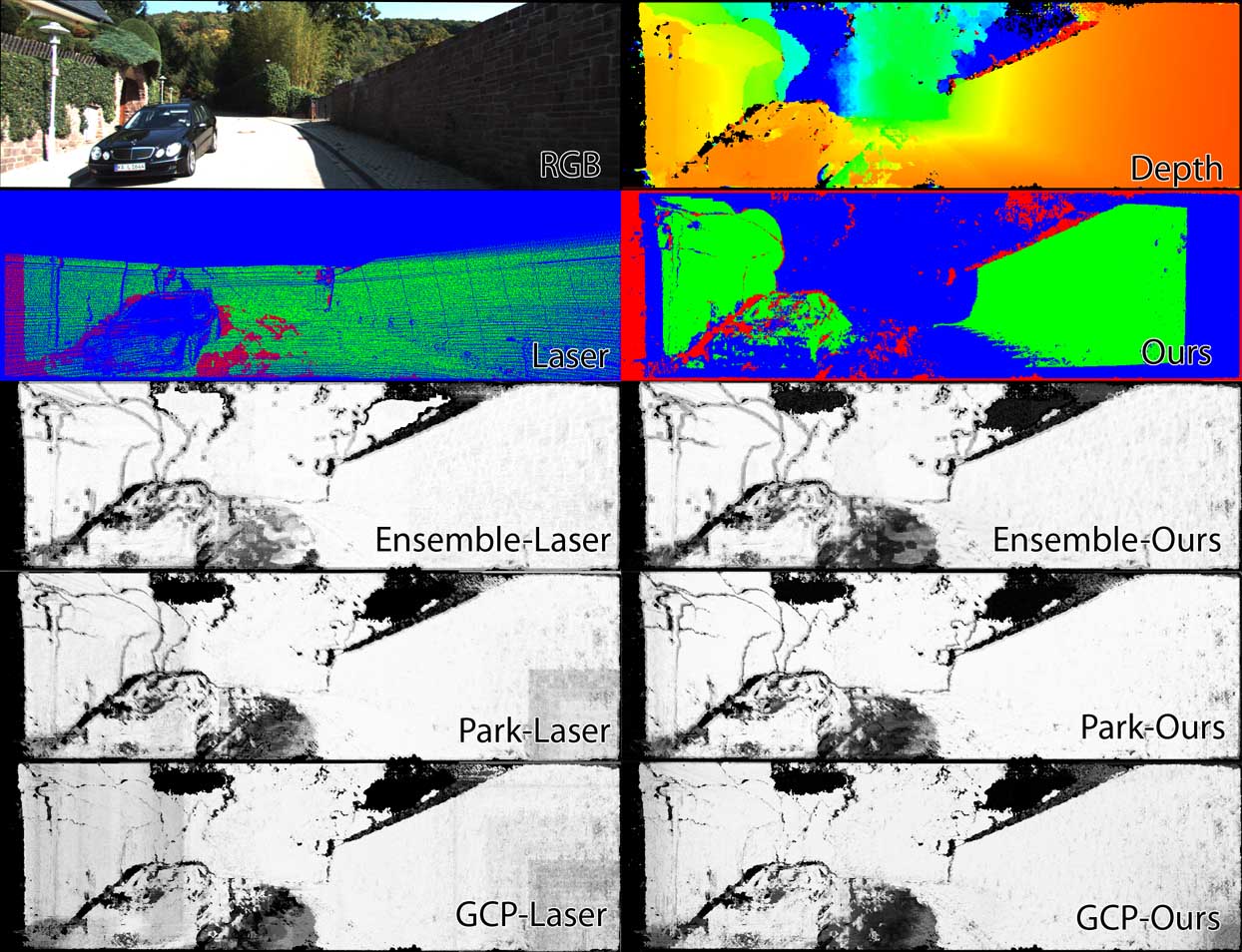}

%\end{minipage}
    \caption{Visual comparison of outputs for frame 126 and SGM~\cite{rothermel12} as query algorithm (rank 93 - median). Top row: RGB input image and SGM~\cite{rothermel12} depthmap.
    The color in the depth images ranges from blue (far away) to red (very close).
    Second row: Label images computed with laser ground truth and with our approach.
    In the label images the color green stands for positive samples,
    red for negative and blue is ignored during training and evaluation.
    In the remaining rows, we display the confidence prediction output for all combinations of 
    confidence prediction algorithm (Ensemble~\cite{haeusler13}, GCP~\cite{spyro14}, Park~\cite{park15}) and
    training data (Laser and Ours). The color ranges from black (low confidence) to white (high confidence).}
  \label{fig:sgm126}
\end{figure*}

\begin{figure*}[p]

  \centering

                \includegraphics[width=1\textwidth]{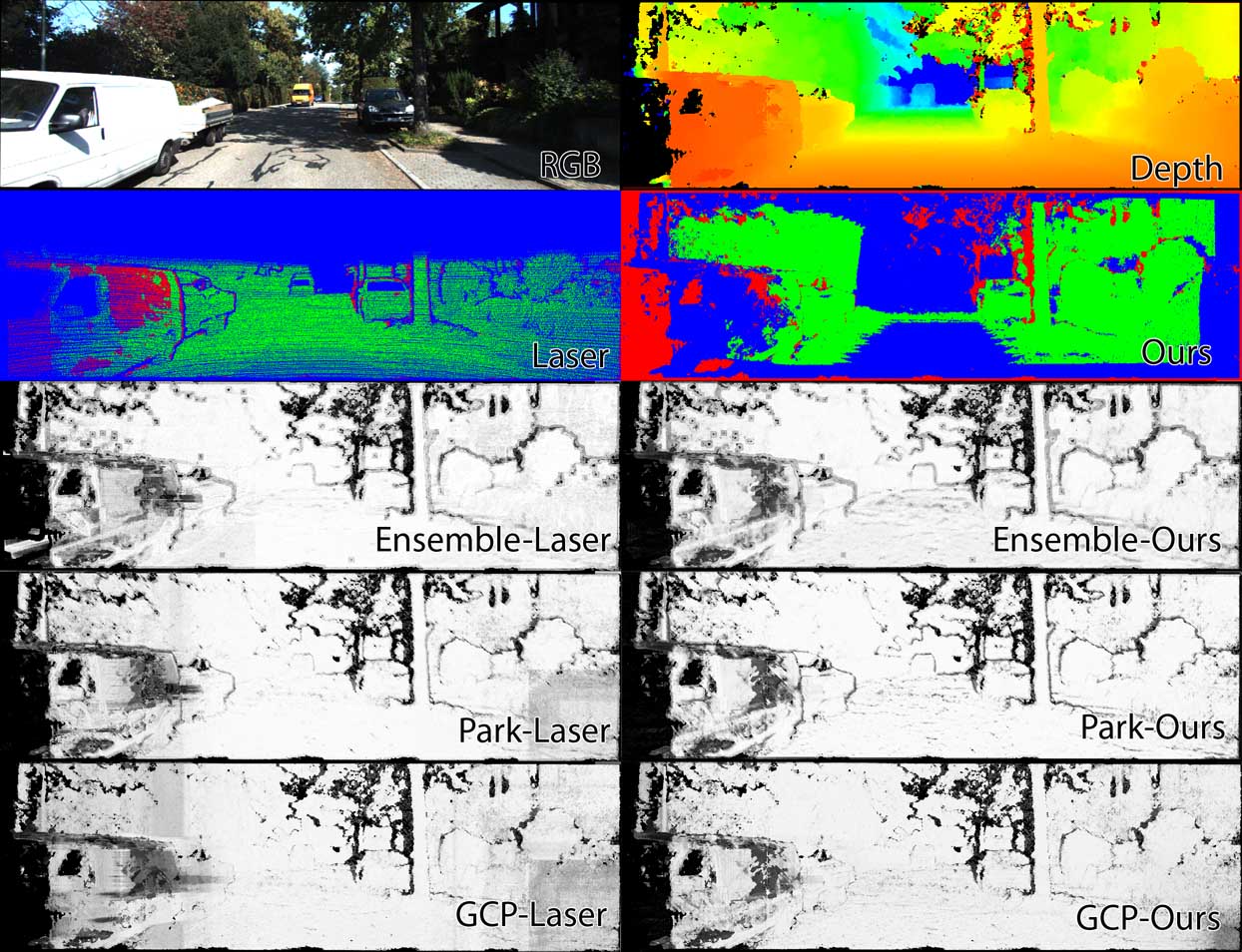}

%\end{minipage}
    \caption{Visual comparison of outputs for frame 151 and SGM~\cite{rothermel12} as query algorithm (rank 184 - worst). Top row: RGB input image and SGM~\cite{rothermel12} depthmap.
    The color in the depth images ranges from blue (far away) to red (very close).
    Second row: Label images computed with laser ground truth and with our approach.
    In the label images the color green stands for positive samples,
    red for negative and blue is ignored during training and evaluation.
    In the remaining rows, we display the confidence prediction output for all combinations of 
    confidence prediction algorithm (Ensemble~\cite{haeusler13}, GCP~\cite{spyro14}, Park~\cite{park15}) and
    training data (Laser and Ours). The color ranges from black (low confidence) to white (high confidence).}
  \label{fig:sgm151}
\end{figure*}

\begin{figure*}[p]

  \centering

                \includegraphics[width=1\textwidth]{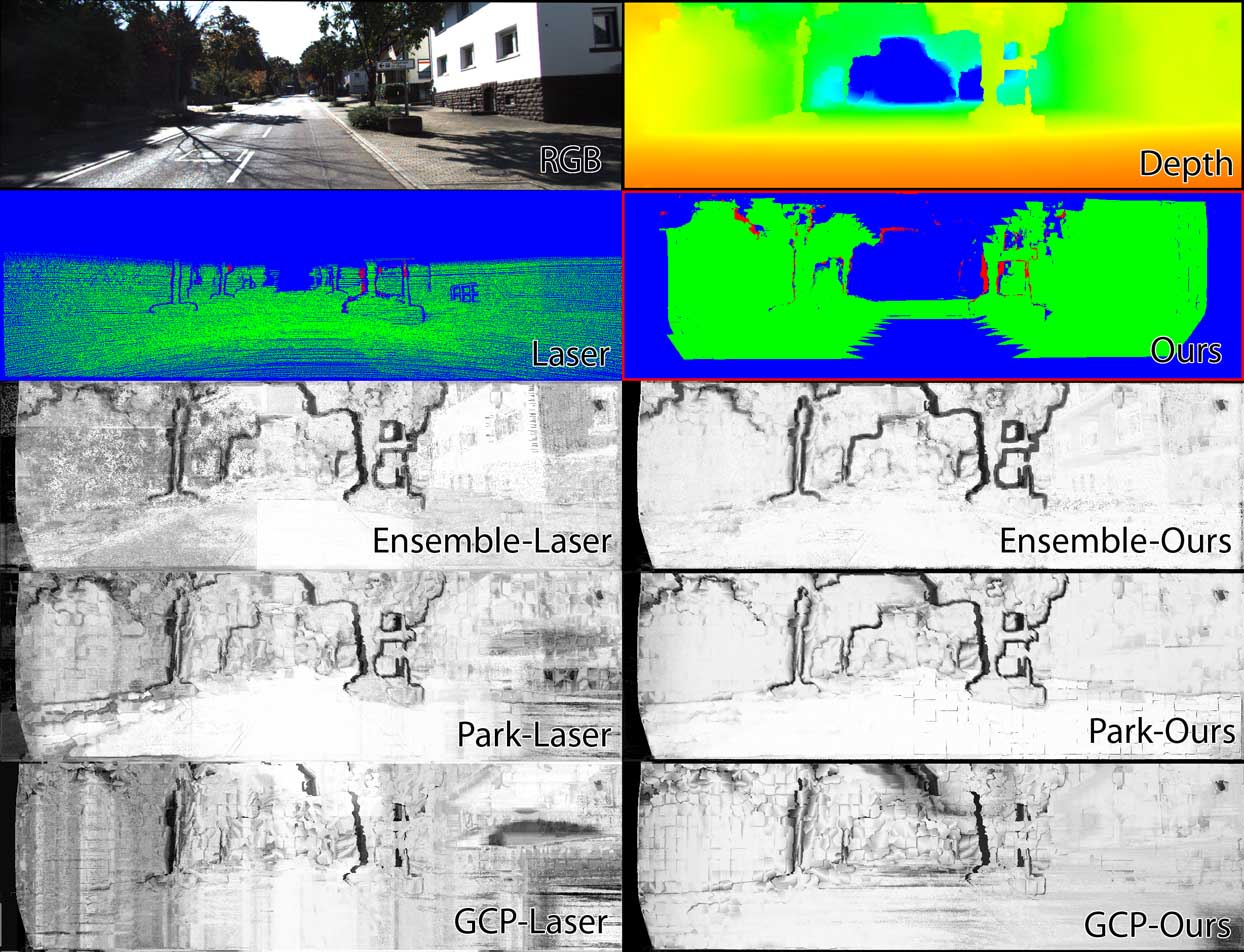}

%\end{minipage}
    \caption{Visual comparison of outputs for frame 0 and SPS~\cite{yamaguchi14} as query algorithm (rank 3 - best). Top row: RGB input image and SPS~\cite{yamaguchi14} depthmap.
    The color in the depth images ranges from blue (far away) to red (very close).
    Second row: Label images computed with laser ground truth and with our approach.
    In the label images the color green stands for positive samples,
    red for negative and blue is ignored during training and evaluation.
    In the remaining rows, we display the confidence prediction output for all combinations of 
    confidence prediction algorithm (Ensemble~\cite{haeusler13}, GCP~\cite{spyro14}, Park~\cite{park15}) and
    training data (Laser and Ours). The color ranges from black (low confidence) to white (high confidence).}
  \label{fig:sps000}
\end{figure*}

\begin{figure*}[p]

  \centering

                \includegraphics[width=1\textwidth]{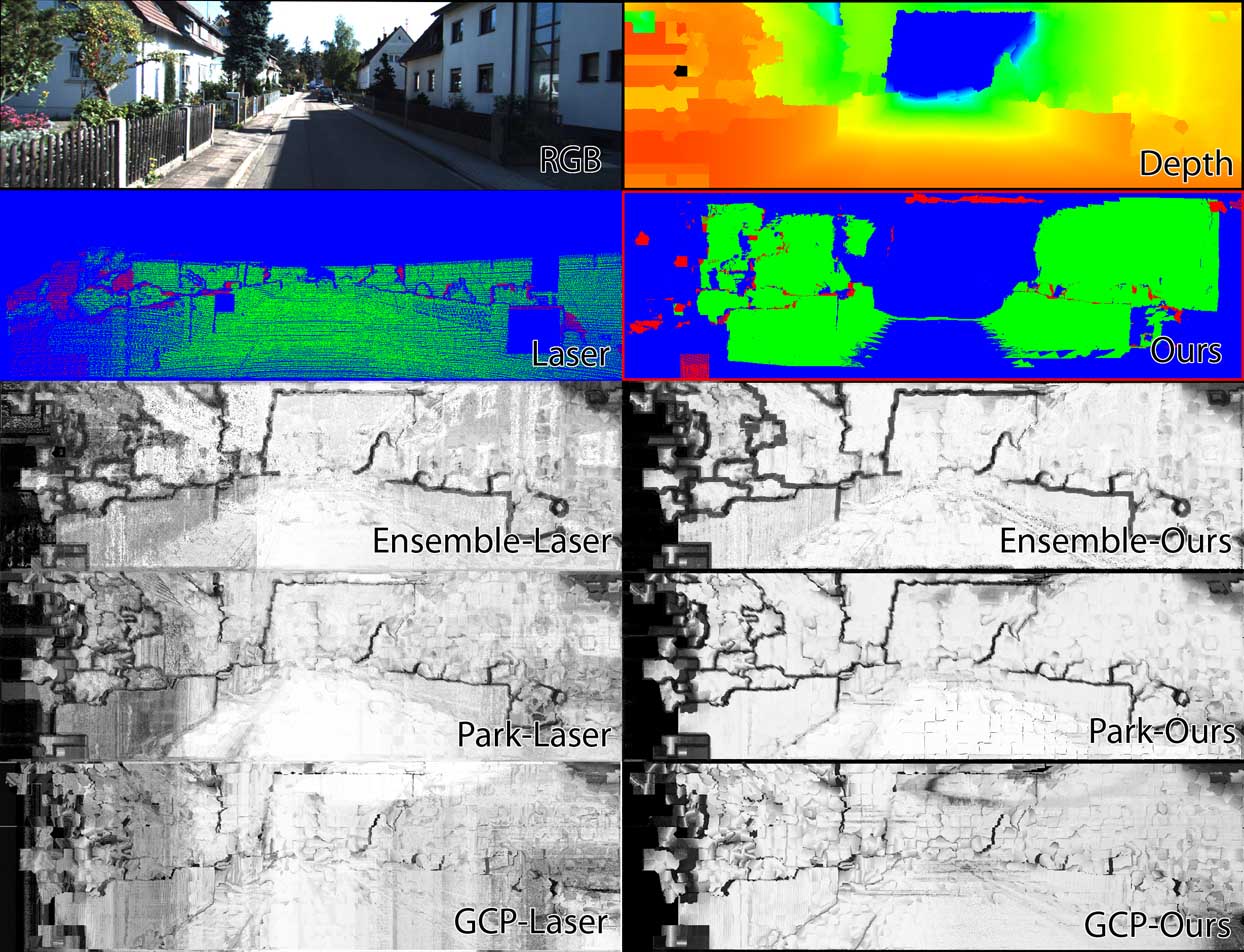}

%\end{minipage}
    \caption{Visual comparison of outputs for frame 35 and SPS~\cite{yamaguchi14} as query algorithm (rank 185 - worst). Top row: RGB input image and SPS~\cite{yamaguchi14} depthmap.
    The color in the depth images ranges from blue (far away) to red (very close).
    Second row: Label images computed with laser ground truth and with our approach.
    In the label images the color green stands for positive samples,
    red for negative and blue is ignored during training and evaluation.
    In the remaining rows, we display the confidence prediction output for all combinations of 
    confidence prediction algorithm (Ensemble~\cite{haeusler13}, GCP~\cite{spyro14}, Park~\cite{park15}) and
    training data (Laser and Ours). The color ranges from black (low confidence) to white (high confidence).}
  \label{fig:sps035}
\end{figure*}

\begin{figure*}[p]

  \centering

                \includegraphics[width=1\textwidth]{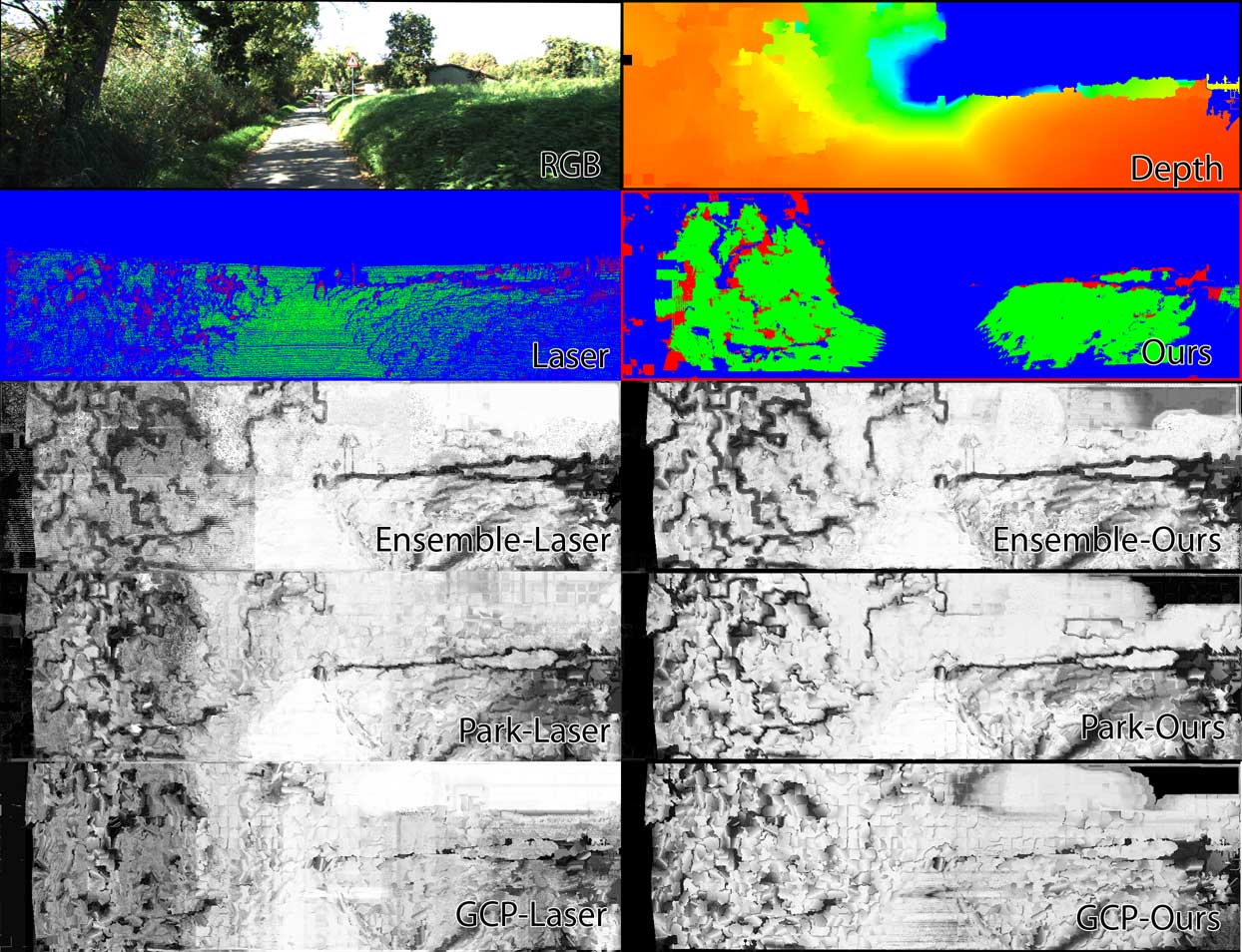}

%\end{minipage}
    \caption{Visual comparison of outputs for frame 42 and SPS~\cite{yamaguchi14} as query algorithm (rank 186 - worst). Top row: RGB input image and SPS~\cite{yamaguchi14} depthmap.
    The color in the depth images ranges from blue (far away) to red (very close).
    Second row: Label images computed with laser ground truth and with our approach.
    In the label images the color green stands for positive samples,
    red for negative and blue is ignored during training and evaluation.
    In the remaining rows, we display the confidence prediction output for all combinations of 
    confidence prediction algorithm (Ensemble~\cite{haeusler13}, GCP~\cite{spyro14}, Park~\cite{park15}) and
    training data (Laser and Ours). The color ranges from black (low confidence) to white (high confidence).}
  \label{fig:sps042}
\end{figure*}

\begin{figure*}[p]
  \centering

                \includegraphics[width=1\textwidth]{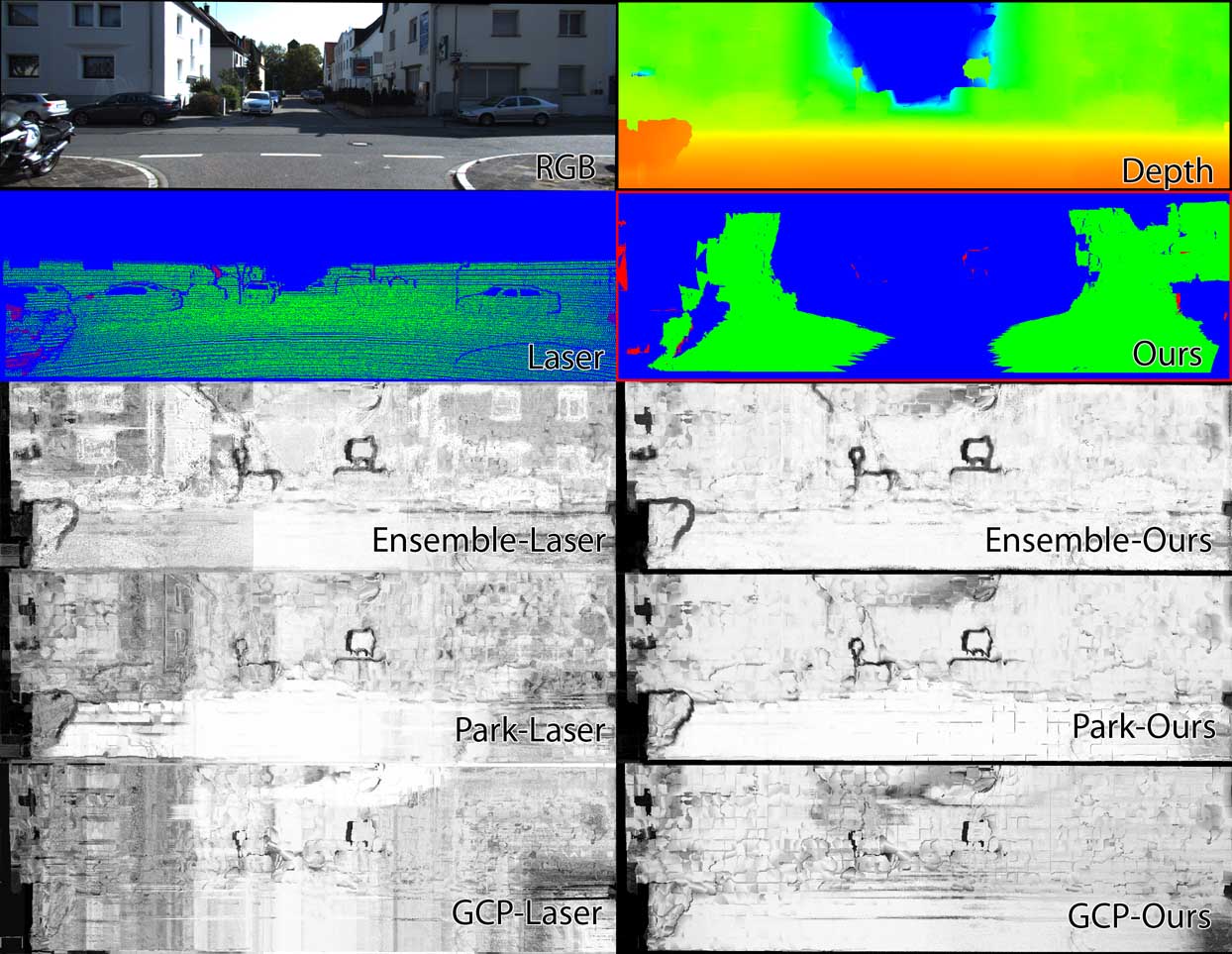}

%\end{minipage}
    \caption{Visual comparison of outputs for frame 56 and SPS~\cite{yamaguchi14} as query algorithm (rank 2 - best). Top row: RGB input image and SPS~\cite{yamaguchi14} depthmap.
    The color in the depth images ranges from blue (far away) to red (very close).
    Second row: Label images computed with laser ground truth and with our approach.
    In the label images the color green stands for positive samples,
    red for negative and blue is ignored during training and evaluation.
    In the remaining rows, we display the confidence prediction output for all combinations of 
    confidence prediction algorithm (Ensemble~\cite{haeusler13}, GCP~\cite{spyro14}, Park~\cite{park15}) and
    training data (Laser and Ours). The color ranges from black (low confidence) to white (high confidence).}
  \label{fig:sps056}
\end{figure*}

\begin{figure*}
  \centering
                \includegraphics[width=1\textwidth]{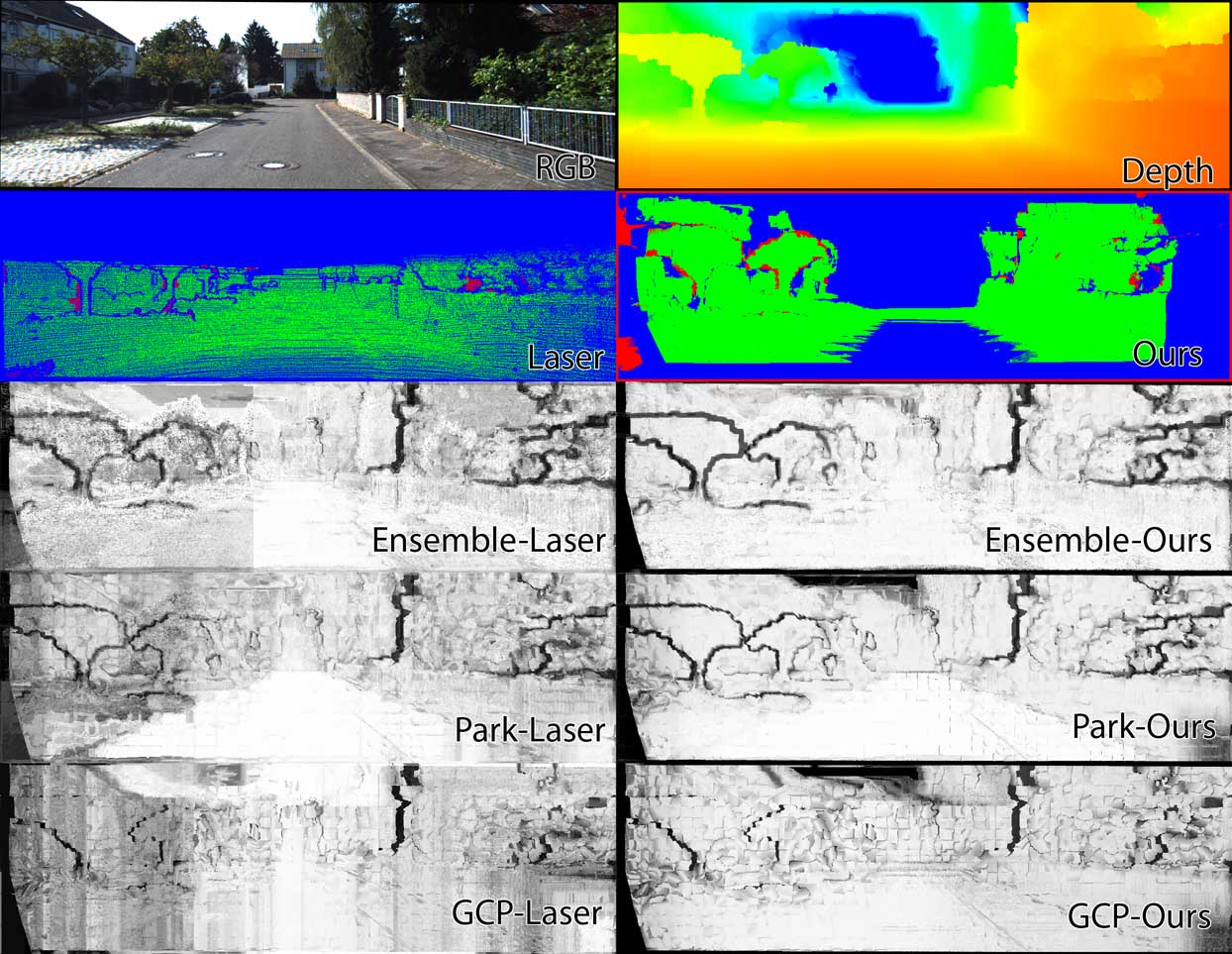}     
    \caption{Visual comparison of outputs for frame 85 and SPS~\cite{yamaguchi14} as query algorithm (rank 92 - median). Top row: RGB input image and SPS~\cite{yamaguchi14} depthmap.
    The color in the depth images ranges from blue (far away) to red (very close).
    Second row: Label images computed with laser ground truth and with our approach.
    In the label images the color green stands for positive samples,
    red for negative and blue is ignored during training and evaluation.
    In the remaining rows, we display the confidence prediction output for all combinations of 
    confidence prediction algorithm (Ensemble~\cite{haeusler13}, GCP~\cite{spyro14}, Park~\cite{park15}) and
    training data (Laser and Ours). The color ranges from black (low confidence) to white (high confidence).}
  \label{fig:sps085}
\end{figure*}

\begin{figure*} 
  \centering
                \includegraphics[width=1\textwidth]{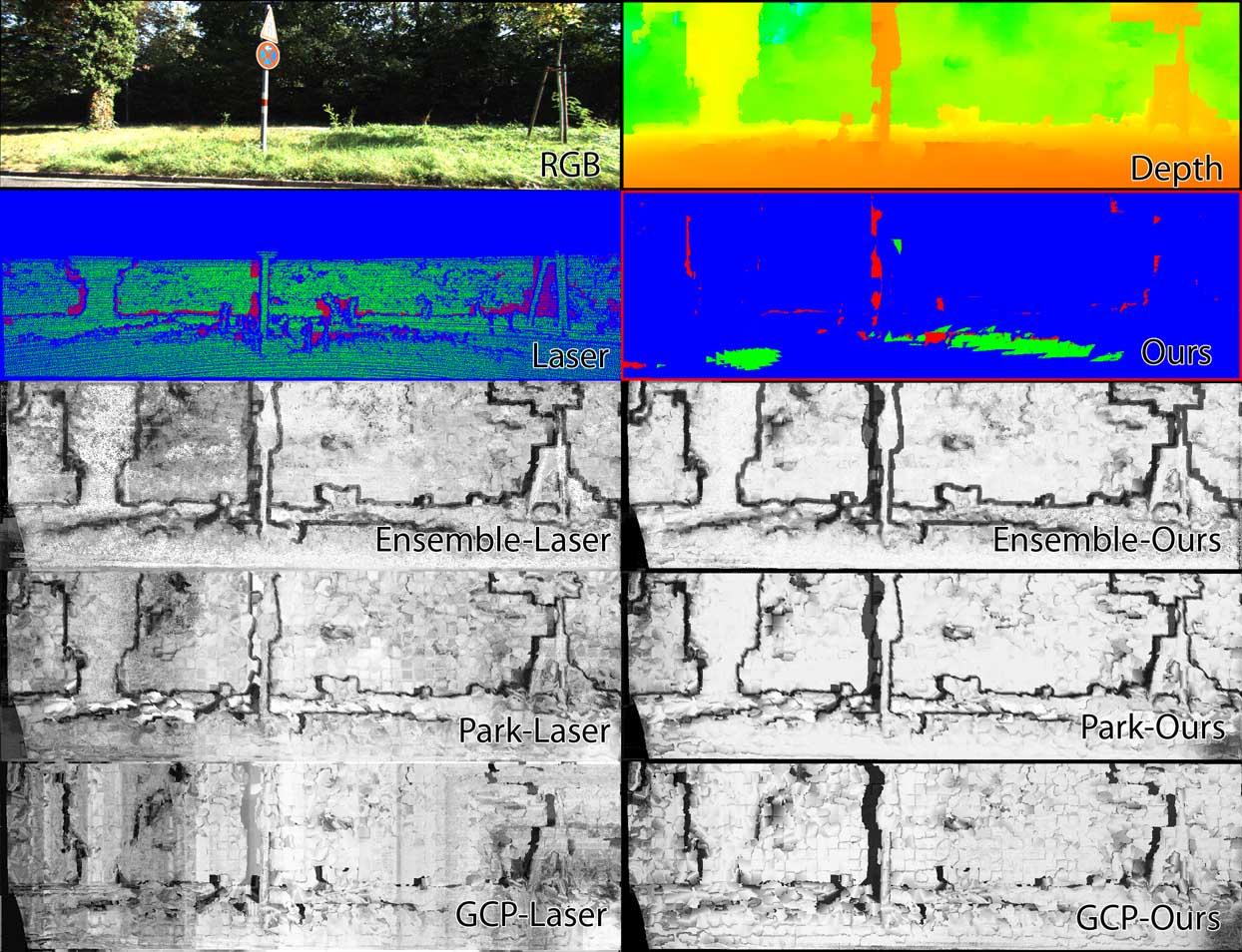} 
    \caption{Visual comparison of outputs for frame 88 and SPS~\cite{yamaguchi14} as query algorithm (rank 94- median). Top row: RGB input image and SPS~\cite{yamaguchi14} depthmap.
    The color in the depth images ranges from blue (far away) to red (very close).
    Second row: Label images computed with laser ground truth and with our approach.
    In the label images the color green stands for positive samples,
    red for negative and blue is ignored during training and evaluation.
    In the remaining rows, we display the confidence prediction output for all combinations of 
    confidence prediction algorithm (Ensemble~\cite{haeusler13}, GCP~\cite{spyro14}, Park~\cite{park15}) and
    training data (Laser and Ours). The color ranges from black (low confidence) to white (high confidence).}
  \label{fig:sps088}
\end{figure*}

\begin{figure*}[p]

  \centering

                \includegraphics[width=1\textwidth]{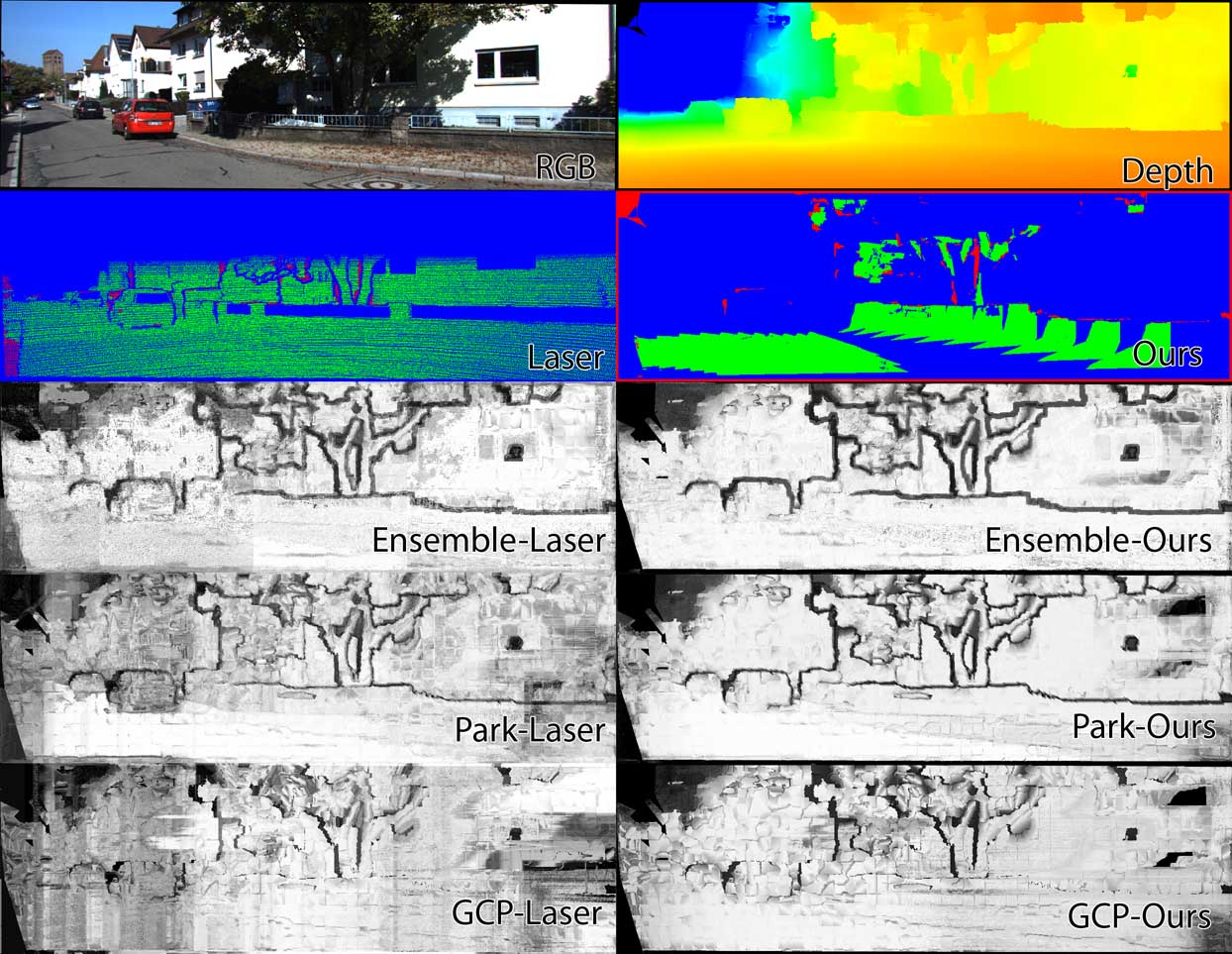}

%\end{minipage}
    \caption{Visual comparison of outputs for frame 102 and SPS~\cite{yamaguchi14} as query algorithm (rank 1 - best). Top row: RGB input image and SPS~\cite{yamaguchi14} depthmap.
    The color in the depth images ranges from blue (far away) to red (very close).
    Second row: Label images computed with laser ground truth and with our approach.
    In the label images the color green stands for positive samples,
    red for negative and blue is ignored during training and evaluation.
    In the remaining rows, we display the confidence prediction output for all combinations of 
    confidence prediction algorithm (Ensemble~\cite{haeusler13}, GCP~\cite{spyro14}, Park~\cite{park15}) and
    training data (Laser and Ours). The color ranges from black (low confidence) to white (high confidence).}
  \label{fig:sps102}
\end{figure*}

\begin{figure*}[p]

  \centering

                \includegraphics[width=1\textwidth]{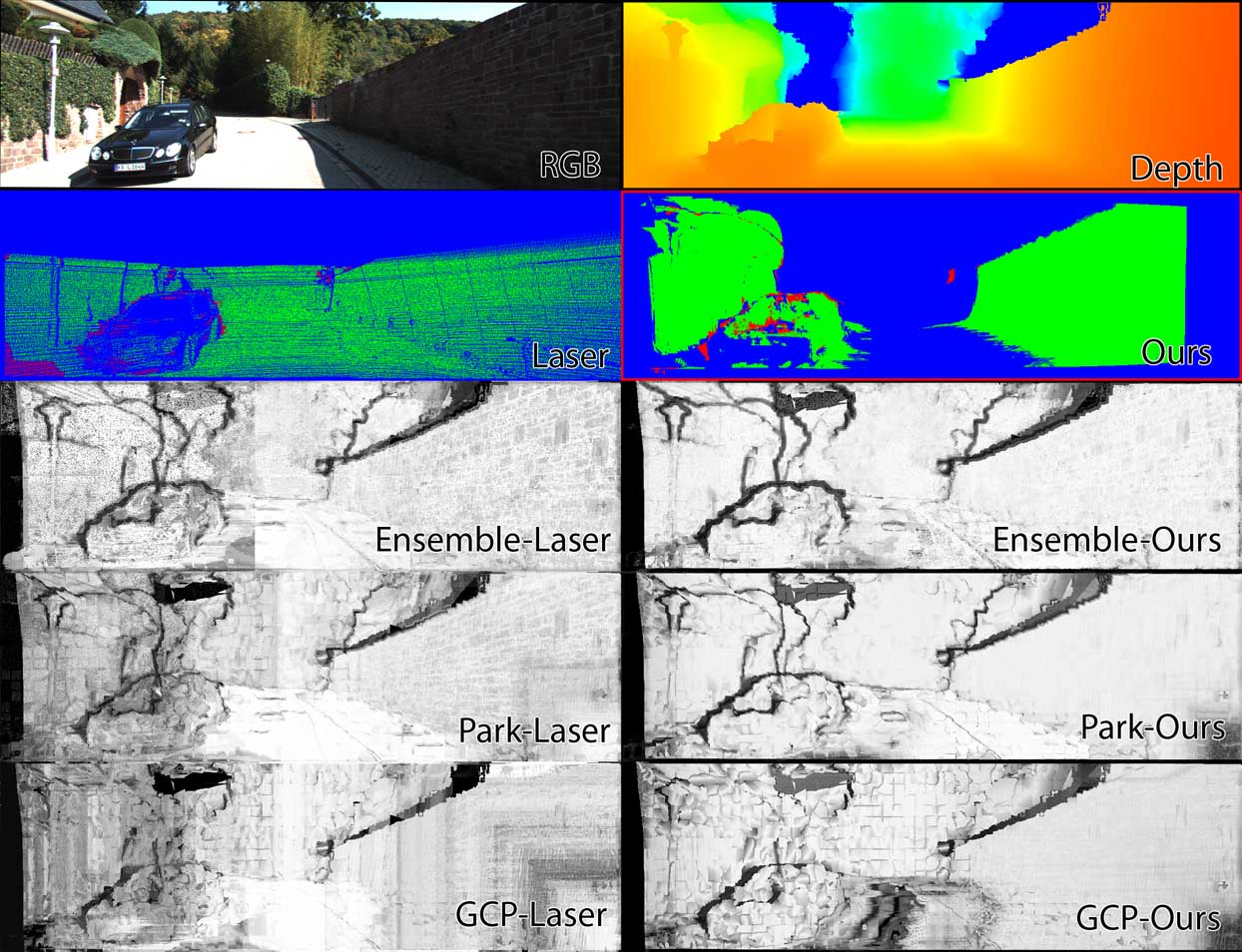}

%\end{minipage}
    \caption{Visual comparison of outputs for frame 126 and SPS~\cite{yamaguchi14} as query algorithm (rank 93 - median). Top row: RGB input image and SPS~\cite{yamaguchi14} depthmap.
    The color in the depth images ranges from blue (far away) to red (very close).
    Second row: Label images computed with laser ground truth and with our approach.
    In the label images the color green stands for positive samples,
    red for negative and blue is ignored during training and evaluation.
    In the remaining rows, we display the confidence prediction output for all combinations of 
    confidence prediction algorithm (Ensemble~\cite{haeusler13}, GCP~\cite{spyro14}, Park~\cite{park15}) and
    training data (Laser and Ours). The color ranges from black (low confidence) to white (high confidence).}
  \label{fig:sps126}
\end{figure*}

\begin{figure*}[p]

  \centering

                \includegraphics[width=1\textwidth]{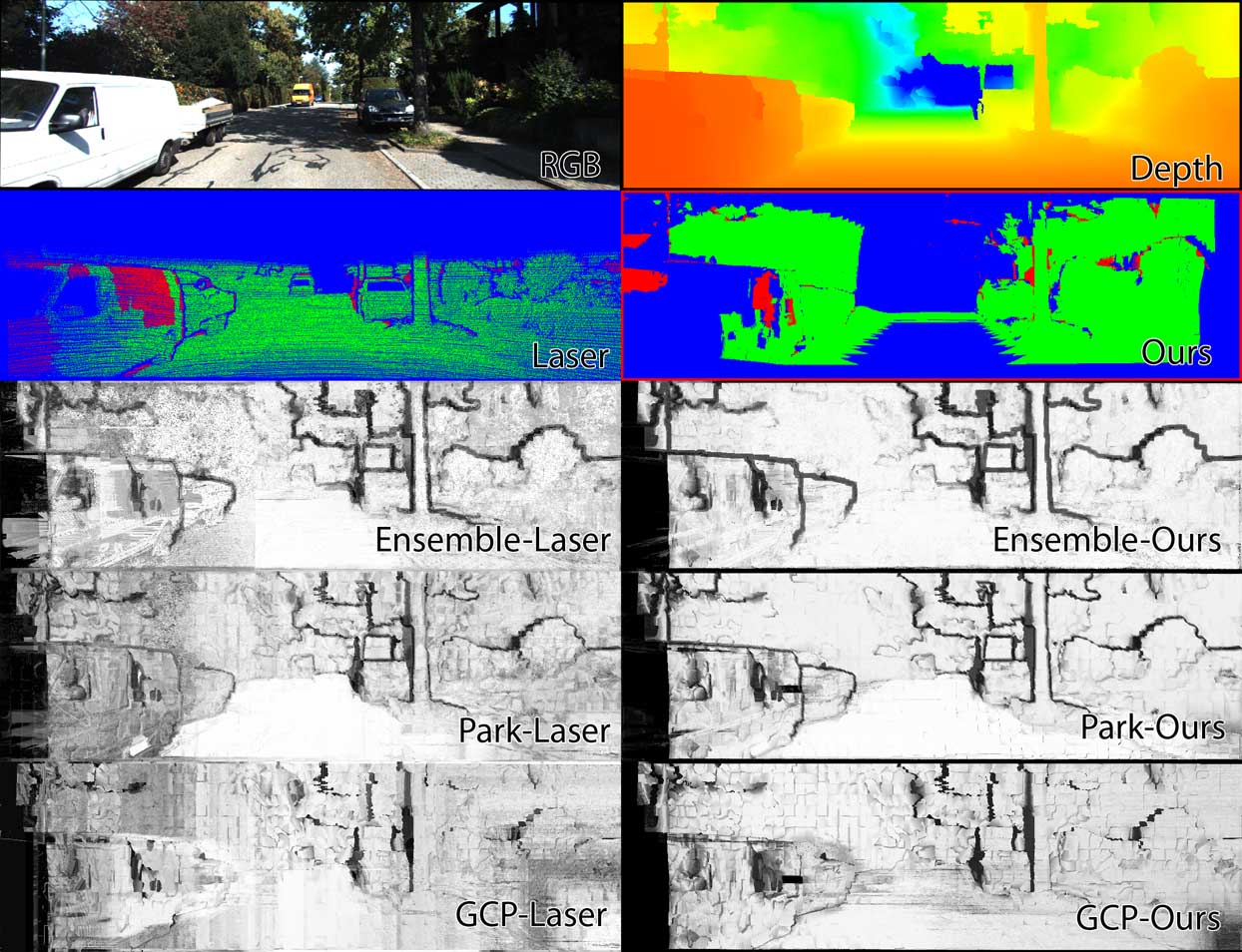}

%\end{minipage}
    \caption{Visual comparison of outputs for frame 151 and SPS~\cite{yamaguchi14} as query algorithm (rank 184 -  worst). Top row: RGB input image and SPS~\cite{yamaguchi14} depthmap.
    The color in the depth images ranges from blue (far away) to red (very close).
    Second row: Label images computed with laser ground truth and with our approach.
    In the label images the color green stands for positive samples,
    red for negative and blue is ignored during training and evaluation.
    In the remaining rows, we display the confidence prediction output for all combinations of 
    confidence prediction algorithm (Ensemble~\cite{haeusler13}, GCP~\cite{spyro14}, Park~\cite{park15}) and
    training data (Laser and Ours). The color ranges from black (low confidence) to white (high confidence).}
  \label{fig:sps151}
\end{figure*}

\FloatBarrier

\begin{figure*}[t]
      %\vspace{-15pt}
  \centering
%\begin{minipage}{1\linewidth}
    \subfigure%[SGM Frame 0.]
    {
\includegraphics[width=1\textwidth]{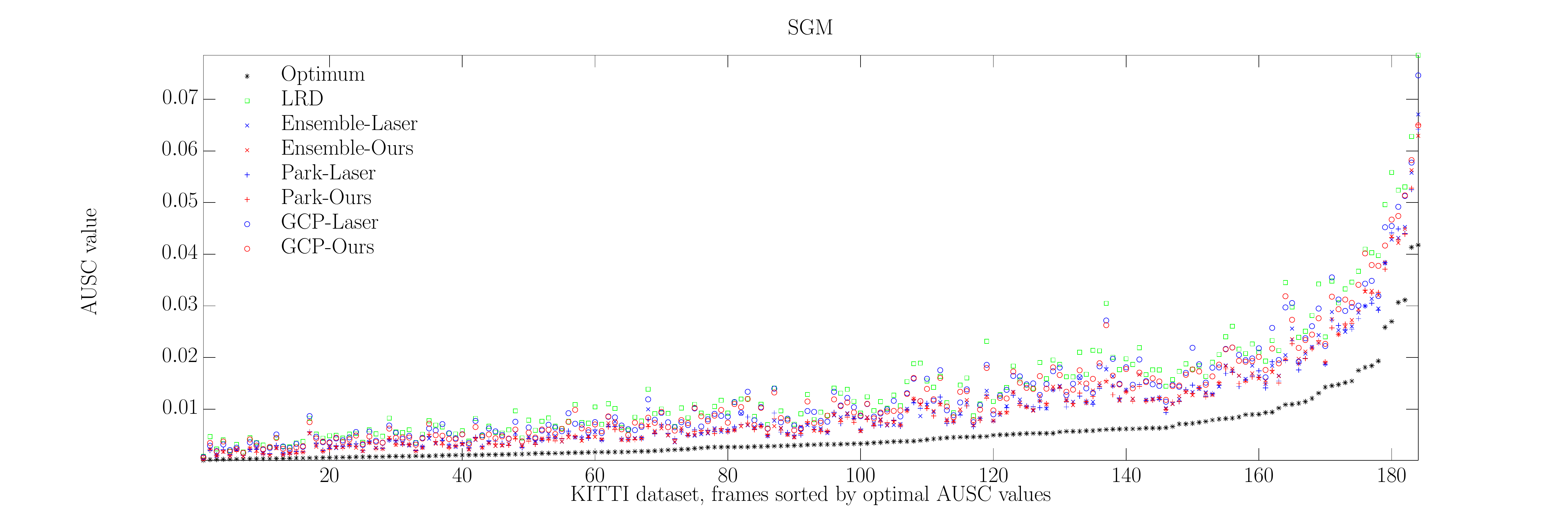} 
} \vspace{-20pt}\quad
\subfigure%[SGM Frame 1.]
    {
\includegraphics[width=1\textwidth]{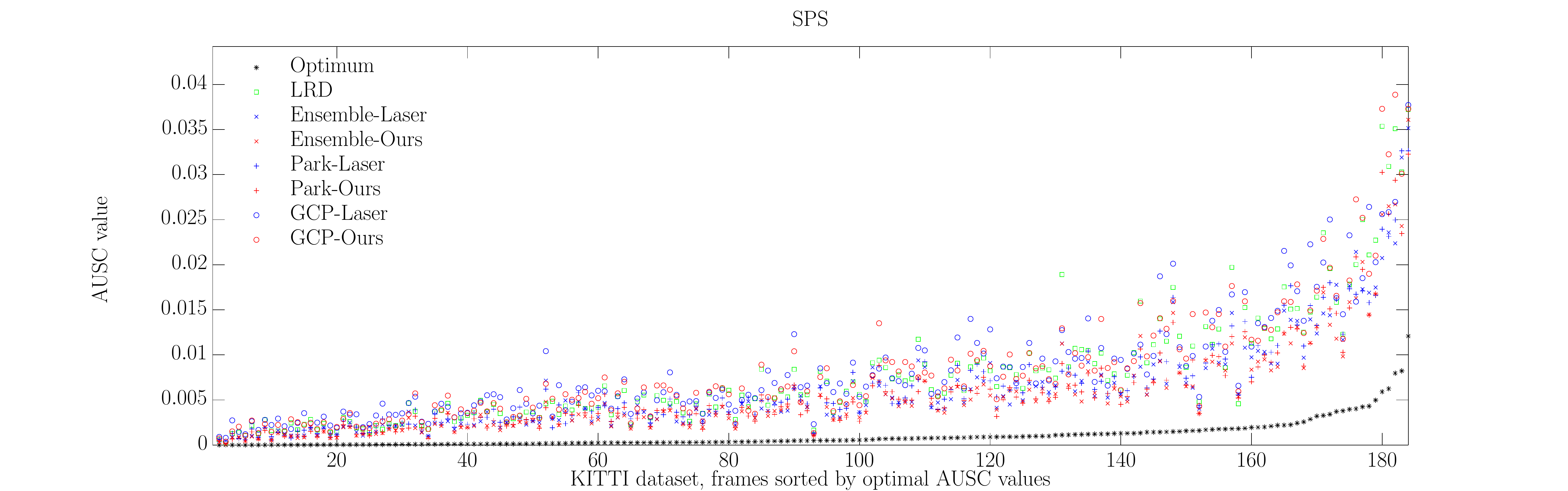} 
    }
  \caption{Area under the Sparsification Curve (AUSC) values for all frames of the KITTI training dataset minus the eight frames used for training. The frames were sorted according to the 
  optimal area under the curve value. We display all combinations of 
    query  algorithm (SGM~\cite{rothermel12} and SPS~\cite{yamaguchi14}), confidence prediction algorithm (Ensemble~\cite{haeusler13},GCP~\cite{spyro14},Park~\cite{park15}) and
    training data(Laser and Ours). As a baseline method we also show the Left-Right disparity Difference (LRD).}
  %\vspace{-15pt}
  \label{fig:kitti_ausc}
\end{figure*}

\end{document}